\documentclass[journal]{IEEEtran}
\usepackage[export]{adjustbox}
\usepackage{subcaption}
\usepackage{amsmath}
\usepackage{commath}
\usepackage[linesnumbered,boxed]{algorithm2e}
\usepackage{graphicx}
\usepackage{multirow}
\usepackage{tabularx}
\usepackage{booktabs}
\usepackage{float}
\graphicspath{ {Figures/} }
\usepackage{amsmath}
\DeclareMathOperator*{\argmax}{argmax} 
\usepackage{balance}

\makeatletter
\newcommand{\removelatexerror}{\let\@latex@error\@gobble}
\makeatother

\ifCLASSINFOpdf
\else
\fi
\begin{document}
\bibliographystyle{IEEEtran}
%

\title{System Architecture for Real-time Surface Inspection Using Multiple UAVs}

%
%
%

\author{V. T.~Hoang, M. D.~Phung, T. H.~Dinh, and~Q. P.~Ha 
\thanks{V. T.~Hoang, T. H.~Dinh, and~Q. P.~Ha are with School of Electrical and Data Engineering, University of Technology Sydney (UTS), 81 Broadway, Ultimo NSW 2007, Australia.
        {\tt\small \{VanTruong.Hoang, TranHiep.Dinh, Quang.Ha\}@uts.edu.au}.}
\thanks{M. D.~Phung is with University of Engineering and Technology, Vietnam National University (VNU), 144 Xuan Thuy, Cau Giay, Hanoi, Vietnam.
		{\tt\small duongpm@vnu.edu.vn.}}
\thanks{}}

\maketitle

\begin{abstract}
This paper presents a real-time control system for surface inspection using multiple unmanned aerial vehicles (UAVs). The UAVs are coordinated in a specific formation to collect data of the inspecting objects. The communication platform for data transmission is based on the Internet of Things (IoT). In the proposed architecture, the UAV formation is established via using the angle-encoded particle swarm optimisation to generate an inspecting path and redistribute it to each UAV where communication links are embedded with an IoT board for network and data processing capabilities. Data collected are transmitted in real time through the network to remote computational units. To detect potential damage or defects, an online image processing technique is proposed and implemented based on histograms. Extensive simulation, experiments and comparisons have been conducted to verify the validity and performance of the proposed system.
\end{abstract}


%
\IEEEpeerreviewmaketitle

\section{Introduction}
%
%
%
%
Continuous monitoring of health conditions of built infrastructure is essential for their safe operation and long-term serviceability. Since structural deterioration is quite often revealed through visual appearance, the foremost task to be conducted in monitoring is typically surface inspection. This task can be carried out by three main approaches including using foot patrol, ground vehicles, and flying-assisted robots. While the first two approaches are common, the use of flying vehicles like UAVs is receiving increasing interest thanks to their flexibility in operation, versatility in task allocation and capability of conducting non-intrusive inspection \cite{tatum2017}. For instance, UAVs have been used for periodical inspection of critical infrastructures such as oil-gas pipelines, wind turbine blades and power lines \cite{wang2017,zhang2017}. They also have been employed to collect data for defect/damage recognition such as cracks, rust or structural misalignment \cite{morgenthal2014, chen2017}. 

Systems for robotic inspection using UAVs are often designed in a layer-based architecture. At the top layer, the UAV is programmed to fly close to the objects to collect inspection data. They are then post-processed to detect potential damage/defects via techniques such as photogrammetry \cite{sanchez2014} or Haar-like features extraction \cite{wang2017}. The middle layer involves the generation of inspection trajectories whereby path planning algorithms  such as A-star, Dijkstra, rapidly-exploring random tree and probabilistic roadmap can be employed \cite{Goerzen2009}. However, algorithms specifically designed for UAVs using particle swarm optimization \cite{phung2017} or iterative viewpoint resampling  \cite{bircher2015} are typically more efficient by considering also the coverage among waypoints. At the lowest layer, various techniques have been proposed to track the generated paths, using  backstepping, model predictive control, PID control or sliding mode control \cite{mathe2015vision} with sufficient performance. 

While the use of single UAVs has proved its applicability in inspection, recent works start to study the coordination of  UAVs in a group to achieve higher time efficiency, performance and resilience \cite{deng2014, lim2016}. A multi-platform UAV system for power line inspection tasks \cite{deng2014} shows that the time to inspect 70 km of the power line can be reduced to 3 hours compared to a week as with the traditional inspection method. Therefore, different aspects of multiple UAV coordination such as formation control, collision avoidance, and data acquisition have been addressed \cite{Lwowski2018,tanner2004}. 

In a control architecture for inspection using multiple UAVs, the synchronization in control, path planning and data processing is typically required not only between layers within single UAVs but also among them. For real-time inspection, communication channels are therefore needed among UAVs, ground control stations (GCS), and other remote computational units (RCU) such as server computers or cloud computing systems to tackle the high demand of data fusion and processing. Those requirements pose the need for a homogeneous communication platform that allows all components of an inspection system to be integrated. The current use of radio transceiver modules with a pulse mode modulation 2.4GHz uplink and 2.4/5.8GHz downlink is insufficient as they can only form a private network among on-site devices \cite{Stingu2009}. Some studies suggested deploying additional UAVs as communication relay stations to extend the network \cite{KIM201442,Cetin2012}. This approach is suitable for tasks covering a local range but not large areas like bridges, power lines or wind turbines which require hundreds of kilometres in communication range. Several studies proposed to use satellite communication to overcome this problem \cite{deng2014,Zeng2016}. It is however expensive, complex and not always feasible. 

In this study, the Internet of Things (IoT) is used as the communication platform to interconnect UAVs and devices in order to efficiently perform inspection tasks. Based on this platform, a robust low-level controller is introduced for each UAV to track the inspection path under harsh working conditions. The high-level control is then developed to coordinate multiple UAVs via formation and path-planning modules. The formation configuration is decided based on the results of the mission assessment and defect detection requirements. A multi-objective particle swarm optimisation algorithm using angle-encoded PSO ($\theta$-PSO) is exploited to generate an optimal path for the centroid of the group. This path is finally translated into individual trajectories for UAVs to follow. The images captured during the flight are streamed to RCU where they are processed to detect potential defects. A detection algorithm based on histogram analysis is proposed for the image processing tasks. Extensive simulations and experiments have been conducted in real and practical scenarios to evaluate the validity and feasibility of the proposed system.

\section{Related works} \label{review}
A number of studies have been devoted to the problem of surface inspection using UAVs. Early works typically targeted simple and linear structures such as highways, roads, and canals \cite{Rathinam2008, metni2007}. They focused on low-level control laws using the data acquired from the incorporated visual systems for quasi-stationary flights above planar surfaces. Later, the inspection of hardly accessible structures in GPS-denied environments using UAVs has been proposed using on-board Simultaneous Localisation and Mapping (SLAM) \cite{Winkvist2013}. While the positioning and self-recovering problems have been addressed, the system was intended for semi-autonomous operations that would require an operator to instruct the UAV. To improve the level of autonomy, images obtained by UAVs can be used to detect inspecting objects and create paths for autonomous flights \cite{Mejias2007, zhang2017}.

For automatic inspection, sophisticated systems featured path planning and path-following control algorithms have been proposed. In \cite{alexis2016}, an aerial robotic system for the contact-based surface inspection has been introduced using not only optimal trajectory tracking but also accurate force control techniques. In \cite{bircher2015}, an iterative viewpoint re-sampling path planning algorithm was proposed for the inspection of complex 3D structures. The inspection of outdoor structures under windy conditions was addressed in \cite{guerrero2013} by using the viewpoint selection and optimal time route algorithms. Besides, an UAV-based inspection system for wind turbines was developed in \cite{Schafer2016} with the capability of creating smooth and collision-free flight paths based on the data recorded from LIDAR sensors. These studies however focus more on data collection rather than defect detection.

In terms of surface inspection, several studies have been conducted for defect detection tasks. In \cite{ellenberg2016}, a fast and effective defect detection method has been proposed using the size-based estimation with data obtained from both color and infrared cameras. In \cite{wang2017}, Haar-like features and a cascading classifier were applied to UAV-taken images to identify cracks on wind turbine blade surfaces. Self-organizing map optimization was introduced in \cite{chen2017}, based on image recognition and processing model for crack detection to reduce human involvement. On another note, \cite{morgenthal2014} assessed the quality and feasibility of images taken by UAVs for defect detection and discussed methods to improve the image quality. 

The use of multi-UAV cooperation for infrastructure inspection has been recently investigated. A multi-platform UAV system for power line inspection was experimented in \cite{deng2014}. Similarly, a multi-UAV positioning and routing for power network damage assessment was introduced in \cite{lim2016} whereas some other studies developed inspection systems for buildings or bridges \cite{banaszek2017, mathe2015vision}. However, those systems are rather limited in communication and coordination capability. In fact, it can be seen from the literature that UAV-based inspection is still in its early development stage. There exists a crucial need for an architecture that not only enables the integration of various functional modules like path planning, high-level and low-level control but also allows for the real-time coordination of multiple UAVs within a homogeneous communication platform for efficient surface inspection.
 
\begin{figure*}
	\centering
	\includegraphics[scale=0.4]{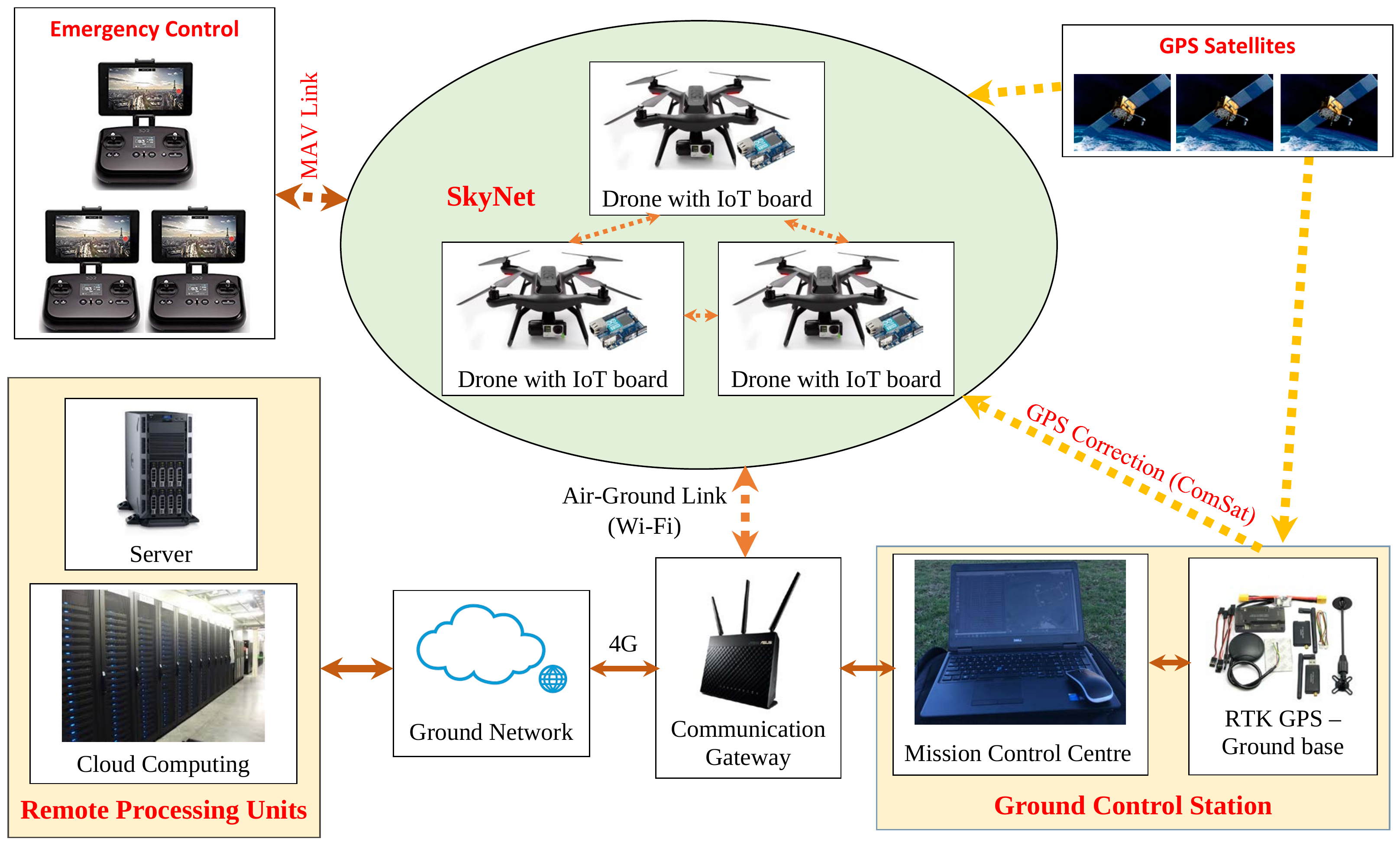}
	\caption{Data communication structure}
	\label{Figcomsat}
\end{figure*}

\section{System Overview and Data Communication} \label{Overview}
Figure \ref{Figcomsat} presents an overview of the proposed inspection system. Its core includes three UAVs equipped with cameras to collect data of the inspected structure. The UAVs, via IoT boards, form a private network called SkyNet for data exchanges during their cooperative flight. At the same time, they also share data with GCS via a wireless gateway router that relays the data via the Internet to RCU to be processed further. The UAVs localization is conducted via GPS modules combined with inertial measurement units (IMUs) to monitor internal states of the drones. Depending on the inspection task, real-time kinematic (RTK) GPS modules will be used to improve positioning accuracy. The operation of each UAV is monitored via its accompanied remote controller which will take over the control in emergency situations. 

\subsection{IoT devices}
As illustrated in Fig.\ref{Figcomsat}, the inspection system involves real-time cooperation of many components within the control architecture. To equip them with networking and data processing capabilities, IoT boards are installed, each includes a processor and a microcontroller as shown in Fig. \ref{drone}. With the processor, the IoT-based devices can connect to the wifi network as well as process the receiving data via the Linux operating system installed. The microcontroller enables the devices to work with other low hardware interfaces such as voltage/current interface, AD/DA converter, PWM, etc. The boards thus turn a normal device into a smart one that can be integrated into the Internet protocol (IP) based networks. Nevertheless, the built-in antennas of IoT boards designed for indoor applications are insufficient for outdoor communication which requires high gain external antennas. Therefore, a 6 dBi detachable antenna is placed on the drone to extend the communication range and provide stable signals, especially for moving objects like UAVs. In addition, mobile broadband networks are made available using wireless gateway routers that have SIM card slots. As mobile broadband networks present almost everywhere, this approach allows the inspection system to be deployed for any surface without the need for relay stations nor satellite communication. In remote areas where the mobile network is not available, this approach would be feasible but requires extra equipment to connect to the nearest available access point.

\subsection{Communication protocols}
Along with the hardware, transport protocols play an important role in ensuring the security and efficiency of the data exchanged. The most popular transport protocols for IoT include the Transmission Control Protocol (TCP), User Datagram Protocol (UDP), and Real-Time Transport Protocol (RTP). While TCP was originally designed for the reliable transmission of static data over low-bandwidth, high-error-rate networks, UDP could send datagrams from a device to another as fast as possible but only under good network conditions without considering the state of the network. Developed  for delivering real-time multimedia data, RTP can facilitate jitter compensation and detection of out-of-sequence arrival in data. According to those protocol features, RTP is chosen in our systems architecture for photo/video streaming and UDP for sensing data transmitting whereas TCP for delivering administrative data and control commands.

\subsection{Data processing}
One of the main benefits of using IoT is the capability of conducting computation and data processing at various layers of the system, i.e., the device layer by IoT boards, the control layer by built-in computers of UAVs and the application layer by networked server computers or cloud computing services, depending on the amount of data to be processed and real-time requirements. In our system, IoT boards are used for processing communication data among UAVs such as position, velocity and other state information to minimize the computational latency. The control algorithm on the other hand is handled by the built-in computers of UAVs to enhance reliability. Other information and imaging data are processed by server computers to cope with the high demand for computation and energy consumption. 

\section{Data Acquisition and Control} \label{control}

\begin{figure}
	\centering
	\includegraphics[width=9cm]{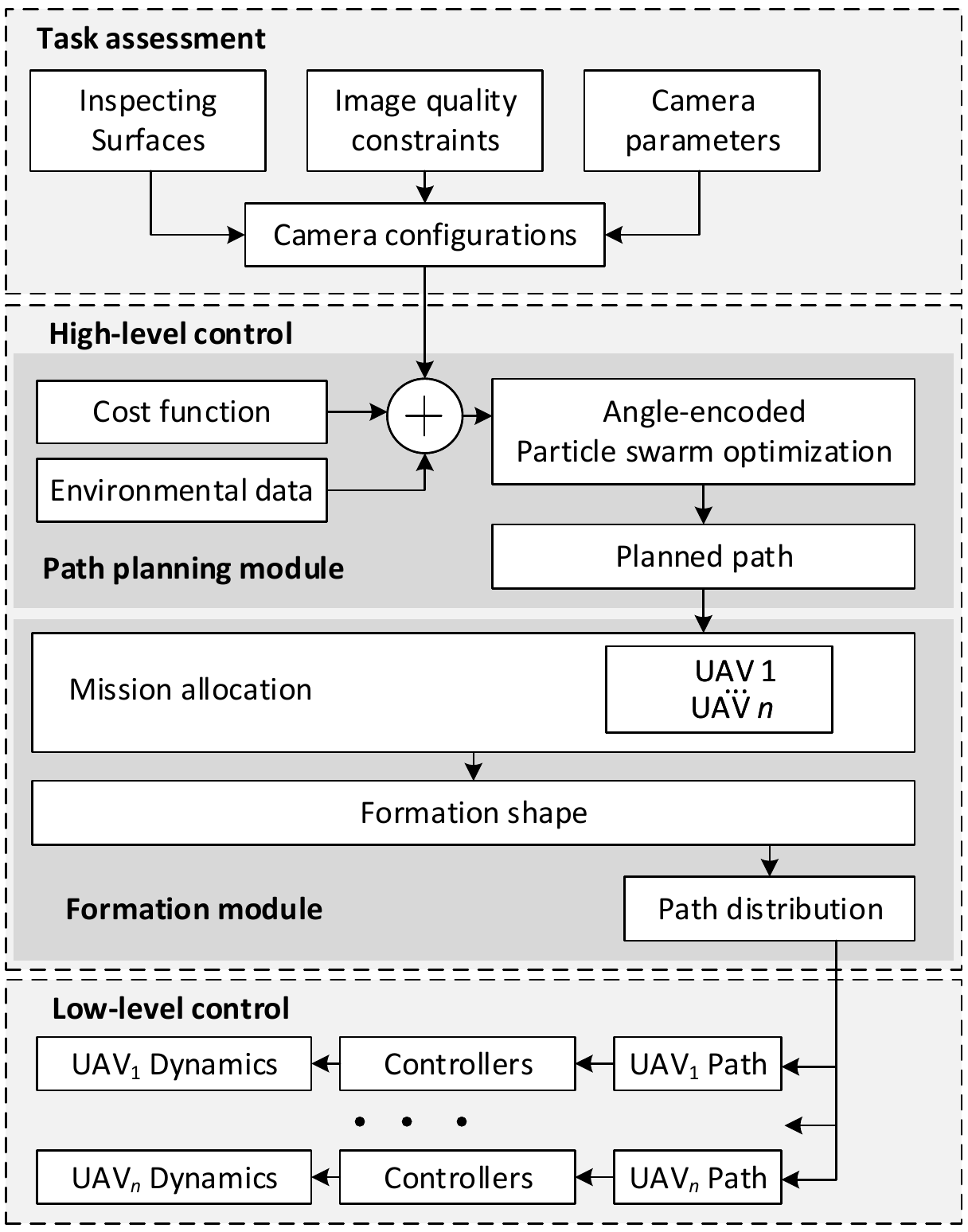}
	\caption{System architecture for multi-UAV surface inspection}
	\label{FigSys_Archi}
\end{figure}

Based on the data communication and processing framework, we propose an architecture for data acquisition and control consisting of three levels: task assessment, high-level control and low-level control as shown in Fig. \ref{FigSys_Archi}. Details are described as follows.

\subsection{Task assessment}
Given the surface to be inspected, at the task assessment level, a set of camera configurations is computed, covering all areas of interest. Each configuration $c_i$ corresponds to a position and an orientation of the camera, and subject to constraints on image quality as per the following requirements: 
\begin{itemize}
\item[(i)] images are taken when the camera directed perpendicularly to the inspected surface; 
\item[(ii)] their resolution is sufficiently high to distinguish the required smallest feature, $s_f$; and
\item[(iii)] image overlapping is specified by the sticking algorithm to a given percentage, $o_p$.
\end{itemize}

The first requirement confines the camera orientation to the normal of the inspected surface. The resolution requirement suggests the computation of the field of view of the camera as:
\begin{equation} \label{eq_cam_fov}
a_{fov} = \dfrac{1}{2} r_c s_f,
\end{equation}
where $r_c$ is the camera resolution. Thus, the distance $d_{f,i}$ from the $i$th UAV to the surface under inspection can be found as: 
\begin{equation} \label{uav_distance}
d_{f,i} = \dfrac{a_{fov} f}{s_s} ,
\end{equation}
where $f$ and $s_s$ are respectively the focal length and sensor size of the camera. Let $G$ be a finite set of geometric primitives $g_i$ for the whole inspected surface, with each $g_i$ corresponding to a surface patch covered by a camera shot. Taken the overlapping percentage into account, the geometric primitive $g_i$ is computed as:
\begin{equation} \label{uav_capture}
g_i = (1 - o_p )a_{fov}.
\end{equation}
By using (\ref{uav_distance}) and (\ref{uav_capture}), configurations $c_i$ can be sufficiently determined to cover the set of primitives $G$ \cite{phung2017}, which is then fed to the high-level control layer for path planning. 

\subsection{High-level control} \label{HL_Control}
The high-level control consists of two modules, path planning and formation control the generation respectively of the reference path computed from the task assessment, and of trajectories for each UAV based on the reference path and the desired formation shape. 

\subsubsection{Path Planning}  \label{Path}
When producing a path for the desired motion of multiple UAVs in a group, a number of constraints are required for formation maintenance, UAV maneuverability, operating space, and obstacle avoidance. In this work, all constraints will be incorporated into a multi-objective function. The path planning problem then can be simplified to creating a reference path for the formation centroid. Since our goal is to construct optimal paths for all UAVs in the group, it is essential to speed up the convergence of the optimization process for the whole formation. Therefore, we propose to use the angle-coded PSO ($\theta$-PSO) \cite{van2018}. 

In $\theta$-PSO, a set of particles is generated, each seeks for an optimal solution by propagating to compromise between its own experience and the social experience. Initially, each particle is assigned a random position, $x_i$. This position is represented by a phase angle $\theta_{i}$ of the UAV. The particle motion is then updated by the following equations:
\begin{align}\label{EqThePSO}
\begin{cases} \Delta \theta_{ij}^{k+1} = w \Delta \theta_{ij}^k + c_1 r_{1i}^k (\lambda_{ij}^k - \theta_{ij}^k) + c_2 r_{2i}^k (\lambda_{gj}^k - \theta_{ij}^k) \\
\theta_{ij}^{k+1} = \theta_{ij}^k + \Delta \theta_{ij}^{k+1}, (i=1,2,...,N; j=1,2,...,S)\\
x_{ij}^k = \dfrac{1}{2}\left[(x_{max} -x_{min}) sin \left( \theta_{ij}^k \right) + x_{max} + x_{min} \right], \end{cases}
\end{align}
where $\theta_{ij} \in [-\pi/2, \pi/2]$ and $\Delta \theta_{ij} \in [-\pi/2, \pi/2]$ are respectively the phase angle and phase angle increment of the $i$th particle at iteration $j$ in the searching space; $\lambda_g = [\lambda_{g1}, \lambda_{g2},..., \lambda_{gS}]$ and $\lambda_i = [\lambda_{i1}, \lambda_{i2},..., \lambda_{iS}]$  are respectively the global and personal best positions; $N$ is the swarm size; $S$ is the searching space dimension; $w$ is the inertial weight; $r_{1i}$ and $r_{2i}$ are pseudorandom scalars; $c_1$ and $c_2$ are the gain coefficients; $x_{max}$ and $x_{min}$ are the upper and lower restrictions of the search space;  and subscript $k$ is the iteration index. 

In Eq. (\ref{EqThePSO}), it can be seen that a solution of a particle covers three alternative options: to track its private trajectory $\Delta \theta_{i}$, to follow its best prior position $\lambda_{i}$, or to move toward the global best position $\lambda_{g}$. The correlation among them depends on coefficients $w$, $c_1$ and $c_2$.  The values of $\lambda_{i}$ and $\lambda_{g}$ are evaluated based on the cost function in the following form:
\begin{equation} \label{cost_function}
J_F({\mathcal T}_{F}) = \sum \limits_{m=1}^3 \beta_m J_m({\mathcal T}_{F}),
\end{equation}
where ${\mathcal T}_{F}$ is the formation path; $\beta_m$ is the weighting factor indicating the corresponding threat intensity; and $J_m({\mathcal T}_{F})$, $m=1,2,3$, are the costs associated with the path length, collision violation and flying altitude, respectively. The cost function (\ref{cost_function}) is formulated from evaluating the length and violation cost of the path. The former helps to minimize the total travelling distance of the path whereas the latter is to avoid inter-UAV collisions and to avoid collisions of UAVs with static obstacles as identified by the Mission Planner incorporating a satellite map. It is noted that only static and known position obstacles are considered in this work. External dynamic obstacles can be avoided by using additional sensors such as ultrasonic sensors or Lidars with an extended path planning module. This topic is however beyond the scope of our paper. In operation, the UAVs need to maintain certain distances to the surface as described in (\ref{uav_distance}). Thus, the constraint in flying attitude is also added to the multi-objective cost function (\ref{cost_function}).

\subsubsection{UAV Formation} \label{Formation}
From the reference path generated by the proposed $\theta$-PSO algorithm for the formation centroid, it is necessary to produce a specific path for each UAV to maintain the shape of the formation during the flight. Those paths can be computed from the formation centroid path and the desired relative distances among the UAVs.

\begin{figure}
	\centering
	\includegraphics[width=7cm]{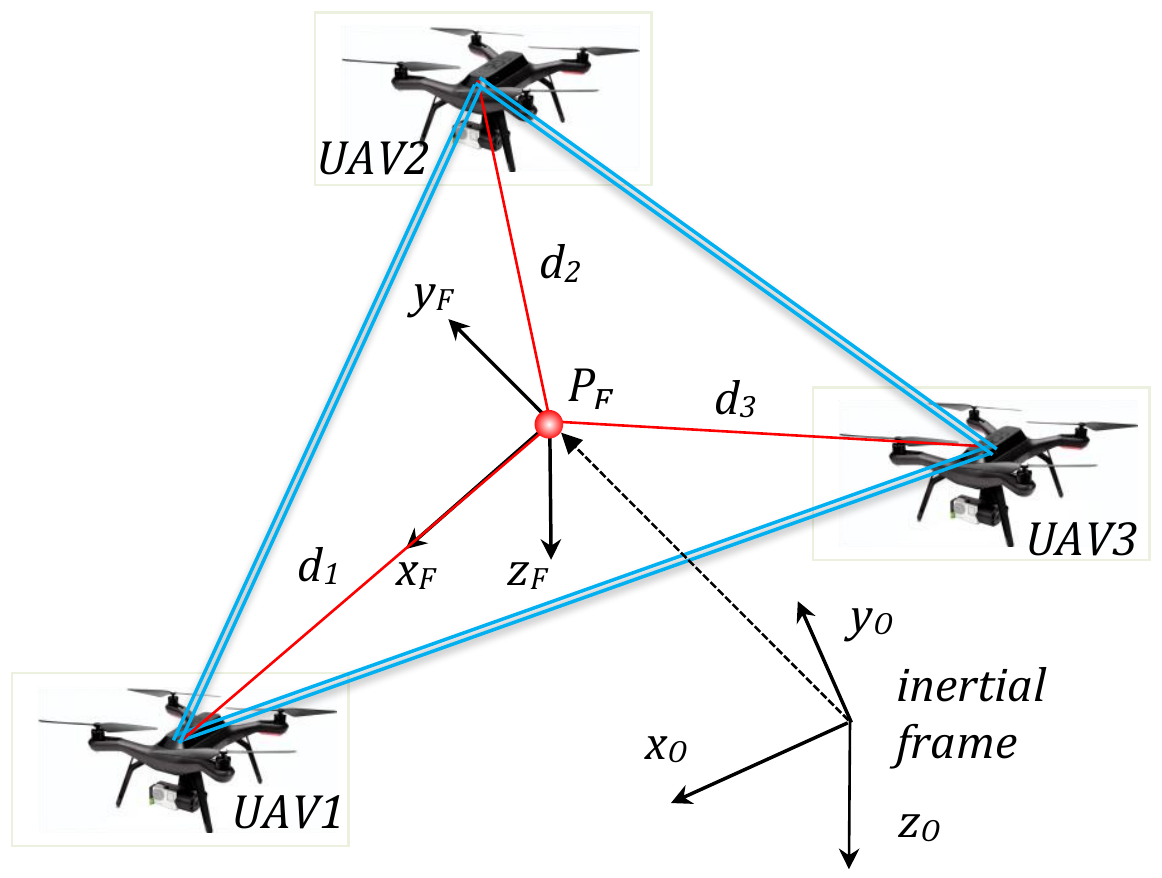}
	\caption{Inertial and formation frames in UAV formation}
	\label{form_fig}
\end{figure}

Figure \ref{form_fig} shows the inertial and formation frames that represent a triangular UAV formation used in this study. All measurements are referred to the inertial frame $O$ with axes $x_O, y_O$ and $z_O$.  Positions of UAV$_n$, $n=1,2,3$, in the inertial frame are denoted as $P_n = \{x_n, y_n,z_n\}$. The formation frame, $\{x_F, y_F, z_F\}$, is defined such that the origin $P_F$ is coincident with the centroid of the triangle. This allows to determine the centroid of the formation from the fixed inertial frame as:
\begin{equation}
P_F = \dfrac{1}{3} \sum\limits_{i=n}^3 P_n.
\end{equation}

The rotation matrix which represents the relation between the formation and inertial frames is given by:
\begin{align}\label{Eq_rotation_matrix2}
R_{OF} =  \left [\begin{array}{ccc} 
c_{\psi}c_{\theta} & c_{\psi}s_{\theta} s_{\phi}-s_{\psi}c_{\phi} & c_{\psi}s_{\theta} c_{\phi}+s_{\psi}s_{\phi}\\
s_{\psi}c_{\theta} & s_{\psi}s_{\theta} s_{\phi}+c_{\psi}c_{\phi} & s_{\psi}s_{\theta} c_{\phi}-c_{\psi}s_{\phi}\\
-s_{\theta} & c_{\theta} s_{\phi} & c_{\theta} c_{\phi}
\end{array} \right],
\end{align}
\noindent where $s_x = sin(x)$, $c_x = cos(x)$, and $\phi$, $\theta$ and $\psi$ are Euler angles of the shape. 

Let  $P_n = [x_n, y_n, z_n]^T$ and $P_{n,d} = [x_{n,d} , y_{n,d} , z_{n,d}]^T$ be respectively the actual and desired position for each UAV during the flight, in which $P_n$ can be obtained from GPS data of the UAV. The relative position errors of the $n$th UAV during the flight in the inertial frame are then defined as:
\begin{align} \label{eq_error}
\left[ \begin{array}{c}
e_{n,x} \\e_{n,y} \\e_{n,z}
\end{array} \right]  = \left[ \begin{array}{c}x_{n,d} -x_n\\ y_{n,d} - y_n \\ z_{n,d} - z_n\end{array} \right].
\end{align}
By using the rotation matrix in (\ref{Eq_rotation_matrix2}), where $R_{FO}(t) = R^{-1}_{OF}(t)$, the errors in (\ref{eq_error}) can be converted from the inertial into formation frame as:
\begin{align} \label{eq_error2}
\left[ \begin{array}{c}
e^F_{n,x} \\e^F_{n,y} \\e^F_{n,z}
\end{array} \right]  = R_{FO}(t)\left[ \begin{array}{c} e_{n,x} \\e_{n,y} \\e_{n,z} \end{array} \right]. 
\end{align} \label{indi_path}
The customized path for each UAV can then be represented in terms of trajectory control command as: 
\begin{equation} \label{formation}
{\mathcal T}_{Fn} = {\mathcal T}_F + \Delta {\mathcal T}_n,
\end{equation}
where $\Delta {\mathcal T}_n$ is the required trajectory change for the $n$-th UAV to deviate from the formation centroid. This difference is calculated based on the desired relative distances among the UAVs and the relative position errors in (\ref{eq_error2}), $\Delta {\mathcal T}_n = [e^F_{n,x}, e^F_{n,y}, e^F_{n,z}]^T$. The output ${\mathcal T}_{Fn}$ will be fed to low-level controllers of UAVs for trajectory tracking.

\subsection{Low-level control} \label{LL_Control}
The aim of low-level control is to derive the control laws that apply to each UAV's actuators to reach the desired position and attitude. They are derived from the mathematical model of the quadcopter with two main frames, i.e., the inertial frame $(x_E, y_E, z_E)$ and the body frame $(x_B, y_B, z_B)$. The translational motion of the quadcopter in the inertial frame is determined by its position, $\xi=({x , y , z})^T$, and velocity, $\dot{\xi}=({\dot x , \dot y , \dot z})^T$. The UAV attitude is described by Euler angles $\Theta = ({\phi , \theta , \psi})^T$ with the corresponding roll, pitch, and yaw rates $\dot \Theta =  ({\dot \phi , \dot \theta , \dot \psi})^T$. Let $\omega = [p,q,r]^T$ be the angular rate of the quadcopter in the inertial frame, i.e.,
\begin{equation}\label{Eq1}
\omega = \left [\begin{array}{ccc} 
1 &0 & -s_\theta \\
0 & c_\phi & c_\theta s_\phi \\
0 & -s_\phi & c_\theta c_\phi 
\end{array} \right ]\dot\Theta.
\end{equation}
The transformation from the body to earth frames is then determined by the same rotation matrix as in (\ref{Eq_rotation_matrix2}).

In this study, the position of the UAVs is controlled by the built-in PID controller of the flight controller whereas the attitude is governed by the adaptive twisting sliding mode control. Therefore, only torque components for orientation of the UAV are considered, with the control input being $\tau = u = [u_\phi, u_\theta, u_\psi]^T$, where $u_\phi$, $u_\theta$ and  $u_\psi$ respectively represent the roll, pitch and yaw components of the thrust torque $\tau$.

The quadcopter dynamics can then be described as follows:
\begin{align}\label{Eq7b}
\begin{cases}
\dot{X}_1 = X_2 \\
\dot{X}_2 = I^{-1}\left[f(X) + u + d\right],
\end{cases}
\end{align}
where $X_1 = \Theta$, $X_2 = \dot{\Theta}$, $X=[X_1, X_2]^T$ is the state vector, $d = [d_\phi, d_\theta, d_\psi]^T$ is the disturbance vector, and $f(X)$ is the vector function of the quadcopter moments of inertia $I_{xx}$, $I_{yy}$, $I_{zz}$ of the intertia matrix $I$, and angular rates in the inertial frame \cite{Hoanga:2017}:
\begin{align}\label{Eq8}
f(X) &= \left( \begin{array}{c}
(I_{yy} - I_{zz}) qr\\
(I_{zz} - I_{xx}) pr\\
(I_{xx} - I_{yy}) pq
\end{array} \right).
\end{align}
Of interest at the low level for UAV attitude control are
techniques for improving robustness against nonlinearities, uncertainties, and external
disturbances. By following the sliding mode control methodology, known to possess highly robust performance, the proposed controller has the form: 
\begin{equation}\label{Eq11}
u(t)= u_{eq}(t) + u_T(t),
\end{equation}
where $u_{eq}(t) = (u_{eq,i})^T$ and $u_T(t) = (u_{T,i})^T$, $i = 1, 2,3$, are respectively the equivalent control and the discontinuous part containing switching elements as per desired Euler angle references $X_{1d} = \{\phi_d, \theta_d, \psi_d\}^T$. The sliding function is chosen as:
\begin{equation}\label{Eq10}
\mathbf{\sigma = \dot{e}} +\Lambda \mathbf{e},
\end{equation}
where $\Lambda = \text{diag}(\lambda_\phi, \lambda_\theta, \lambda_\psi)$ is a positive definite matrix to be selected, and $\mathbf{e}$ is the control error, $\mathbf{e} = X_1 -X_{1d}$. 

The equivalent control $u_{eq}$ is obtained at no disturbance by driving $\dot{\sigma}$ to zero, as follows:
\begin{equation}\label{Eq16a}
u_{eq} = I\left(\ddot{X}_{1d}  - \Lambda\dot{e}\right)- f(X).
\end{equation}
Motivated by \cite{Levant:1993}, the discontinuous control here is the twisting controller, $u_{T,i}, ~i=1,2,3$, of the form:
\begin{align}\label{Eq17}
u_{T,i} = \begin{cases} 
-\mu_i \alpha_i \text{sign}(\sigma_i) & \text{if} \enskip \sigma_i\dot{\sigma}_i \leq 0 \\
-\alpha_i \text{sign}(\sigma_i) & \text{if} \enskip \sigma_i\dot{\sigma}_i > 0,
\end{cases}
\end{align}
where $\mu_i<1$ is a fixed positive number and $\alpha_i >0$ is the control gain. By considering the one-stage  accelerated twisting algorithm \cite{Dvir:2015} and the adaptive sliding mode control techniques \cite{Plestan:2010} for improving the control transient and tracking performance, the gain $\alpha_i$ in (\ref{Eq17}) is 
further adjusted by the adaptation law:
\begin{align}\label{Eq24}
\dot{\alpha}_i &= \begin{cases}
\bar{\omega_i} \abs{\sigma_i(\omega,t)} \text{sign}(|\sigma_i(\omega,t)|^{\rho_i}-\epsilon_i) &\text{if} \enskip \alpha_i > \alpha_{m,i}\\
\eta_i &\text{if} \enskip\alpha_i \leq \alpha_{m,i},
\end{cases}
\end{align}
where  $\bar{\omega_i}$, $\epsilon_i$ and $\eta_i$ are positive constants and $\alpha_{m,i}$ is a sufficiently large threshold for adaptation. By assuming the external disturbance $d$ to be bounded, the convergence of the system under the proposed adaptive twisting sliding mode (ATSM) control is obtained in \cite{VanASCC}.

\subsection{Path planning and formation control implementation}
\begin{figure}
	\centering
	\includegraphics[width=8.5cm]{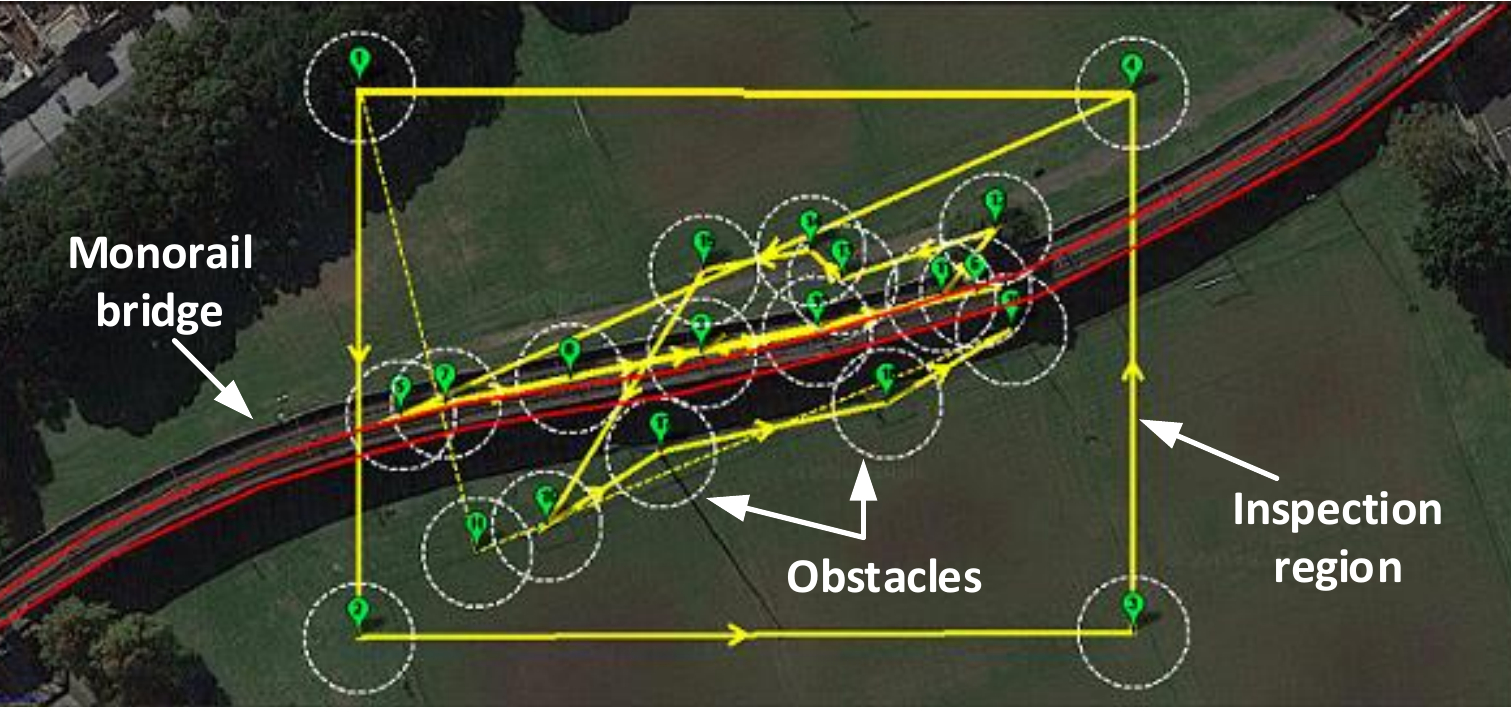}
	\caption{Mission Planner incorporating Google Satellite Map to create initial information and an inspection plan}
	\label{WW_Bridge_plan}
\end{figure}

First, the operation space of UAVs is selected according to the infrastructure surface to be inspected. This can be done by using a navigation map with satellite images and a ground control software called ``Mission Planner''. The software is used to collect initial information about the surface of interest and its surrounding environment based on Google satellite maps (GST). For illustration, Fig. \ref{WW_Bridge_plan} shows the map of a monorail bridge as a testbed subject to inspection loaded by Mission Planner. Correspondingly, the inspection areas, waypoints together with obstacles are identified providing parameters for the cost function and constraints to be defined as in (\ref{cost_function}). Then, the $\theta$-PSO algorithm, with the pseudo-code presented in Fig.\ref{figPSOpseudocode}, is run to obtain the desired path. The path is then uploaded to the UAVs by using Mission Planner for the autonomous flight.

\begin{figure}[h!]
	{\fontsize{08}{08}\selectfont
		\removelatexerror
		\begin{algorithm}[H]
			\SetAlgoLined
			\tcc{Preparation:}
			\quad Determine the inspection surface\;
			\quad Identify the upper and lower boundaries of the working space, start and target positions of the swarm\;
			\quad Identify obstacles in the working space, check and adjust obstacles' parameters if needed \;
			\quad Group all the above data and save in a common file (init file)\;
			\tcc{Initialization:}
			\quad Initialize the working environment by loading the init file to global memory\;
			\quad Initialize the $\theta$-PSO parameters, i.e., $w$, $swarm\_population$, and $swarm\_iteration$ and generate a random path to connect the start and the target points\;
			\quad Set the range of constraints for each particle's phase angle and angular increment in [-$\pi$/2, $\pi$/2]\;
			\tcc{$\theta$-PSO:}
			\ForEach {$i < (swarm\_iteration)$} {
				\ForEach {$ j < (swarm\_population)$} {
					Calculate new phase angle increment value in the range of limitation; \tcc*[f]{using 1st equation in (\ref{EqThePSO})} \\		
					Calculate new phase angle value in the range of limitation; \tcc*[f]{using 2nd equation in (\ref{EqThePSO})} \\					
					Calculate new position; \tcc*[f]{using 3rd equation in (\ref{EqThePSO})} \\		
					Check $Violation$ cost\;
					Evaluate each path based on the $Best\_Costs$ and $Violation$ cost\;
					Update each $particle\_personal\_best$ and the $global\_best$ positions\;
				}
				\quad Update $global\_best$ and $Violation$ costs\;
			}
			\quad Save $global\_best$ and $Violation$ cost\;
			\tcc{Path generation:}
			\quad Final path is chosen as the maximum number of iterations is reached\;
			\quad Generate individual paths for UAVs using (\ref{indi_path})\;
		\end{algorithm}
	}
	\caption{Pseudo-code for path generation process}
	\label{figPSOpseudocode}
\end{figure}

Given the reference path, the formation flight starts with an initialisation phase in which each UAV needs to reach its desired initial position from an arbitrary location without collision. After that, the low-level control is applied to UAVs based on (\ref{formation}) to maintain the shape. During the flight, on-board computers calculate the inverse kinematics, obtain position errors with respect to their neighbours and the formation centroid, and then drive these errors to zero with the tracking control.

\section{Surface inspection} \label{inspection}
For defect detection, images of the inspected surface taken by UAVs are sent to RCU. Due to a large amount of data to be processed in real time, a fast yet effective image processing algorithm is required. Choosing a particular color space in color image segmentation is largely application dependent \cite{Kwok2009}. Here, for the sake of camera-based inspection, a surface patch quite often has typically similar or repeated background patterns. Therefore, processing color images may not be as effective as processing grey scale images for the purpose of detecting potential defects, which are defined as abnormal changes in the gray level of pixels. Histogram-based techniques for conversion from 3$\times$2D of RGB images into 1D thus can be used to speed up the calculation. Since existing methods such as the global thresholding (Otsu) \cite{Bi_Otsu}, valley emphasis (VE) \cite{VE}, adaptive thresholding (Sauvola) \cite{Bi_Sauvola}, iterative analysis (ITTH) \cite{Bi_ITTH} and slope difference distribution (SDD) \cite{SDD} do not yield satisfactory results if required a higher sensitivity to identify a relatively small number of pixels indicating defects, we seek a new algorithm with an optimal threshold to separate defects from the background.

\begin{figure}
	\centering
	\includegraphics[width=8.5cm]{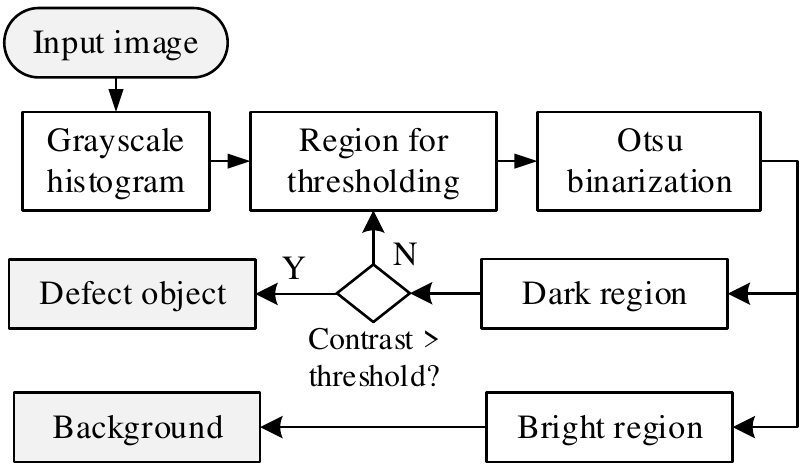}
	\caption{Flowchart of the defect detection algorithm}
	\label{detectionAl}
\end{figure}

The proposed algorithm, called iterative thresholding (IT) is illustrated in Fig.\ref{detectionAl}. It begins with the determination of a threshold value, $T^*$, similar to the Otsu threshold that can maximize the variance between two classes in the histogram. Given a threshold $T$, the variance $\sigma^2(T)$ is calculated as:
\begin{equation}
\sigma^2(T)=\omega_0(T)\omega_1(T)(\epsilon_0(T)-\epsilon_1(T))^2, 
\label{fuc:eq1}	
\end{equation}
where $\omega_0(T)$ and $\omega_1(T)$ are the weight of foreground and background pixels in the whole image, and $\epsilon_0(T)$ and $\epsilon_1(T)$ are the expectations of the foreground and background intensity: 
\begin{equation}
\omega_0(T) = \dfrac{\sum_{x=0}^{T}y(x)}{\sum_{x=0}^{255}y(x)},	
\end{equation}

\begin{equation}
\omega_1(T) = \dfrac{\sum_{x=T+1}^{255}y(x)}{\sum_{x=0}^{255}y(x)},	
\end{equation}

\begin{equation}
\epsilon_0(T) = \sum_{x=0}^{T}xy(x),	
\end{equation}

\begin{equation}
\epsilon_1(T) = \sum_{x=T+1}^{255}xy(x).	
\end{equation}

The threshold $T^*$ is then given by:
\begin{equation} \label{eq:threshold}
T^*=\argmax \limits_{T \in (0,255)} \sigma^2(T).
\end{equation} 
$T^*$ will evenly segment the input image into the dark and bright regions. Without loss of generality, assume that the defect lies in the dark region, our algorithm then continues with an iterative process applying on the dark region.

\begin{figure}[h!]
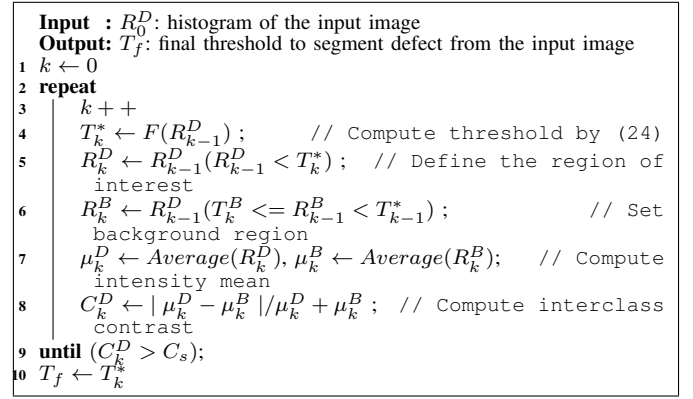

	{\fontsize{08}{08}\selectfont
		\removelatexerror
		\begin{algorithm}[H]
			\SetKwInOut{Input}{Input}
			\SetKwInOut{Output}{Output}	
			\Input{$R^D_0$: histogram of the input image}
			\Output{$T_f$: final threshold to segment defect from the input image}
			
			$k \gets 0$\\
			\Repeat{$(C^D_k>C_s)$}{
				$k++$\\
				$T^*_k \gets F(R^D_{k-1})$ \tcp*{Compute threshold by (\ref{eq:threshold})} 
				$R^D_k \gets R^D_{k-1}(R^D_{k-1}<T^*_k)$ \tcp*{Define the region of interest}
				$R^B_k \gets R^D_{k-1}(T^B_k<=R^B_{k-1}<T^*_{k-1})$ \tcp*{Set background region}
				$\mu^D_k \gets Average(R^D_k)$, $\mu^B_k \gets Average(R^B_k)$\tcp*{Compute intensity mean}
				$C^D_k \gets {\mid\mu^D_k-\mu^B_k\mid}/{\mu^D_k+\mu^B_k}$ \tcp*{Compute interclass contrast}} 
			
			$T_f \gets T^*_k$
		\end{algorithm}
	}
	\caption{Pseudo code for defect detection}
	\label{alg:frontthresh}
\end{figure}

At iteration $k$, the threshold $T_k^*$ in (\ref{eq:threshold}) will segment the dark region $R^D_{k-1}$ into two new regions $R^D_{k}$ and $R^B_{k}$ so that:
\begin{equation}
R^D_{k-1} =  R^D_{k} \cup R^B_{k}, 
\end{equation}
where $R^B_{k}$ is the bright region to be treated as the background. To detect defects through the iteration, we use the concept of interclass contrast, a measure for evaluating the segmentation quality, assuming an average intensity for all pixels inside a class \cite{Evaluation}. For region $R^D_{k-1}$, the interclass contrast $C^D_k$ is calculated as:
\begin{equation}
C^D_k =\frac{\mid\mu^D_k-\mu^B_k\mid}{\mu^D_k+\mu^B_k},
\label{eq:contrast}
\end{equation}
where $\mu^D_k$ and $\mu^B_k$ are the intensity means of $R^D_k$ and $R^B_k$, respectively. As the number of pixels in the thresholding region decreases after each iteration, the denominator in (\ref{eq:contrast}) will decrease leading to the increase of $C^D_k$. A large value of $C^D_k$ thus indicates a sharp difference in the intensity between the segmenting classes, which implies the existence of defect-like objects. Let $C_s$ be the threshold for that occurrence, the pseudo-code of our algorithm is then presented as in Fig. \ref{alg:frontthresh}.

\section{Results} \label{result}
The performance of our proposed system has been evaluated in a number of surface inspection tasks. This section describes the system testbed and experimental results.

\subsection{Experimental setup}
\begin{figure}
	\centering
	\includegraphics[width=8cm]{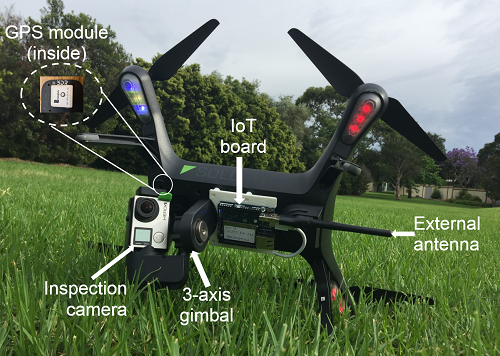}
	\caption{The 3DR Solo testbed}
	\label{drone}
\end{figure}

The UAVs used in this study is the 3DR Solo drone shown in Fig. \ref{drone}. It has three processors, two are Cortex M4 168 MHz running Pixhawk firmware for low-level control and the other is an ARM Cortex A9 running Linux operating system for data processing and high-level control. The UAV is retrofitted with an RTK compatible GPS receiver, an IoT board, a detachable antenna, a camera, a 3D gimbal and other sensors for data acquisition.

The camera used is a Hero 4 Black with the focal length of 34.4 mm, resolution of 12 megapixels, and wireless network capability. It is attached to a three-axis gimbal with one degree of freedom for controlling its elevation (pitch) angle. The photos taken by this camera will be streamed to RCU. The IoT board named Arduino-yun is attached to the accessory port of the drone and interfaced with the embedded Linux operating system via USB protocol. A 6 dBi Ralink RT5370 detachable antenna is attached to the IoT board to extend the wireless communication range. The gateway router is a TP-Link wireless N 4G LTE router that has a SIM card slot to use mobile broadband networks. The RTK compatible GPS receiver is based on the u-blox NEO-M8P-2-10 module. It periodically updates correction data from the base GPS station to improve positioning precision.

The task assigned in experiments is to inspect a monorail bridge using three UAVs. The operation space is of dimension 141 m $\times$ 101 m $\times$ 40 m equivalent to GST coordinates $\{ -33.87601, 151.191182, 0 \}$ and $\{-33.875086, 151.192676, 40\}$, as illustrated in Fig. \ref{WW_Bridge_plan}. The initial and final positions of the formation centroid (6) are $P_{i} = \{40.0,8.0,30\}$ and $P_{f} =\{64,108,34\}$, respectively. Therein, ten obstacles are identified, each with a different radius. 

\begin{figure}
	\centering
	\includegraphics[width=8cm]{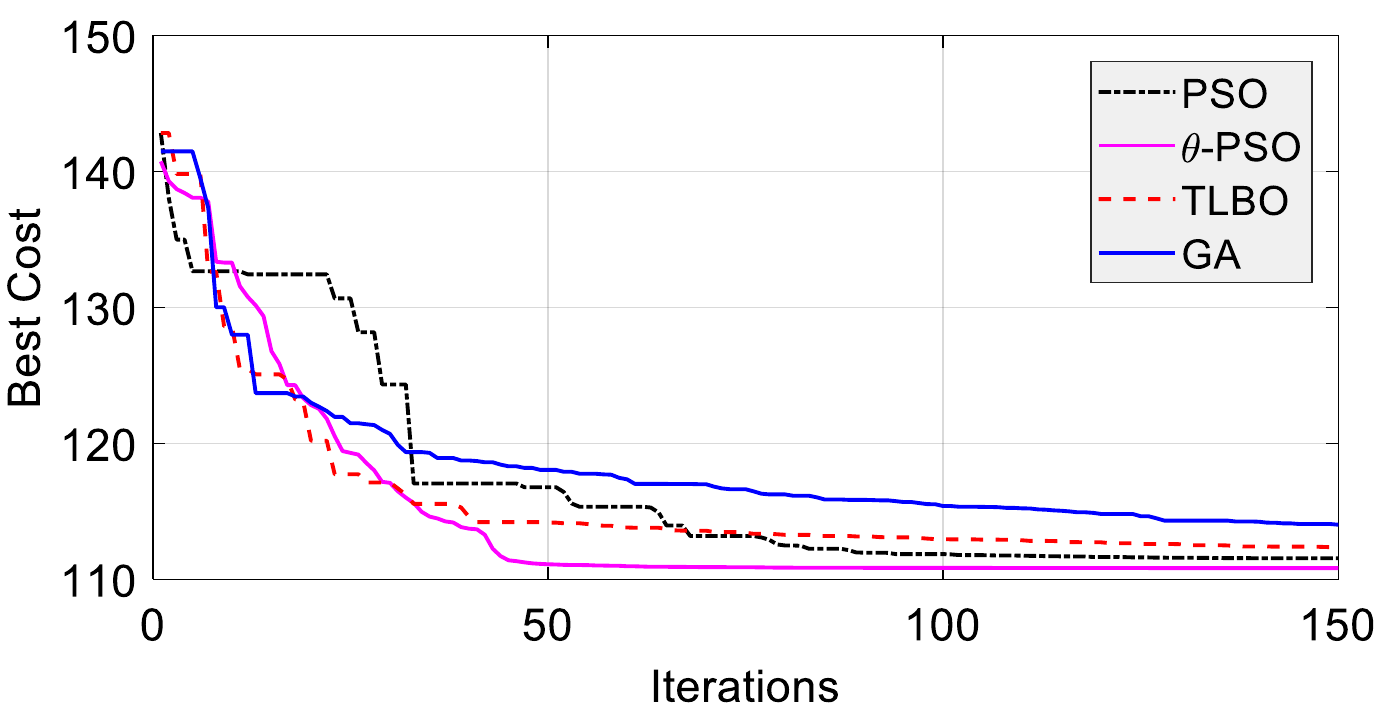}
	\caption{Convergence comparison between conventional PSO and $\theta$-PSO}
	\label{Comparison}
\end{figure}

In our path planning algorithm, the number of particles, waypoints, and iterations are respectively selected as 150, 10, and 300. Parameters of the three quadcopters with respect to the centroid are $\Delta \mathcal{T}_1 = (0, 0, 2)$ m, $\Delta \mathcal{T}_2 = (3, 0, -1)$ m and $\Delta \mathcal{T}_3 = (-3, 0, -1)$ m.

\begin{center}
	\captionof{table}{Performance comparison between GA, TLBO, PSO and $\theta$-PSO} \label{Table2} 
	\begin{tabular}{ |c|c|c|c|}
		\hline
		Algorithm	& Initial cost & Min cost 			& Iterations \\ \hline
		GA 		& 142.91	& 114.17	 	& 147	\\
		TLBO 		& 142.84	& 113.80	 	& 84	\\
		PSO 		& 143.00	& 112.43	 	& 102	\\
		$\theta$-PSO & 142.84	& 111.02		& 68\\
		\hline 
	\end{tabular}
\end{center}

\subsection{High-level control results} \label{PP_results}
The path planning and formation results are presented in this subsection to illustrate the generation of collision-free paths for the three UAVs with sufficiently fast convergence using the proposed algorithm. For this, let us first compare the performance of the proposed $\theta$-PSO with a conventional PSO algorithm and two other bio-inspired algorithms, Genetic Algorithm (GA) \cite{Pehlivanoglu2012} and Teaching-Learning-Based Optimization (TLBO) \cite{RAO2011303}. Figure \ref{Comparison} shows the cost values over iterations, wherein the $\theta$-PSO algorithm exhibits a faster and more stable conversion. The results are confirmed as recorded in Table \ref{Table2} showing the cost values and convergence iterations.

\begin{figure}[!t]
	\begin{subfigure}{\linewidth}
		\centering
		\includegraphics[width=7cm]{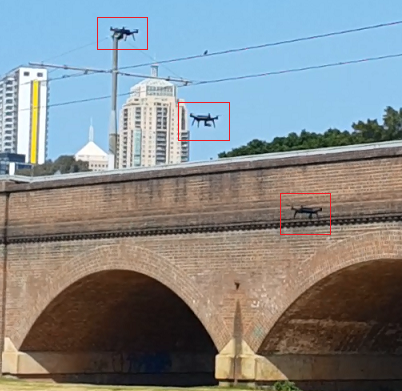}
		\caption{Triangular UAV formation}
	\end{subfigure}\label{Solo_form_fig}
    \hskip2em
	\begin{subfigure}{\linewidth}
		\centering
		\includegraphics[width=7cm]{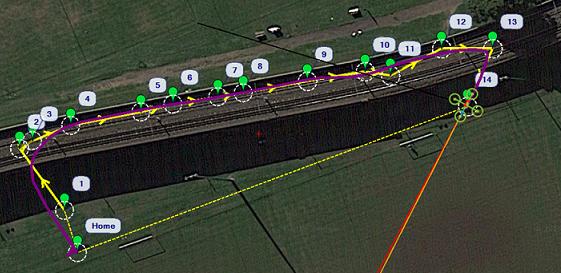}
		\caption{Planned path (yellow) and flown path (violet)}
	\end{subfigure}\label{form_path_fig}
	\caption{Bridge inspection with UAV formation}
		\label{fig6}
\end{figure}

\begin{figure}
	\centering
	\includegraphics[width=8cm]{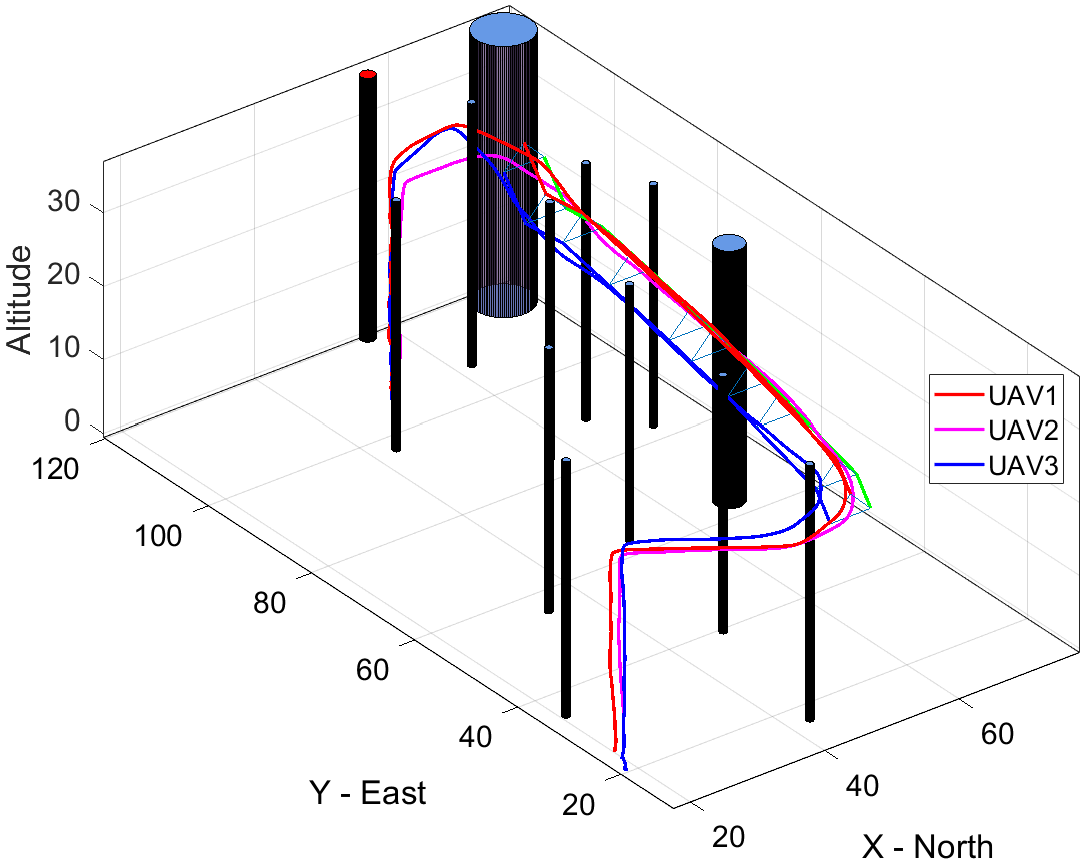}
	\caption{Trajectories of three UAVs tracking the planned paths}
	\label{path_3Solo}
\end{figure}

\begin{figure}
	\centering
	\includegraphics[width=8cm]{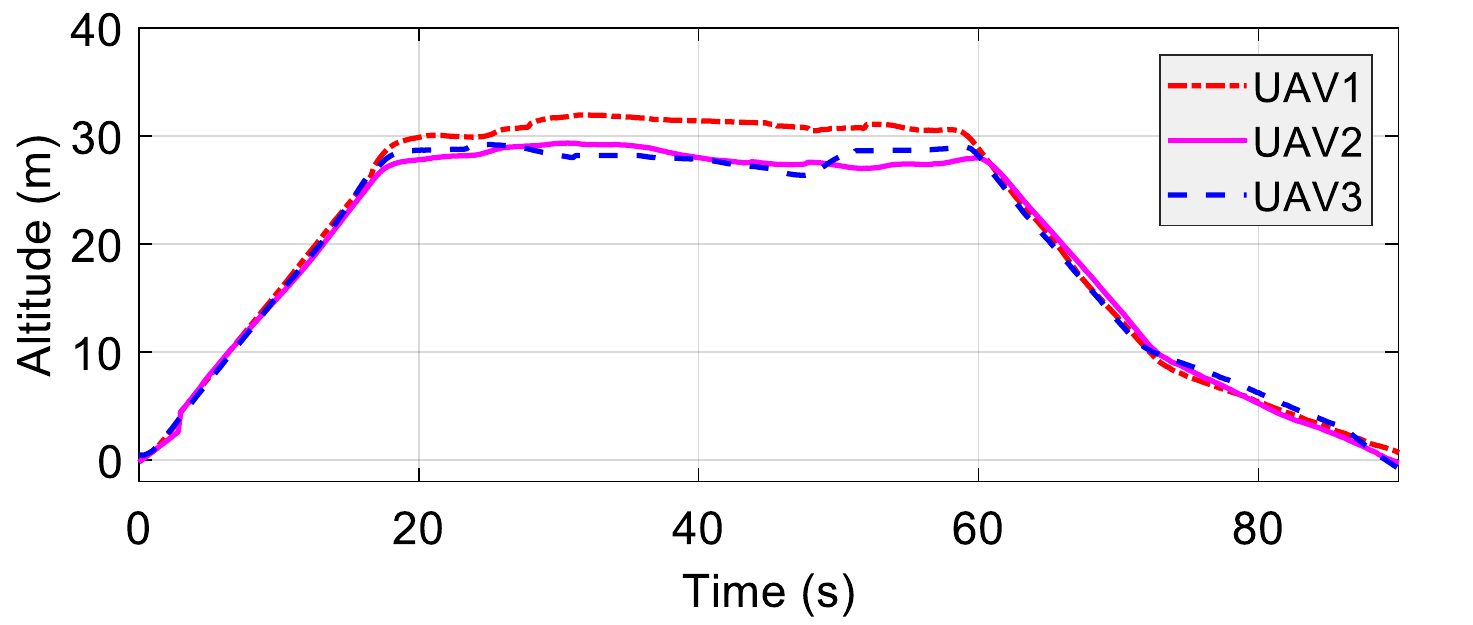}
	\caption{Altitudes of the three UAVs in the formation test}
	\label{Alt_Solo_Formation}
\end{figure}

\begin{figure}
	\centering
	\includegraphics[width=8cm]{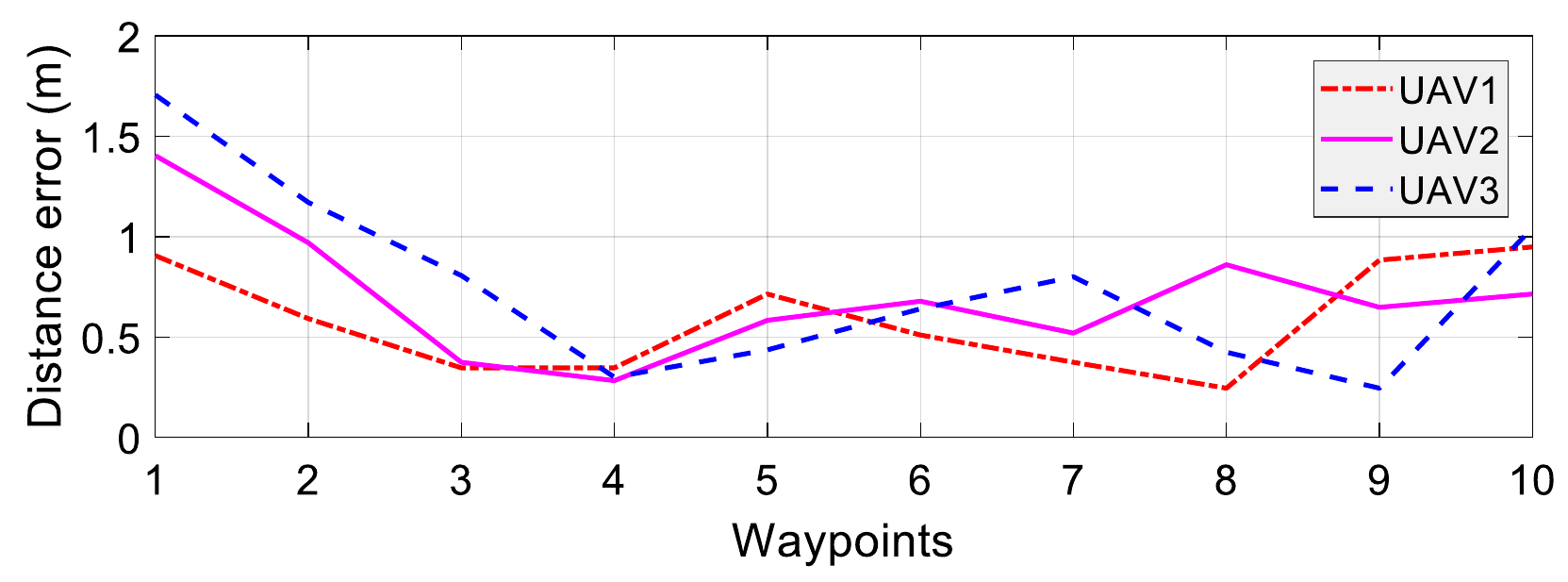}
	\caption{Errors between the planned and flown paths}
	\label{Path_error}
\end{figure}

Field tests were conducted with the triangular formation automatically navigating along the inspected surface, as depicted in Fig. \ref{fig6}. The 3D trajectories generated from the formation centroid path are shown in Fig. \ref{path_3Solo}, illustrating the three UAVs in taking off, reaching to their individual altitude set-point, descending and finally arriving their target position at almost the same interval while also maintaining the desired triangular shape. This result can be further verified via the altitude time responses of the three UAVs as recorded in Fig. \ref{Alt_Solo_Formation}. It is clear that the UAVs are capable of avoiding obstacles and preserving the desired formation configuration during the inspection task. For further evaluation, Fig. \ref{Path_error} shows the error between the planned and flying paths to indicate the feasibility and reasonable smoothness of the generated path for the deployment of UAV formation.

\subsection{Low-level control results} \label{LL_results}
\begin{figure}
	\vspace{-2mm}
	\centering
	\includegraphics[width=8.5cm]{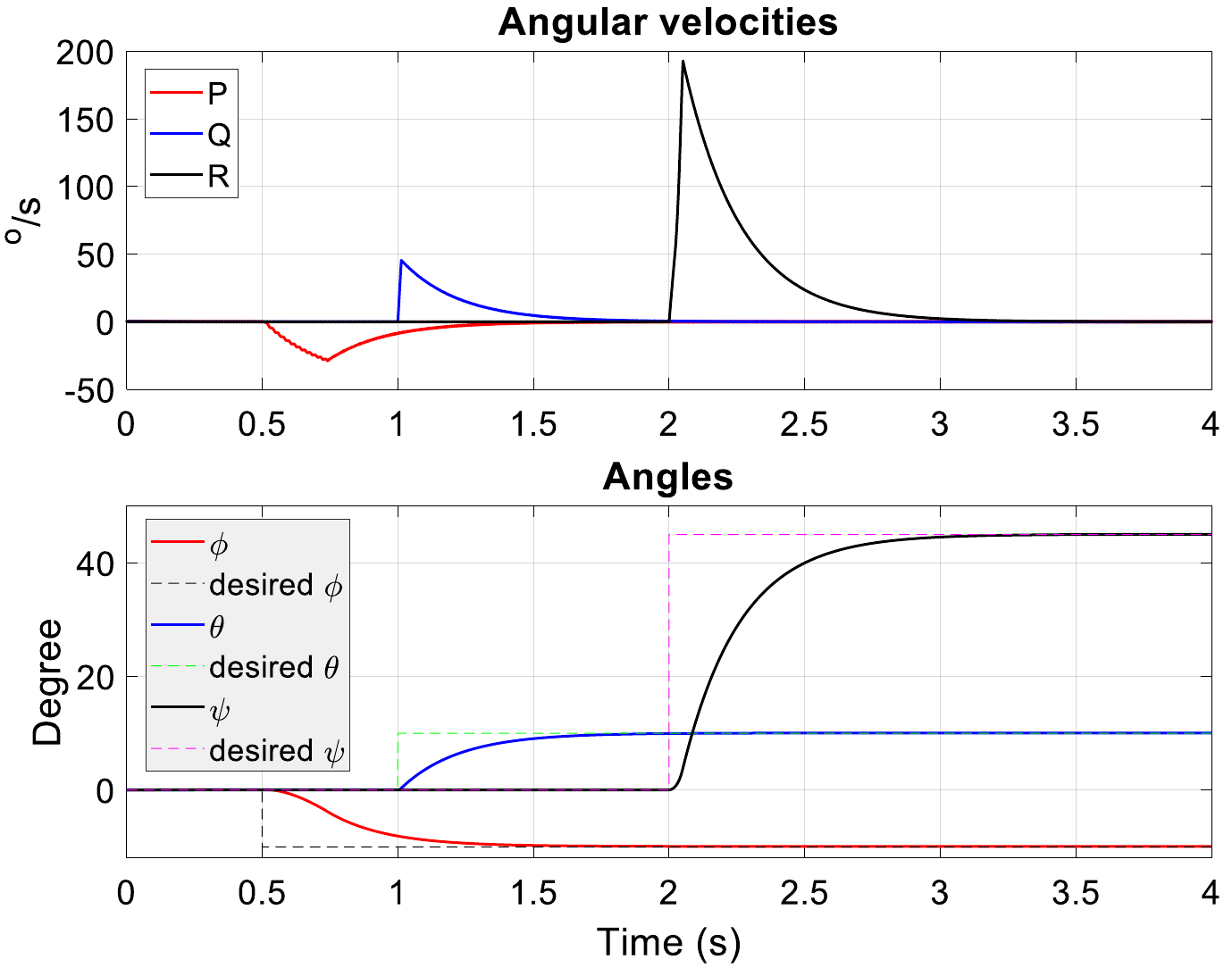}
	\caption{Angular velocity and angle responses in the presence of disturbances}
	\label{dist}
	\vspace{-2mm}
\end{figure}

Control performance of the controller can be judged under disturbances and parametric variations. In our experiments, the UAV was initially staying at the steady state in the air, where all angles and angular velocities are zeros. To test with the real inspection situation, it some sudden and significant changes in reference values are applied as $\phi = -10^\circ$, $\theta = 10^\circ$ and $\psi = 45^\circ$ at time 0.5 s, 1 s and 2 s, respectively. Simulation results in Fig. \ref{dist} illustrate that ATSM effectively rejects disturbances by driving the three angles to their reference values within 2 seconds. Further comparison and evaluation can be found in \cite{VanASCC}.

\subsection{Surface inspection results}
The effectiveness of our proposed defect detection algorithm has been tested on images taken by the camera mounted on UAVs. The results are compared with other methods including Otsu, VE, ITTH, Sauvola and SDD as shown in Figs. \ref{fig:ImCrack1} and \ref{fig:ImCrack2} for various surface defects. The input images were taken at different light conditions with various levels of background complexity, i.e., uniform background (Image 1 and 2), patterned background (Image 3), noisy to very noisy background (Image 4 - 10).  It can be seen that our algorithm outperforms other methods in all test cases regardless of background complexity or image contrast. Specifically, Otsu and ITTH return a high level of noise on all test images except Image 1 where the contrast is high and the background is uniform. Sauvola and VE return a proper segmentation on Images 1-3 where the contrast between the object and the background is rather large, but fail to extract cracks on other images. Similarly, SDD could not detect defects in Images 5-8 as it treats bigger classes more favorably.

\begin{table*}[]
	\renewcommand{\arraystretch}{1.3}
	\footnotesize\addtolength{\tabcolsep}{-3pt}
	\begin{center}
		\begin{tabular}{p{1.2cm}cccc}
			Image 1&\begin{subfigure}{0.15\textwidth}\centering\includegraphics[width=\linewidth]{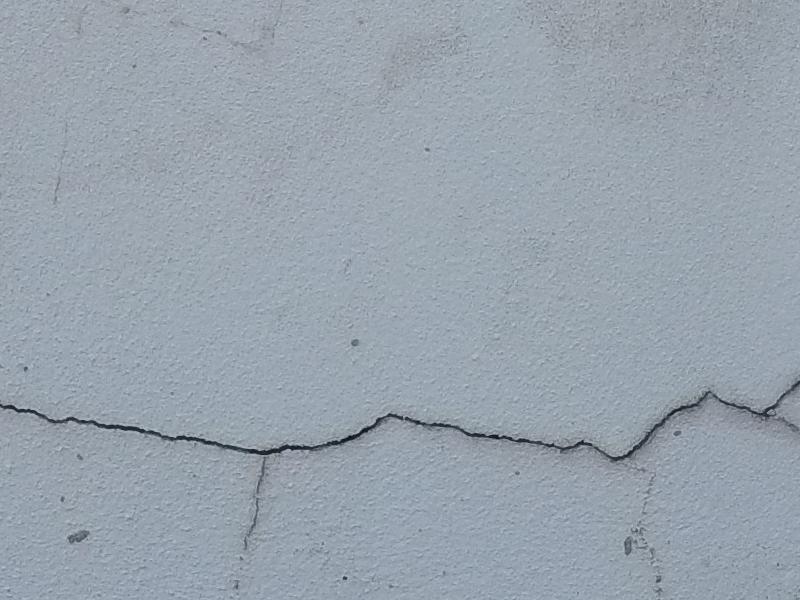}\label{fig:taba}\end{subfigure}&
			\begin{subfigure}{0.15\textwidth}\centering\includegraphics[width=\linewidth]{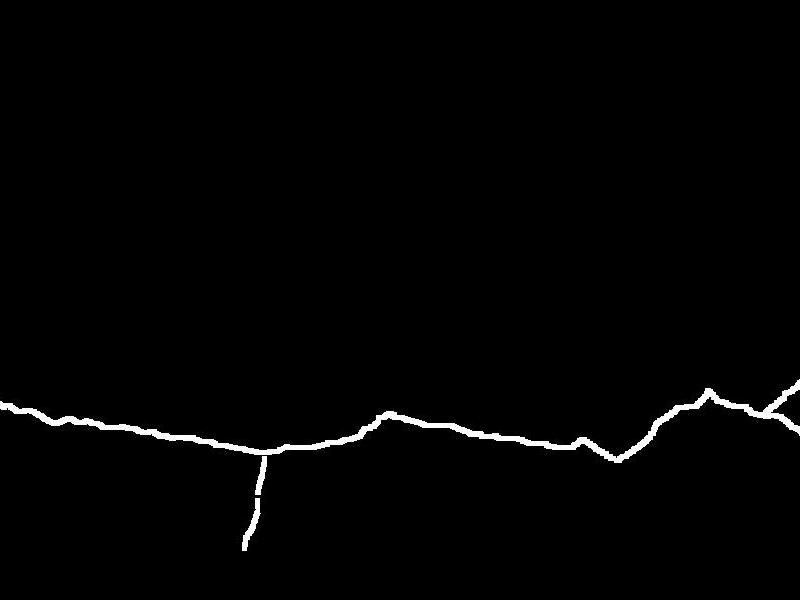}\label{fig:tabc}\end{subfigure}&
			\begin{subfigure}{0.15\textwidth}\centering\includegraphics[width=\linewidth]{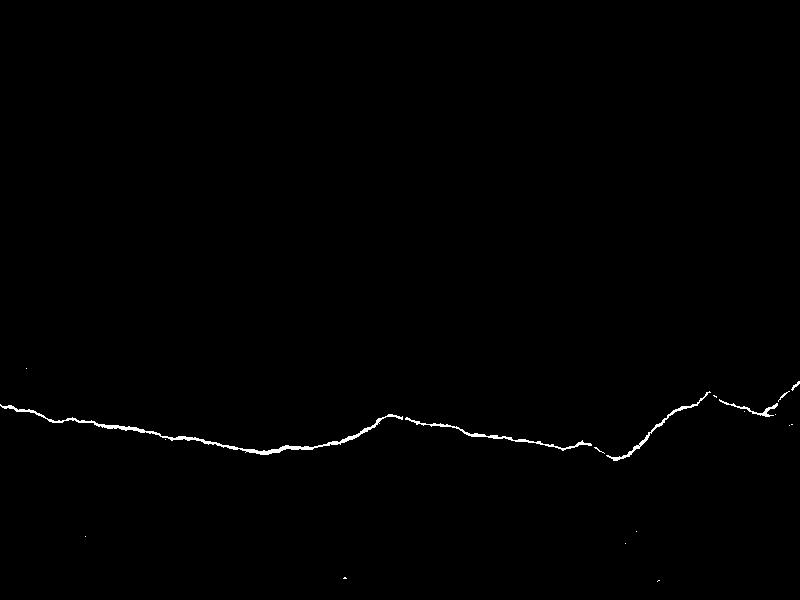}\label{fig:tabc}\end{subfigure}&
			\begin{subfigure}{0.15\textwidth}\centering\includegraphics[width=\linewidth]{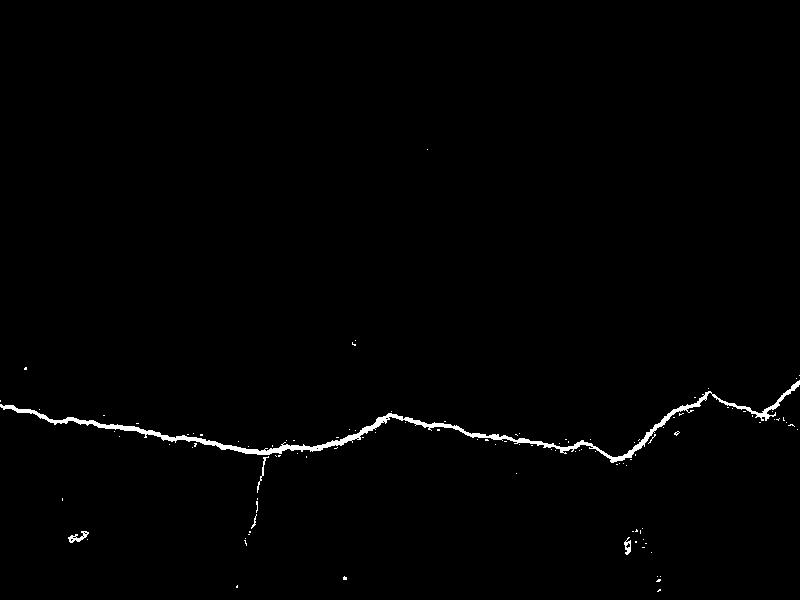}\label{fig:tabc}\end{subfigure}\\[7ex]
			&\begin{subfigure}{0.15\textwidth}\centering\includegraphics[width=\linewidth]{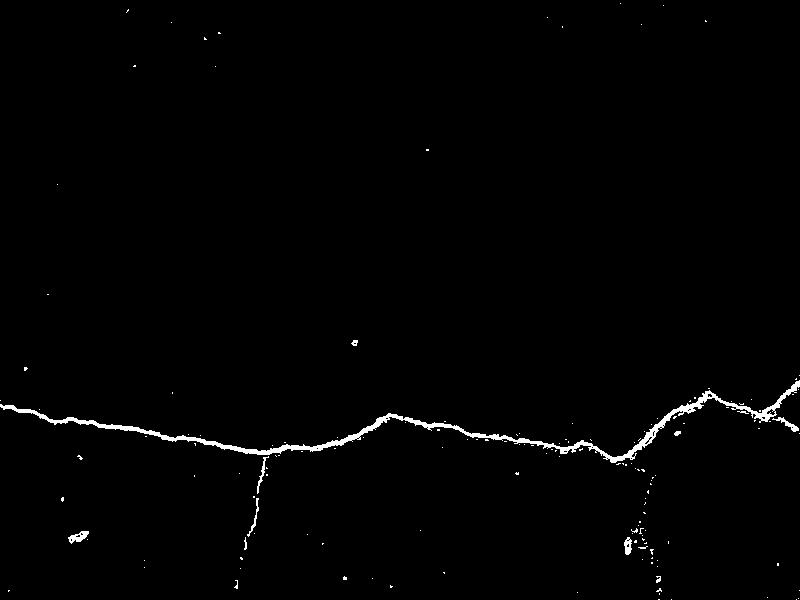}\label{fig:taba}\end{subfigure}&
			\begin{subfigure}{0.15\textwidth}\centering\includegraphics[width=\linewidth]{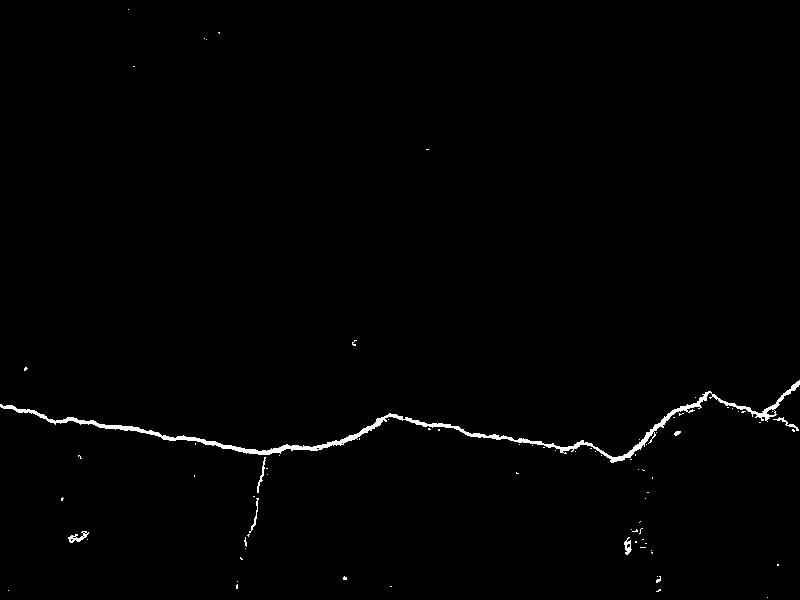}\label{fig:tabc}\end{subfigure}&
			\begin{subfigure}{0.15\textwidth}\centering\includegraphics[width=\linewidth]{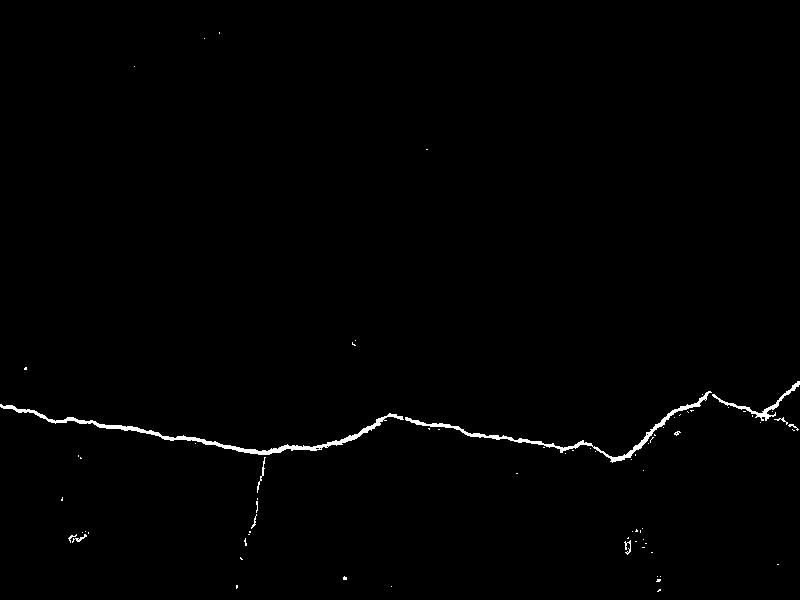}\label{fig:tabc}\end{subfigure}&
			\begin{subfigure}{0.15\textwidth}\centering\includegraphics[width=\linewidth]{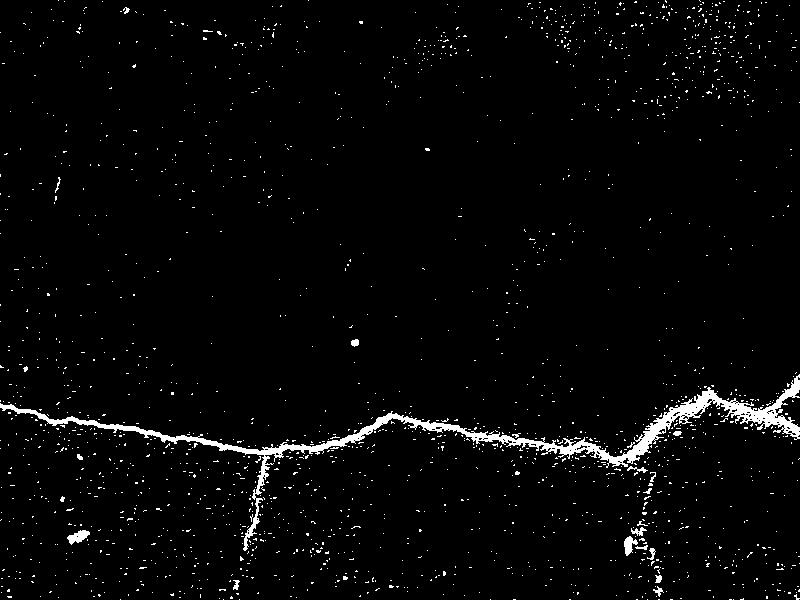}\label{fig:tabc}\end{subfigure}\\[8ex]
			Image 2&\begin{subfigure}{0.15\textwidth}\centering\includegraphics[width=\linewidth]{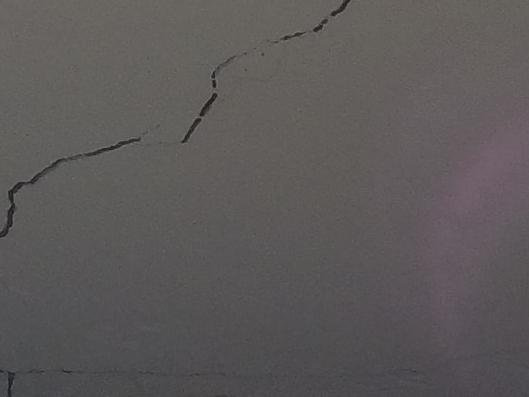}\label{fig:taba}\end{subfigure}&
			\begin{subfigure}{0.15\textwidth}\centering\includegraphics[width=\linewidth]{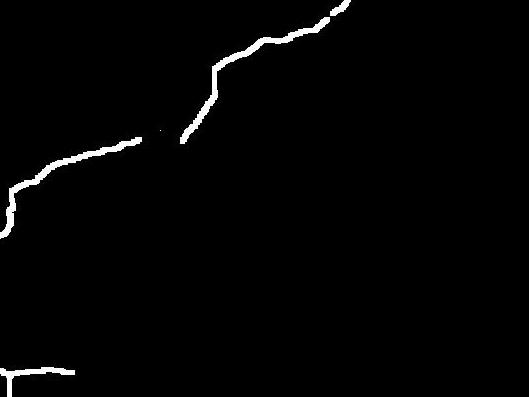}\label{fig:tabc}\end{subfigure}&
			\begin{subfigure}{0.15\textwidth}\centering\includegraphics[width=\linewidth]{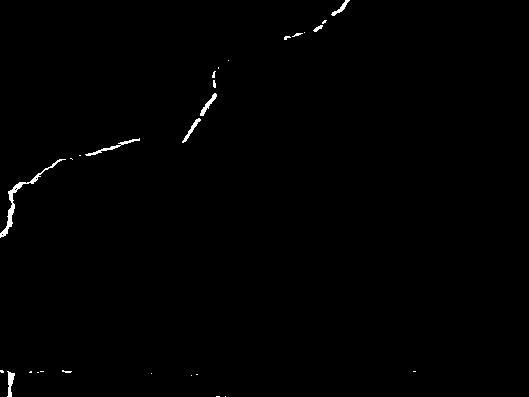}\label{fig:tabc}\end{subfigure}&
			\begin{subfigure}{0.15\textwidth}\centering\includegraphics[width=\linewidth]{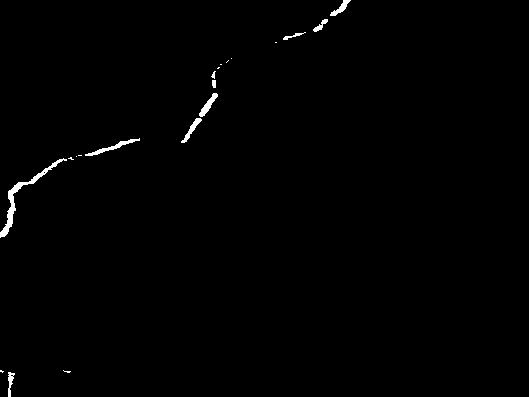}\label{fig:tabc}\end{subfigure}\\[7ex]
			&\begin{subfigure}{0.15\textwidth}\centering\includegraphics[width=\linewidth]{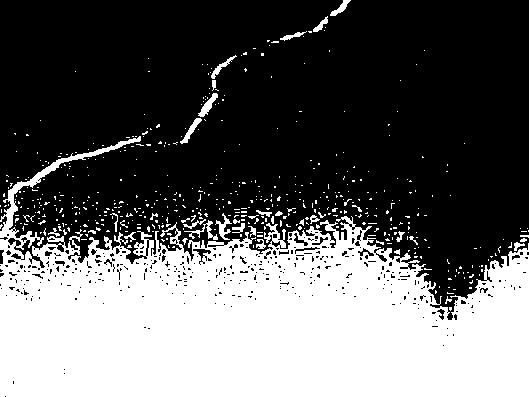}\label{fig:taba}\end{subfigure}&
			\begin{subfigure}{0.15\textwidth}\centering\includegraphics[width=\linewidth]{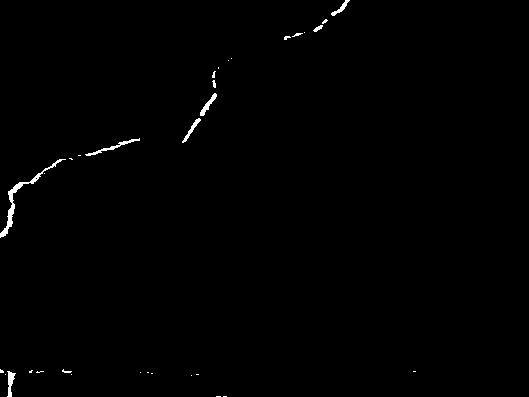}\label{fig:tabc}\end{subfigure}&
			\begin{subfigure}{0.15\textwidth}\centering\includegraphics[width=\linewidth]{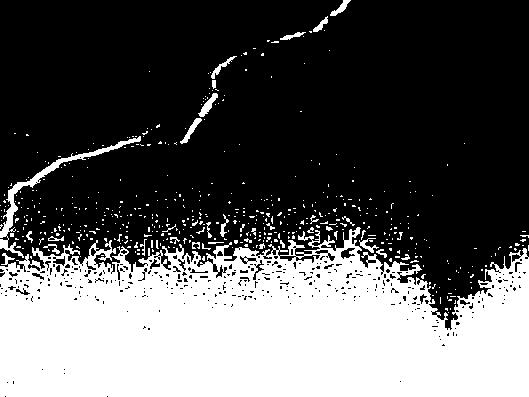}\label{fig:tabc}\end{subfigure}&
			\begin{subfigure}{0.15\textwidth}\centering\includegraphics[width=\linewidth]{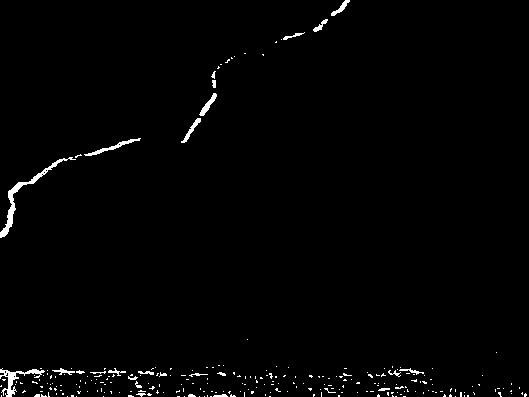}\label{fig:tabc}\end{subfigure}\\[8ex]
			Image 3&\begin{subfigure}{0.15\textwidth}\centering\includegraphics[width=\linewidth]{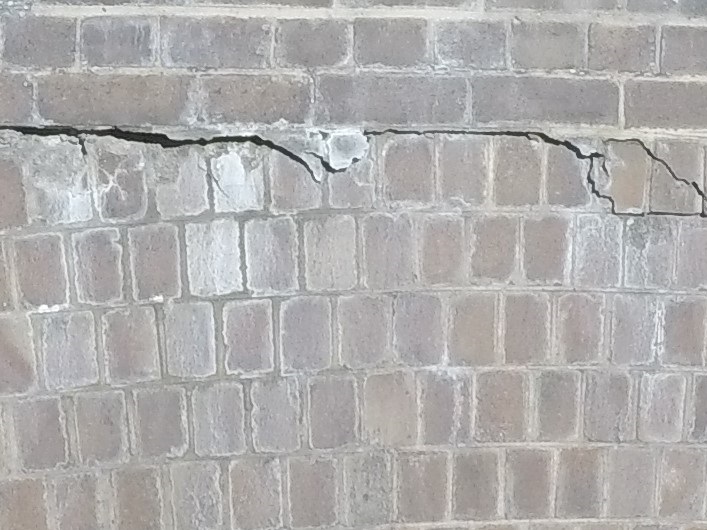}\label{fig:taba}\end{subfigure}&
			\begin{subfigure}{0.15\textwidth}\centering\includegraphics[width=\linewidth]{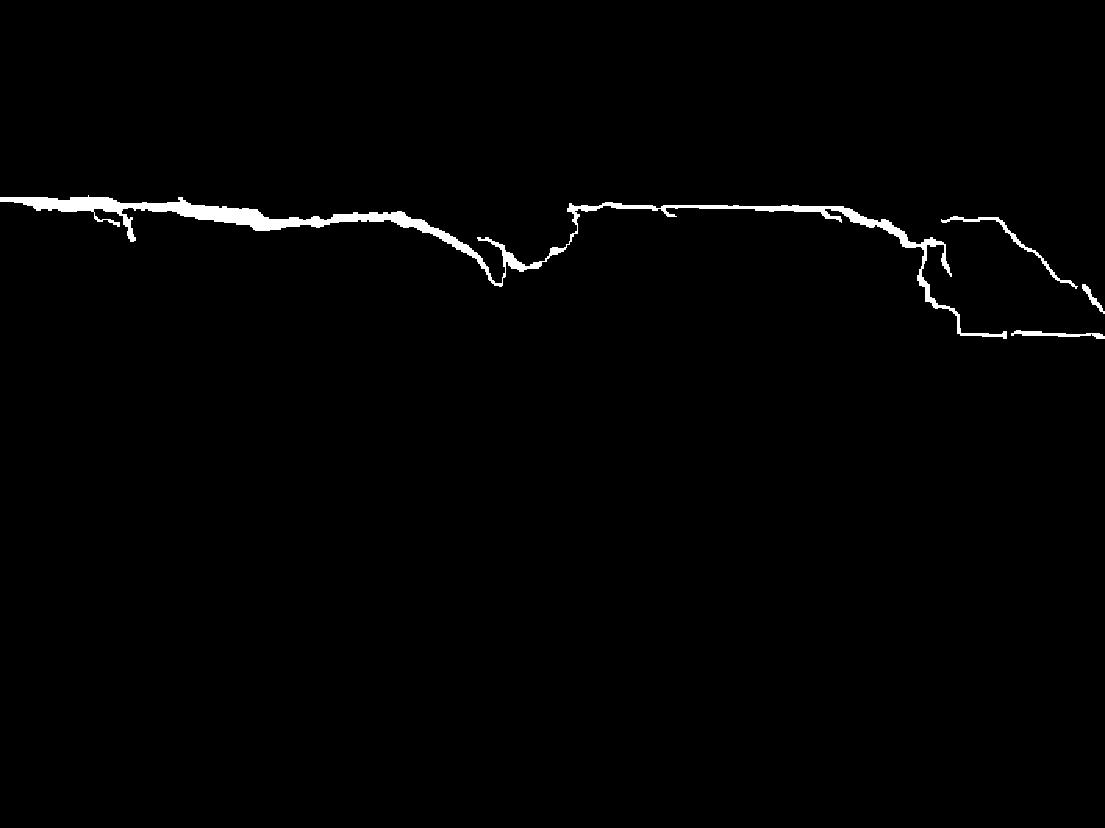}\label{fig:tabc}\end{subfigure}&
			\begin{subfigure}{0.15\textwidth}\centering\includegraphics[width=\linewidth]{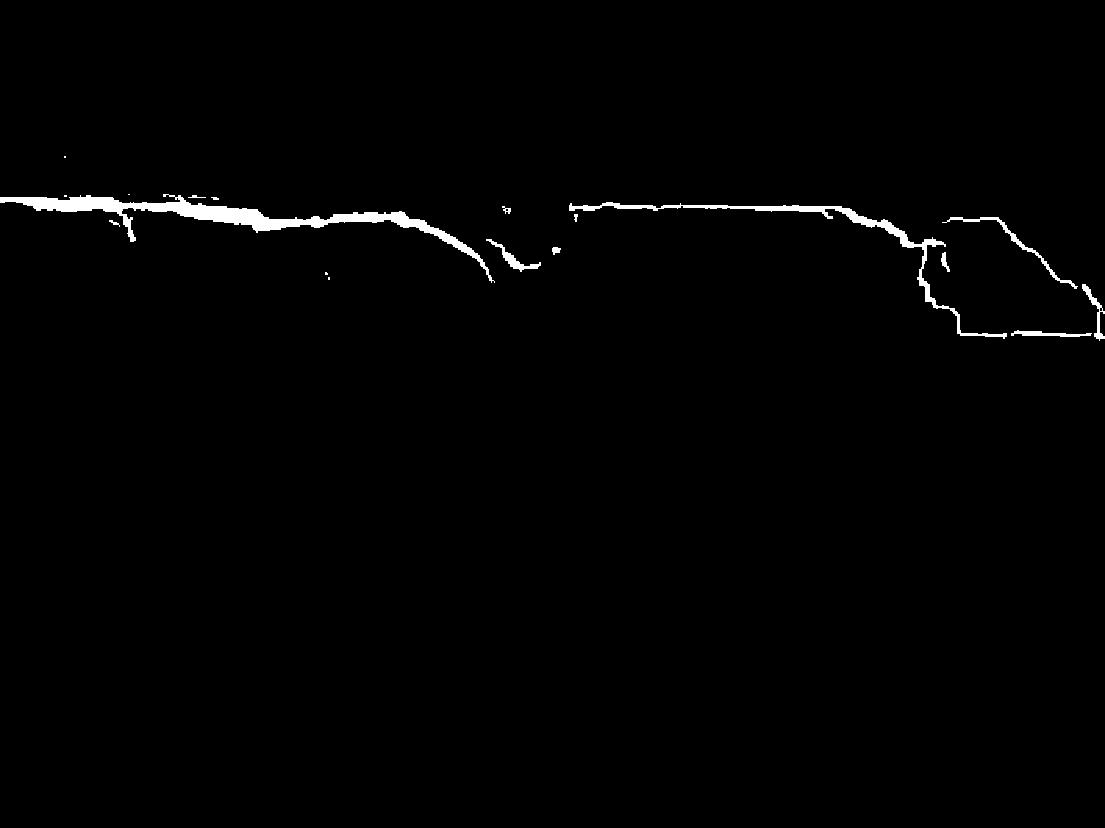}\label{fig:tabc}\end{subfigure}&
			\begin{subfigure}{0.15\textwidth}\centering\includegraphics[width=\linewidth]{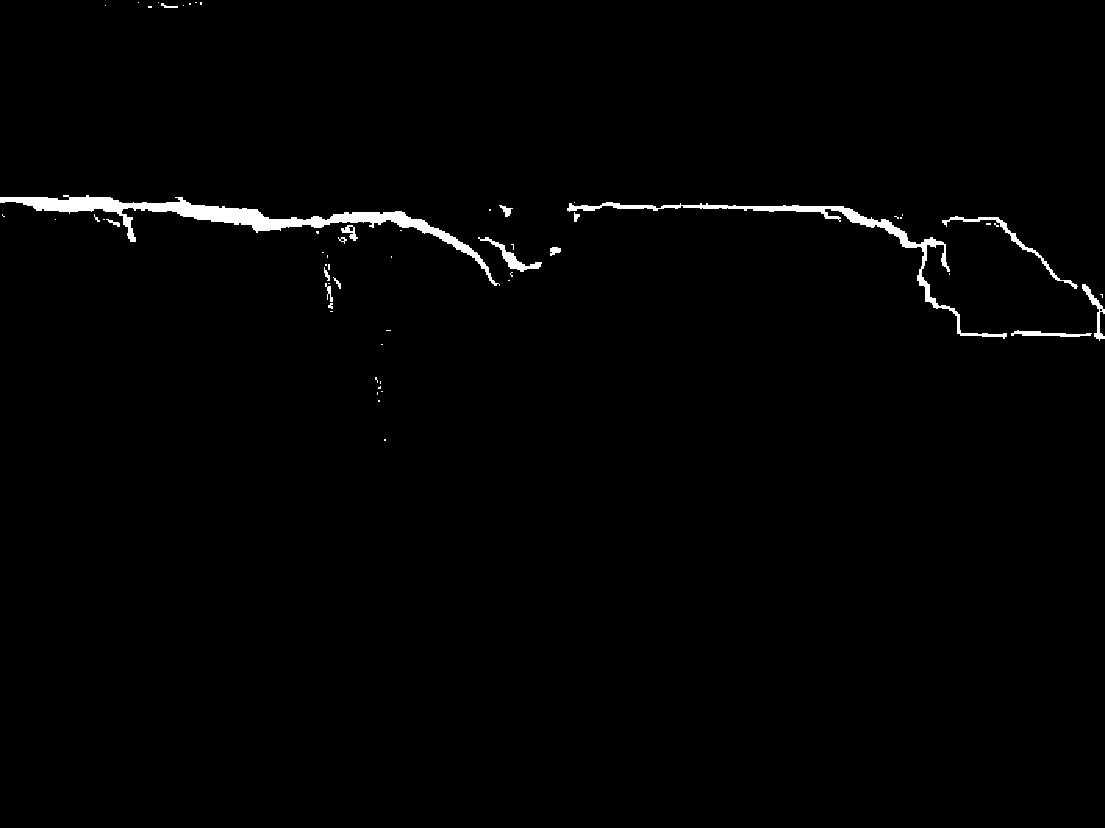}\label{fig:tabc}\end{subfigure}\\[7ex]
			&\begin{subfigure}{0.15\textwidth}\centering\includegraphics[width=\linewidth]{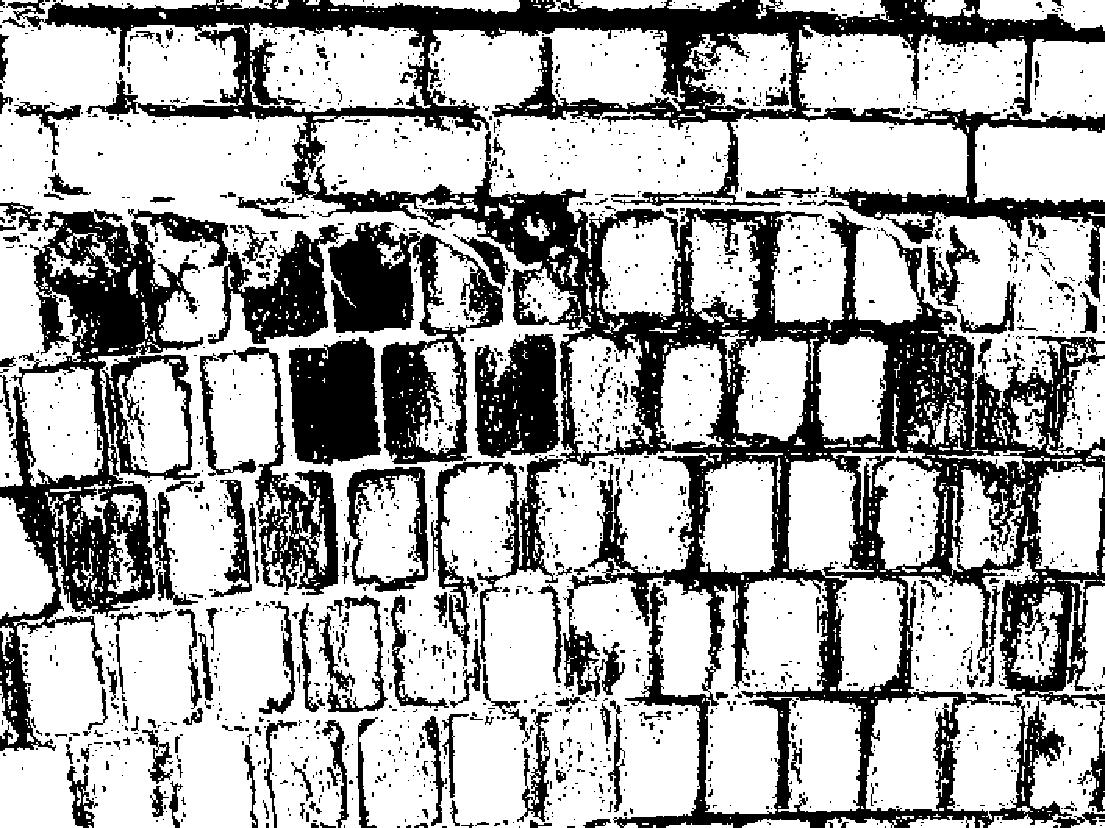}\label{fig:taba}\end{subfigure}&
			\begin{subfigure}{0.15\textwidth}\centering\includegraphics[width=\linewidth]{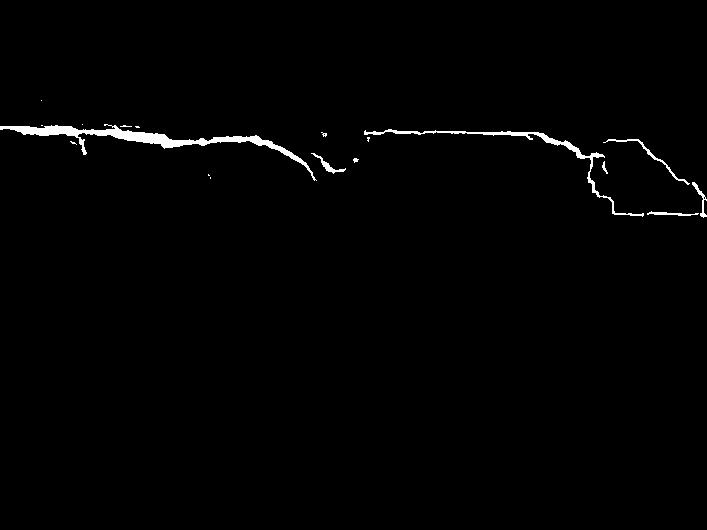}\label{fig:tabc}\end{subfigure}&
			\begin{subfigure}{0.15\textwidth}\centering\includegraphics[width=\linewidth]{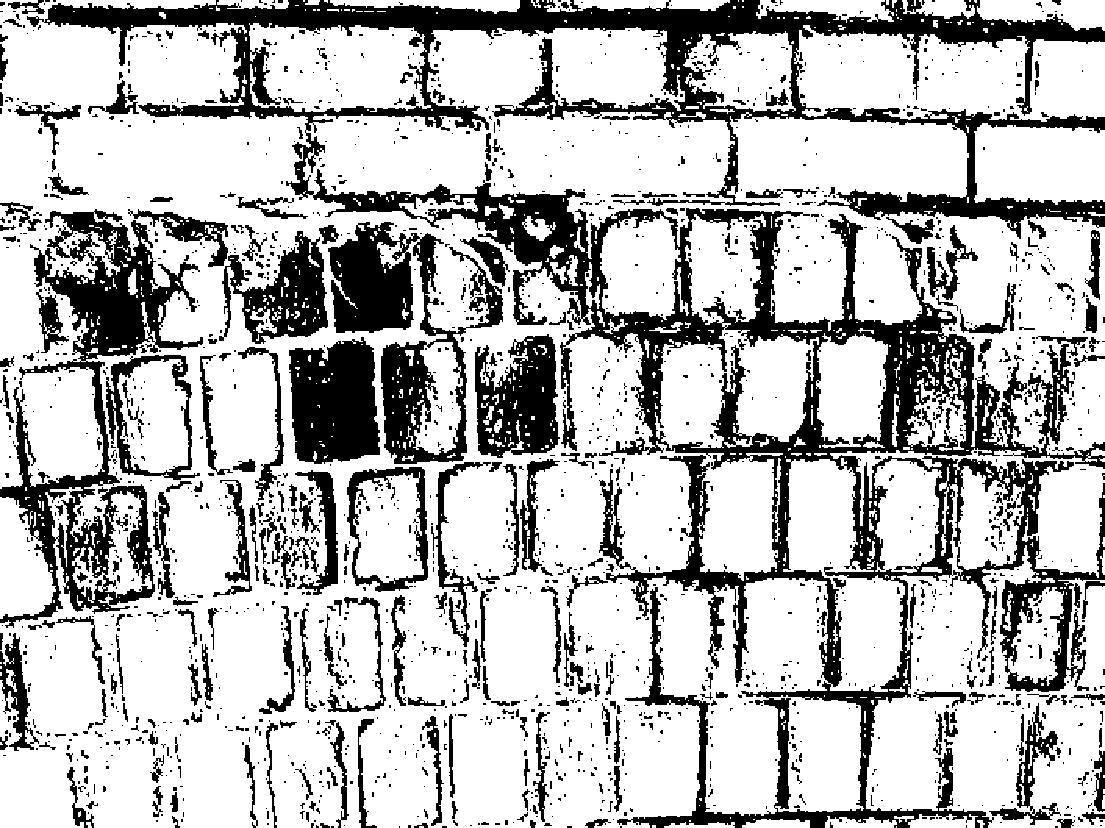}\label{fig:tabc}\end{subfigure}&
			\begin{subfigure}{0.15\textwidth}\centering\includegraphics[width=\linewidth]{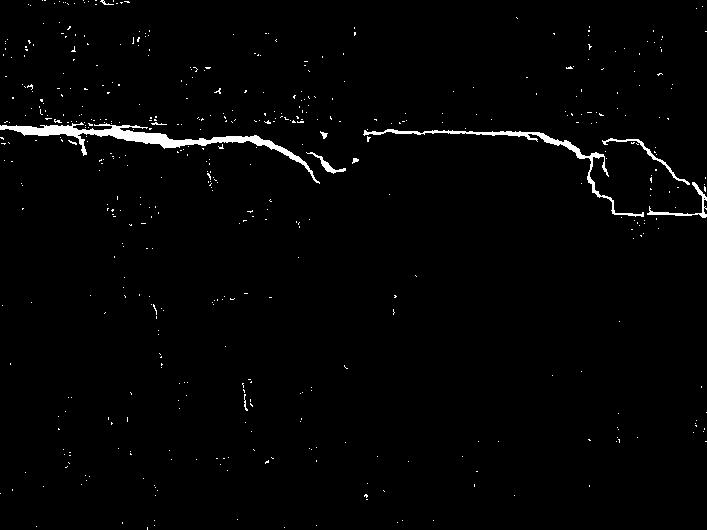}\label{fig:tabc}\end{subfigure}\\[8ex]
			Image 4&\begin{subfigure}{0.15\textwidth}\centering\includegraphics[width=\linewidth]{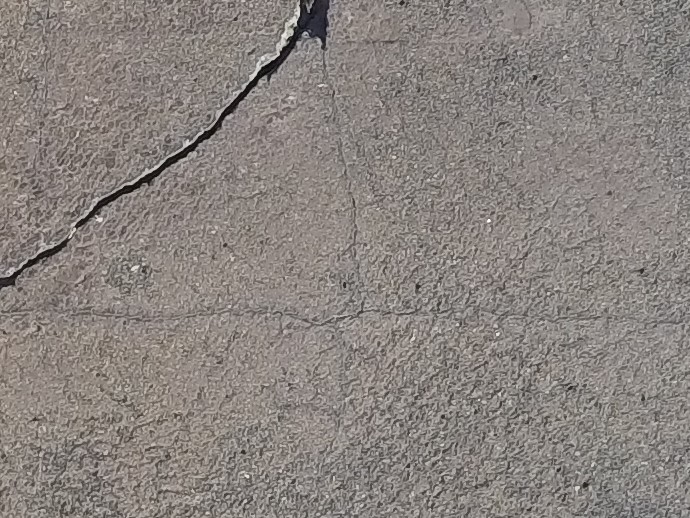}\label{fig:taba}\end{subfigure}&
			\begin{subfigure}{0.15\textwidth}\centering\includegraphics[width=\linewidth]{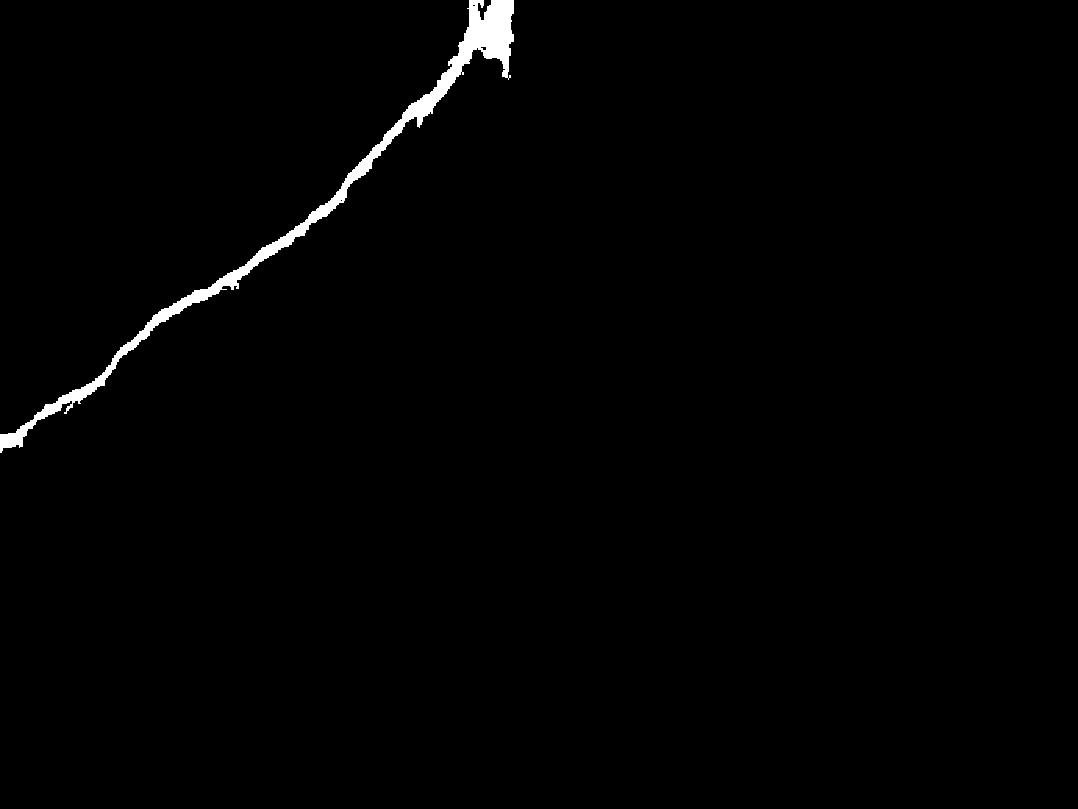}\label{fig:tabc}\end{subfigure}&
			\begin{subfigure}{0.15\textwidth}\centering\includegraphics[width=\linewidth]{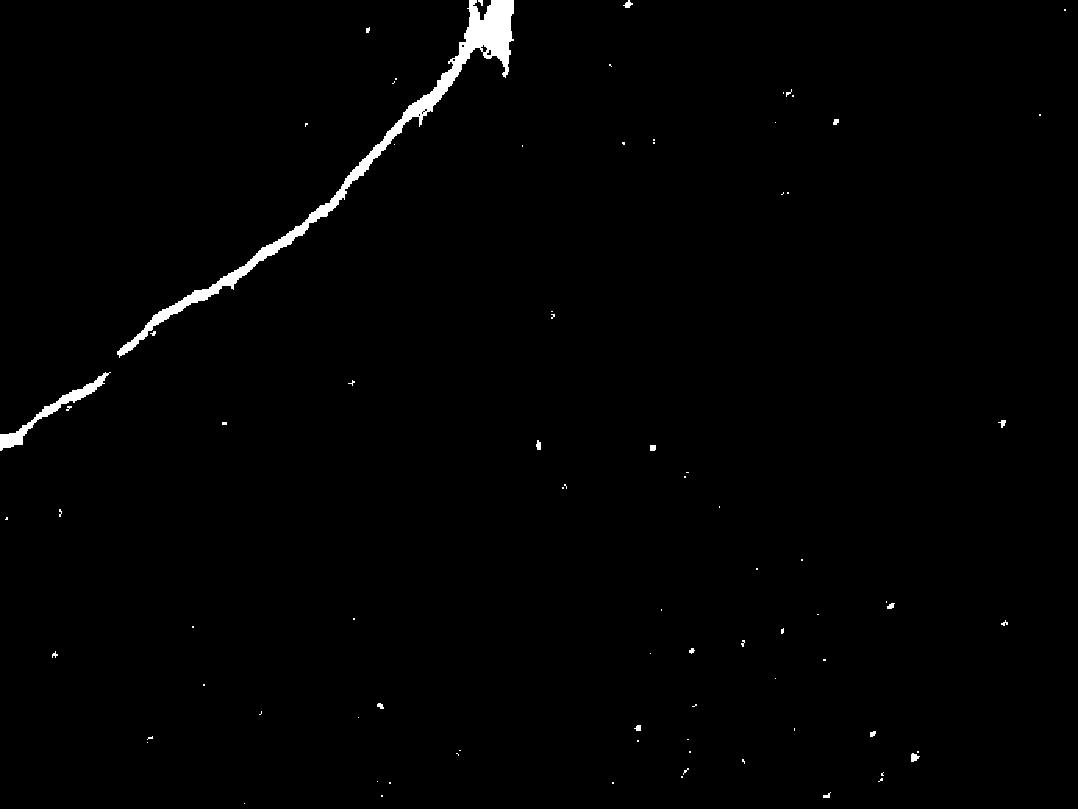}\label{fig:tabc}\end{subfigure}&
			\begin{subfigure}{0.15\textwidth}\centering\includegraphics[width=\linewidth]{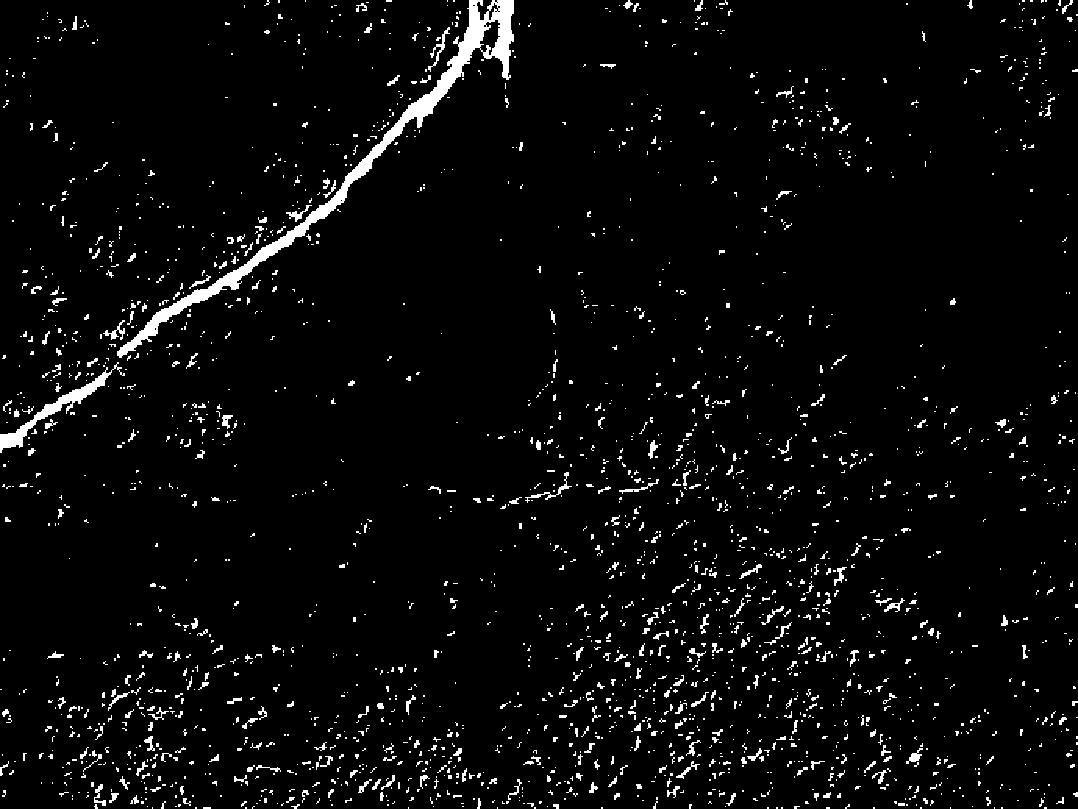}\label{fig:tabc}\end{subfigure}\\[7ex]
			&\begin{subfigure}{0.15\textwidth}\centering\includegraphics[width=\linewidth]{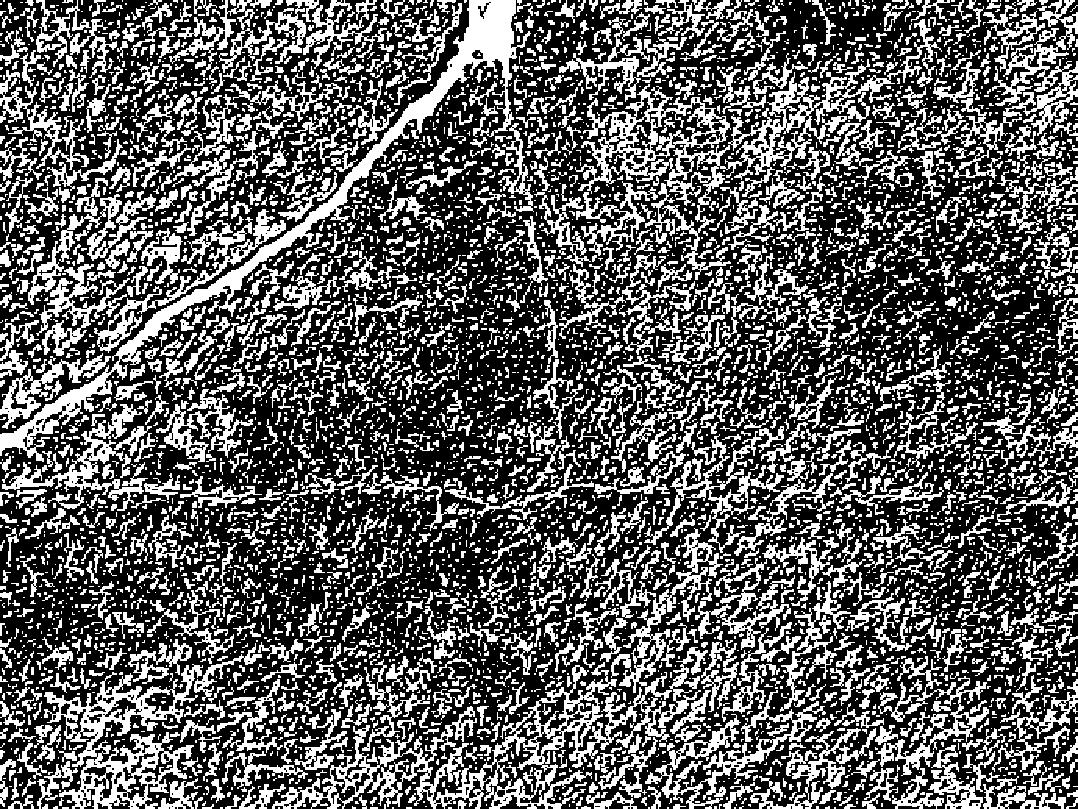}\label{fig:taba}\end{subfigure}&
			\begin{subfigure}{0.15\textwidth}\centering\includegraphics[width=\linewidth]{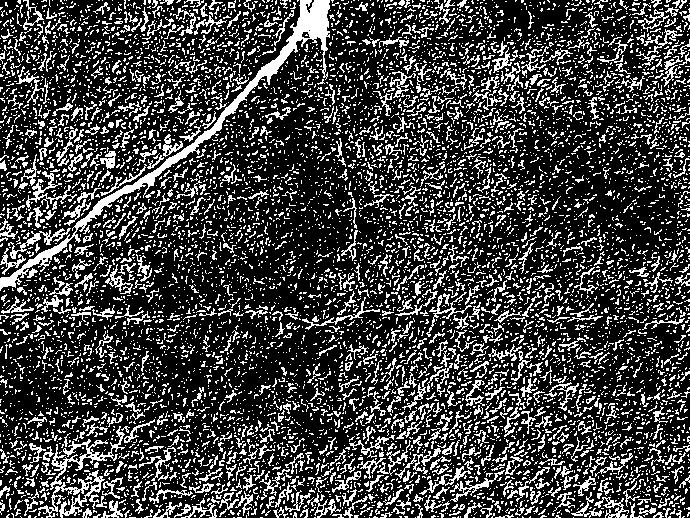}\label{fig:tabc}\end{subfigure}&
			\begin{subfigure}{0.15\textwidth}\centering\includegraphics[width=\linewidth]{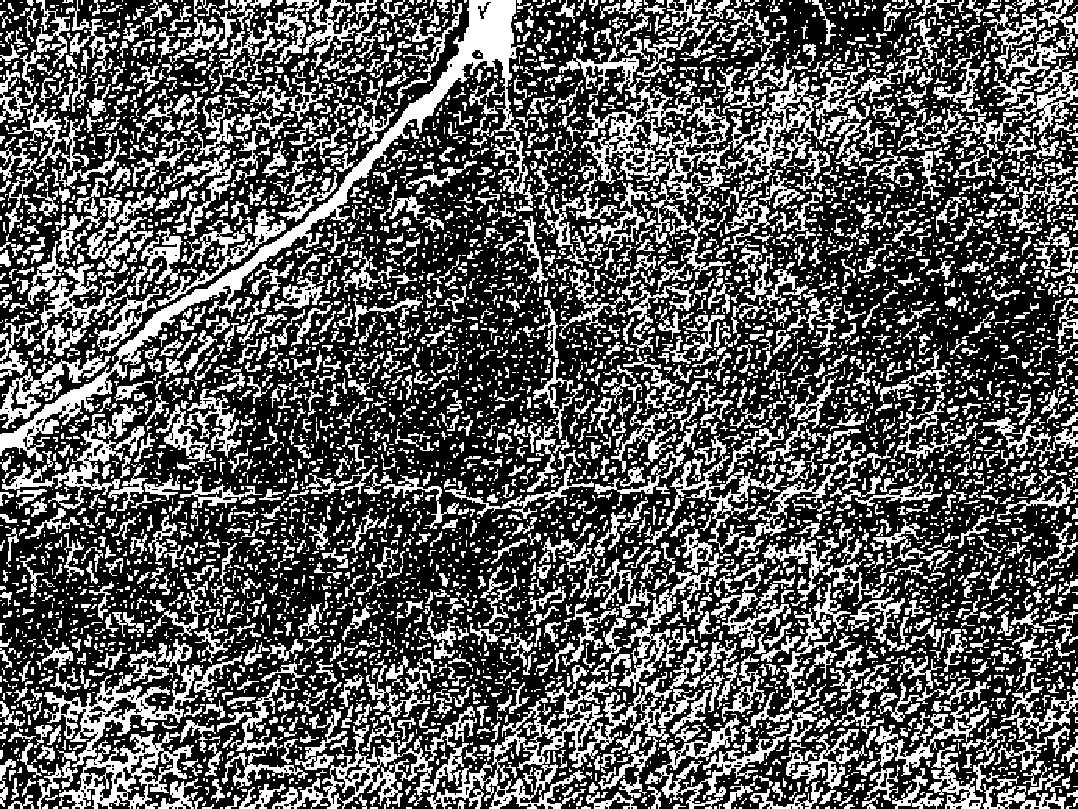}\label{fig:tabc}\end{subfigure}&
			\begin{subfigure}{0.15\textwidth}\centering\includegraphics[width=\linewidth]{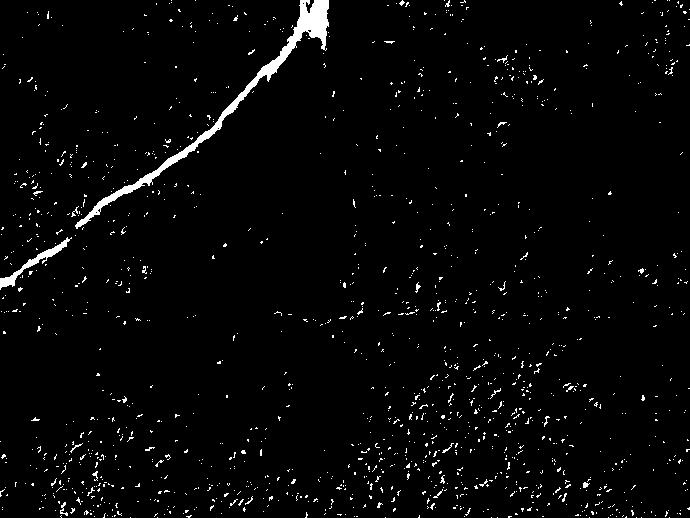}\label{fig:tabc}\end{subfigure}\\[8ex]
			Image 5&\begin{subfigure}{0.15\textwidth}\centering\includegraphics[width=\linewidth]{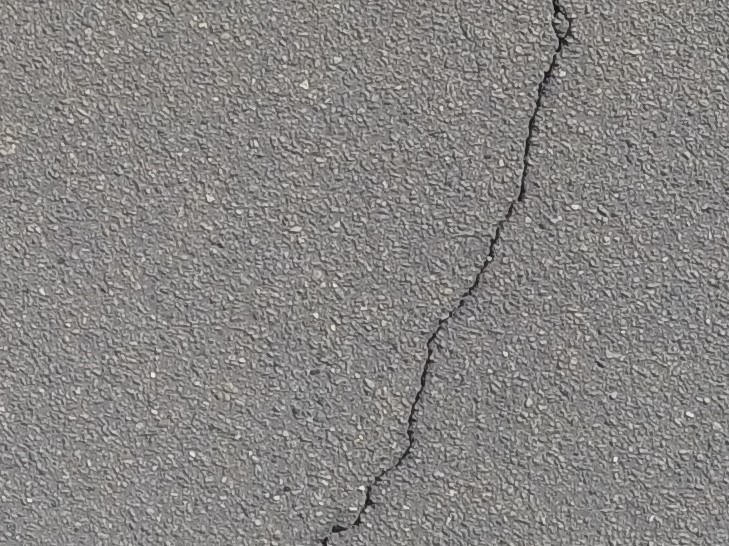}\label{fig:taba}\end{subfigure}&
			\begin{subfigure}{0.15\textwidth}\centering\includegraphics[width=\linewidth]{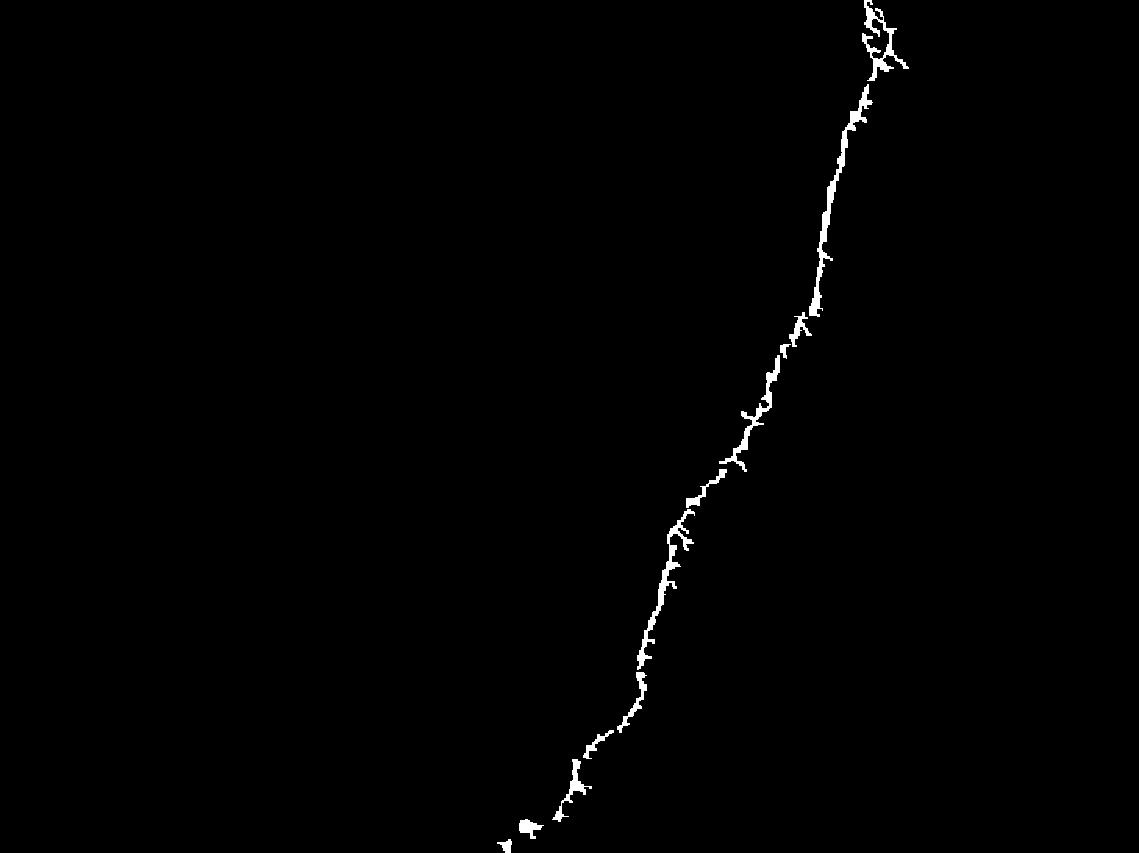}\label{fig:tabc}\end{subfigure}&
			\begin{subfigure}{0.15\textwidth}\centering\includegraphics[width=\linewidth]{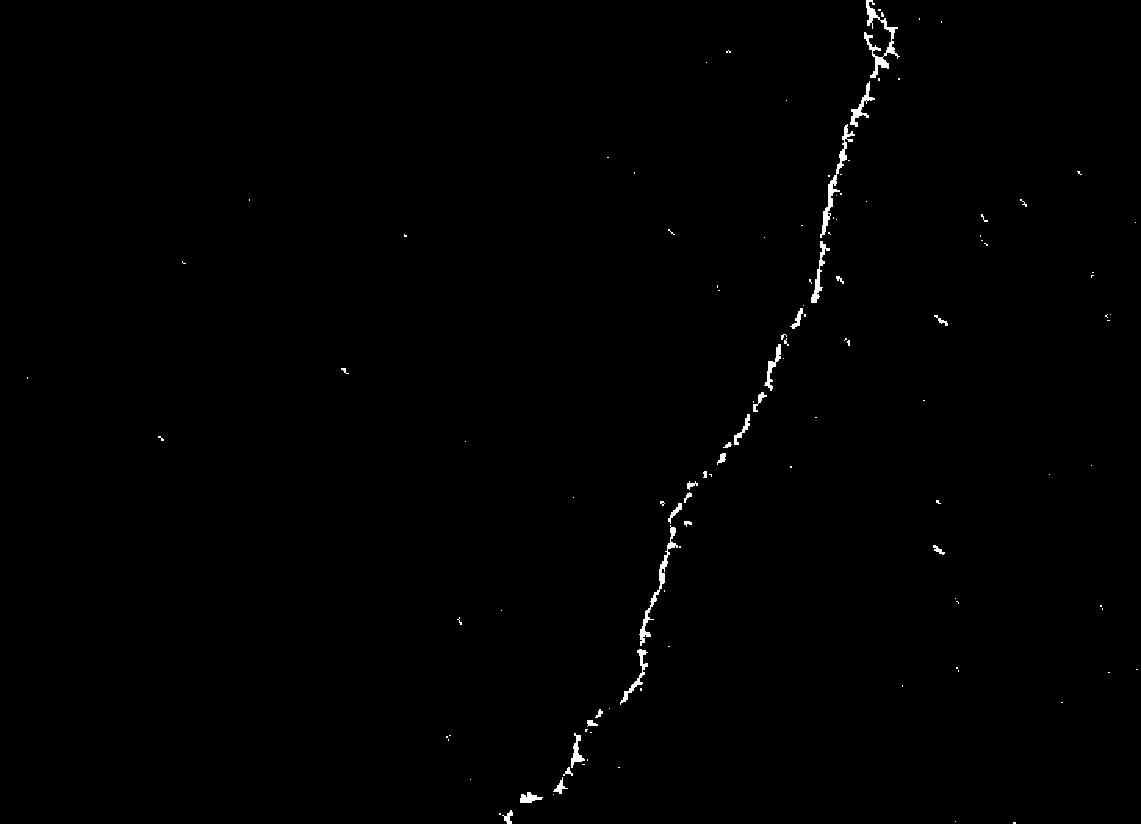}\label{fig:tabc}\end{subfigure}&
			\begin{subfigure}{0.15\textwidth}\centering\includegraphics[width=\linewidth]{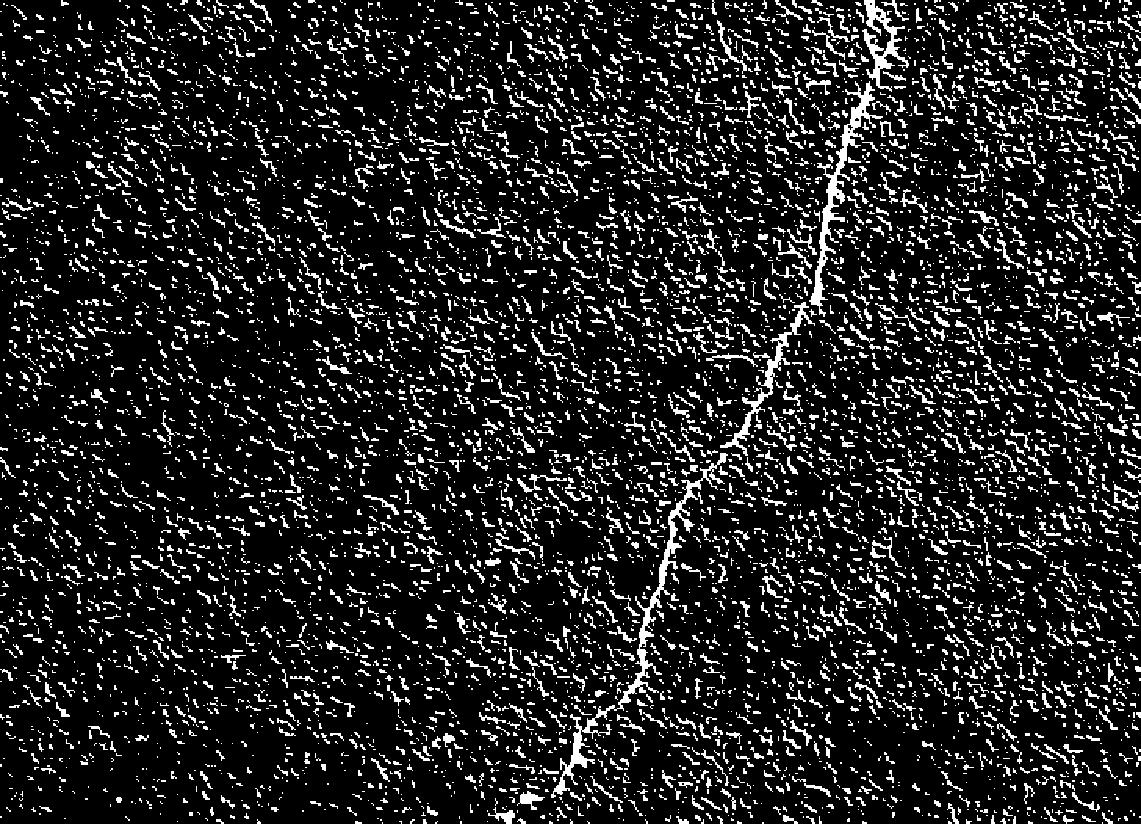}\label{fig:tabc}\end{subfigure}\\[7ex]
			&\begin{subfigure}{0.15\textwidth}\centering\includegraphics[width=\linewidth]{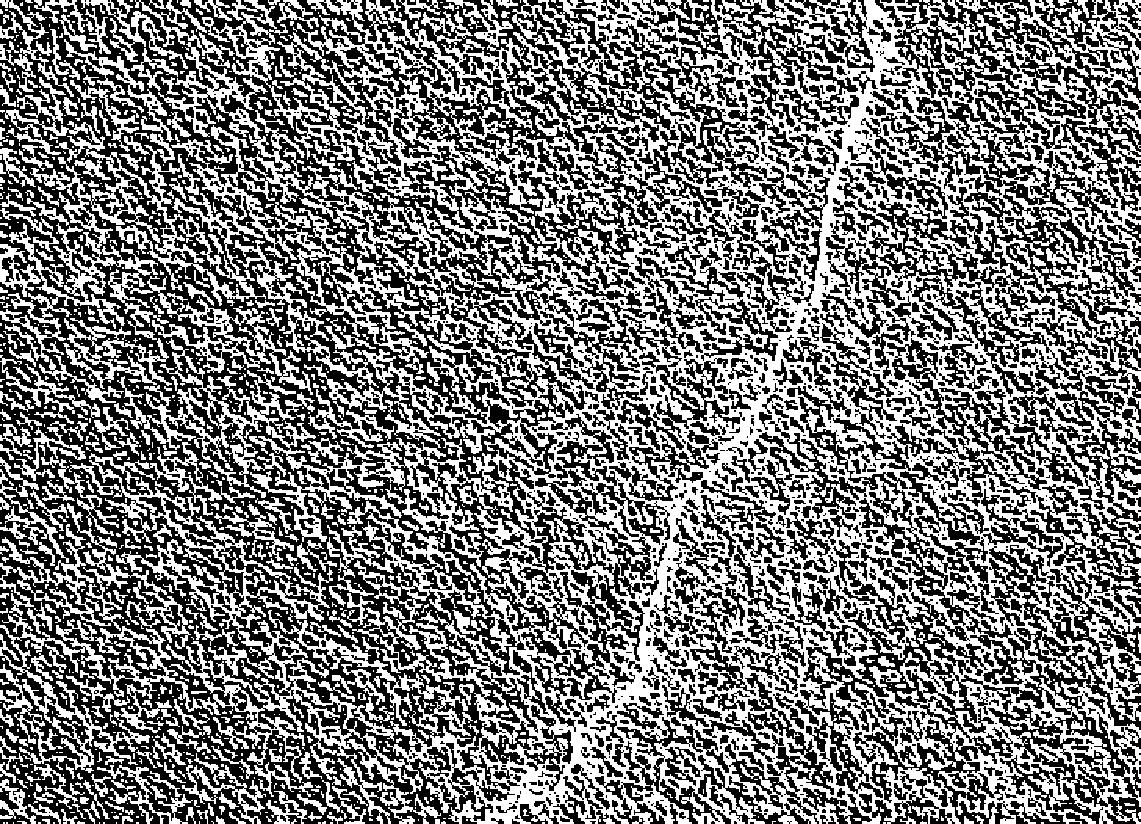}\label{fig:taba}\end{subfigure}&
			\begin{subfigure}{0.15\textwidth}\centering\includegraphics[width=\linewidth]{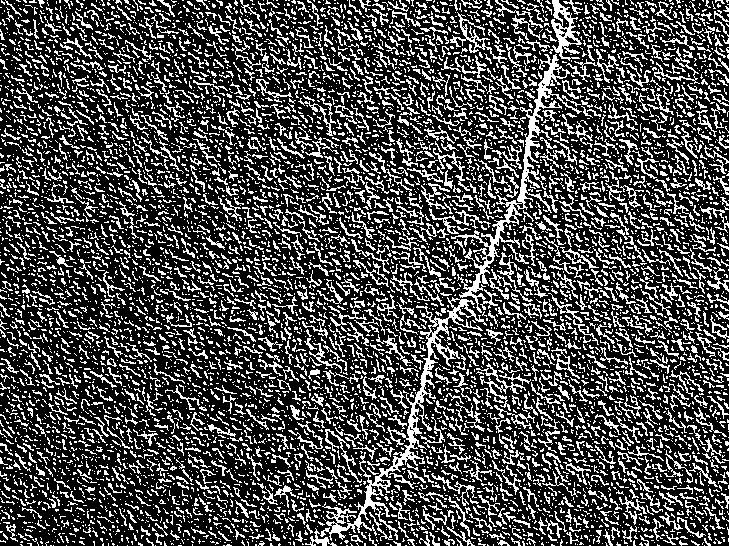}\label{fig:tabc}\end{subfigure}&
			\begin{subfigure}{0.15\textwidth}\centering\includegraphics[width=\linewidth]{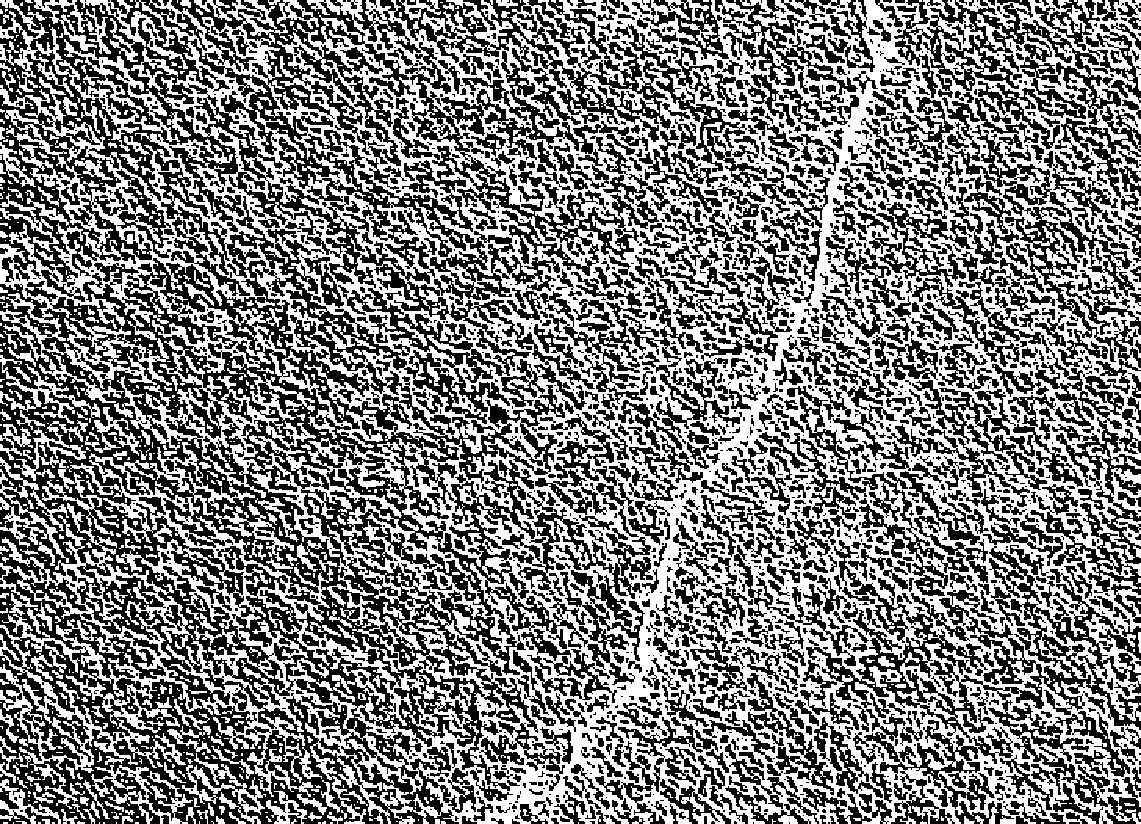}\label{fig:tabc}\end{subfigure}&
			\begin{subfigure}{0.15\textwidth}\centering\includegraphics[width=\linewidth]{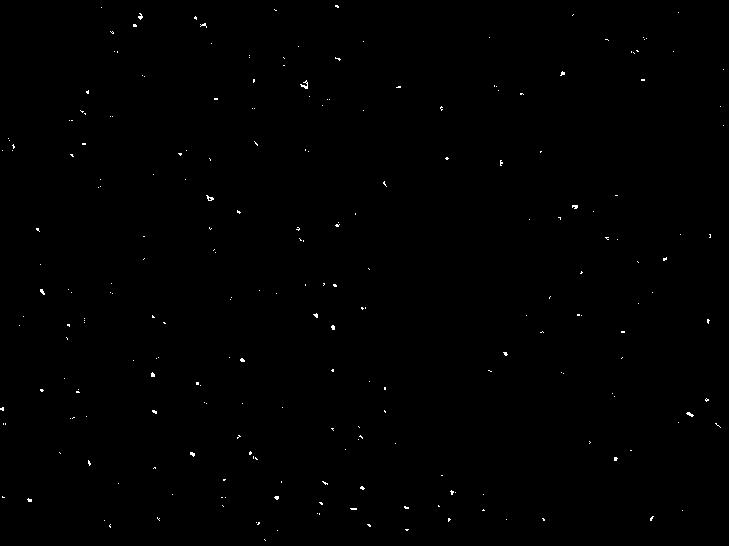}\label{fig:tabc}\end{subfigure}\\[8ex]
			
			
		\end{tabular}
	\end{center}
	\captionof{figure}{Defect detection results. First row: image name, original image, ground truth, our result, Sauvola; second row: detection respectively by Otsu, VE, ITTH and SDD.}
	\label{fig:ImCrack1}
\end{table*}

\begin{table*}[]
	\renewcommand{\arraystretch}{1.3}
	\footnotesize\addtolength{\tabcolsep}{-3pt}
	\begin{center}
		\begin{tabular}{p{1.2cm}cccc}
			Image 6&\begin{subfigure}{0.15\textwidth}\centering\includegraphics[width=\linewidth]{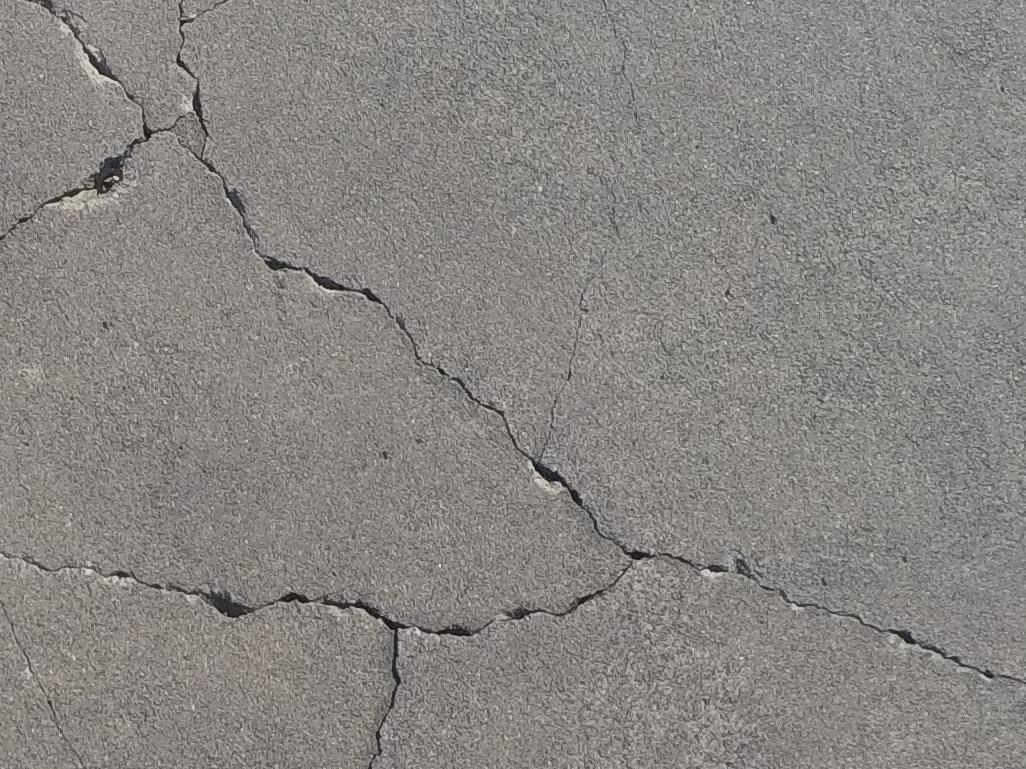}\label{fig:taba}\end{subfigure}&
			\begin{subfigure}{0.15\textwidth}\centering\includegraphics[width=\linewidth]{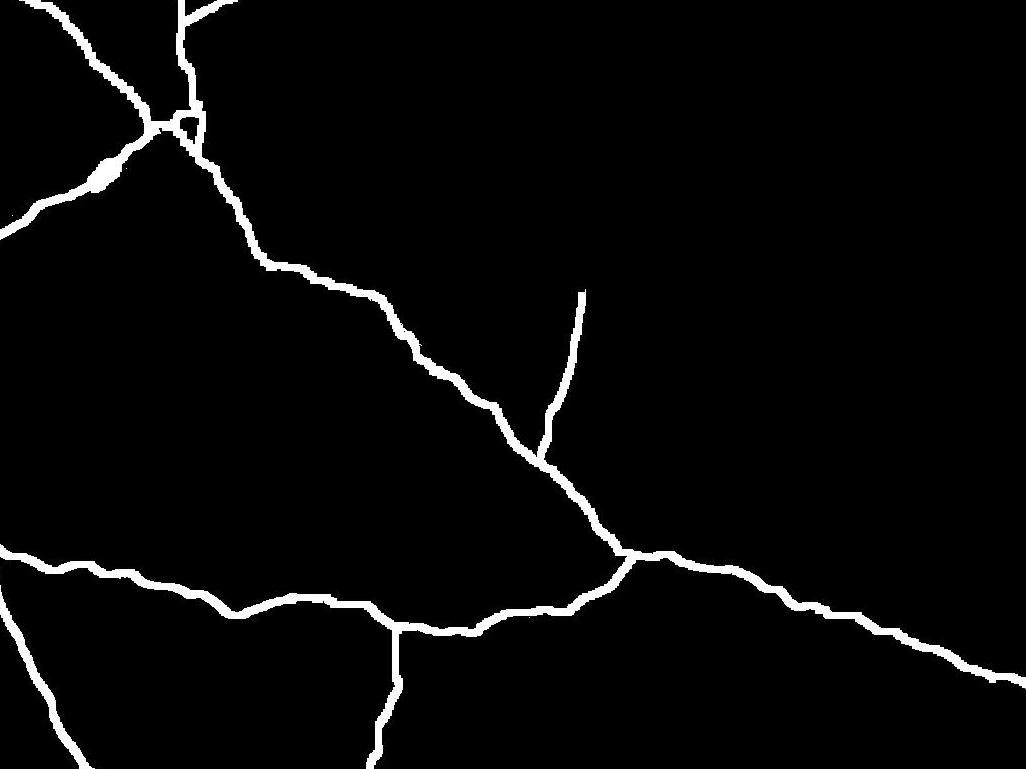}\label{fig:tabc}\end{subfigure}&
			\begin{subfigure}{0.15\textwidth}\centering\includegraphics[width=\linewidth]{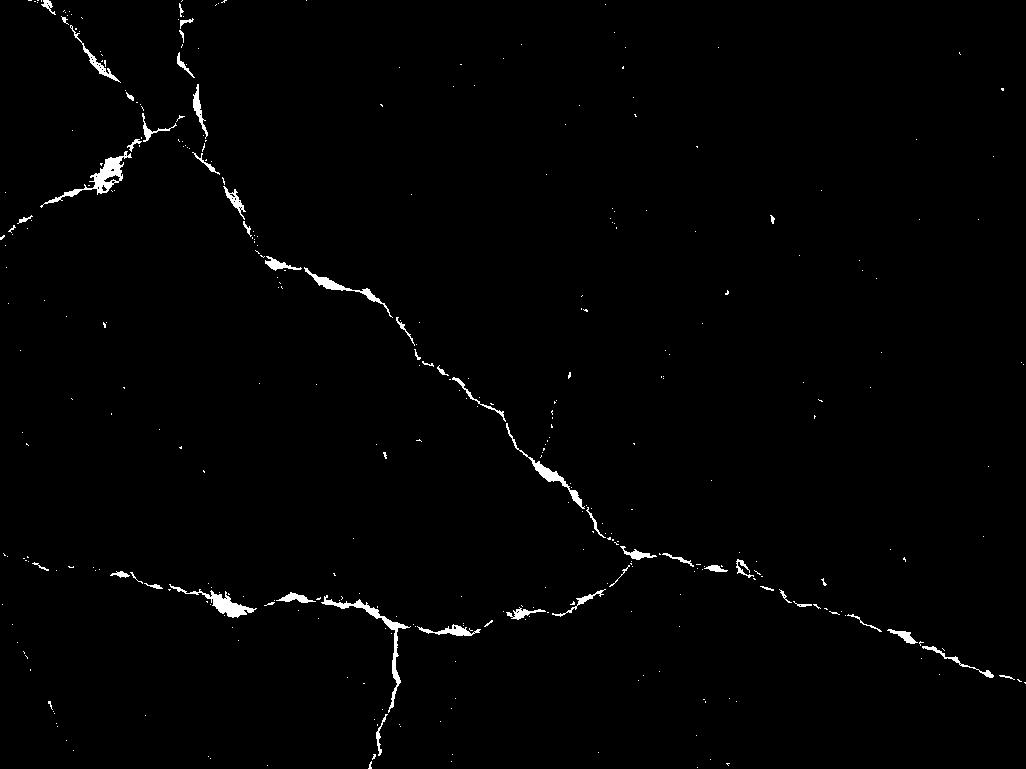}\label{fig:tabc}\end{subfigure}&
			\begin{subfigure}{0.15\textwidth}\centering\includegraphics[width=\linewidth]{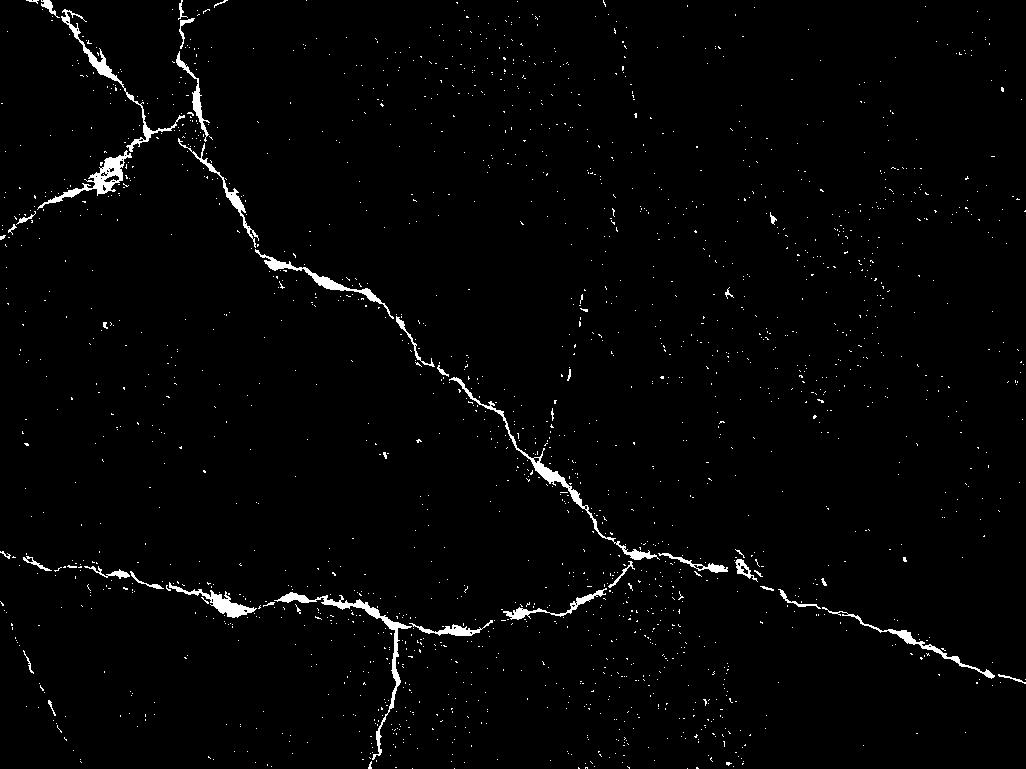}\label{fig:tabc}\end{subfigure}\\[7ex]
			&\begin{subfigure}{0.15\textwidth}\centering\includegraphics[width=\linewidth]{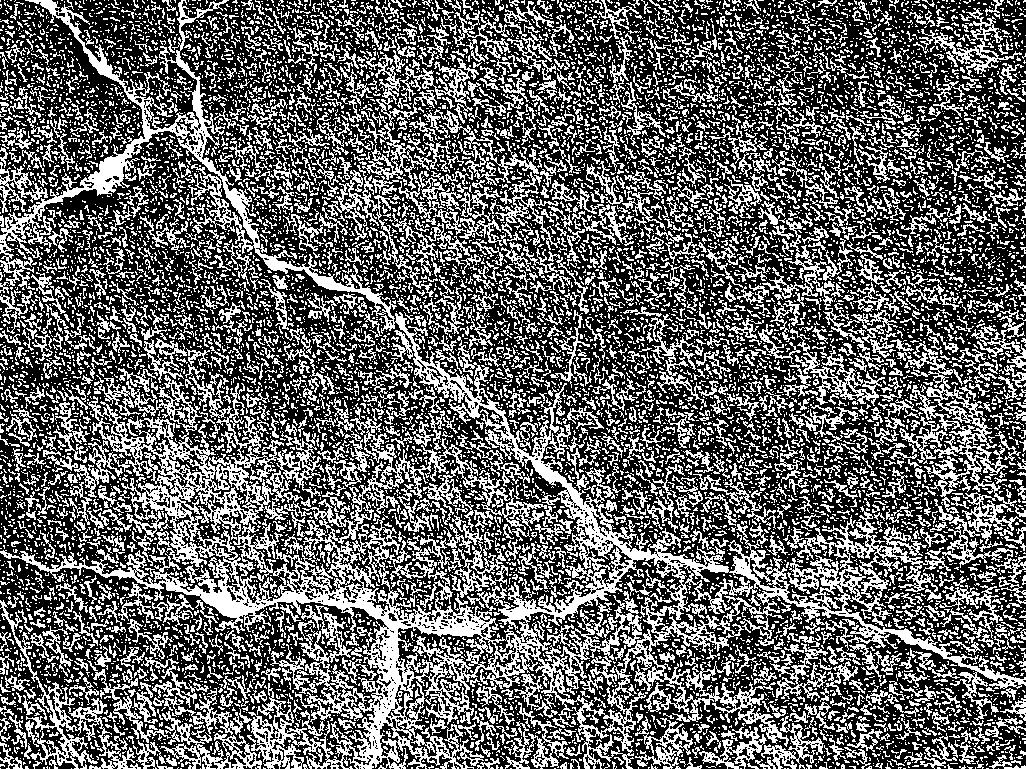}\label{fig:taba}\end{subfigure}&
			\begin{subfigure}{0.15\textwidth}\centering\includegraphics[width=\linewidth]{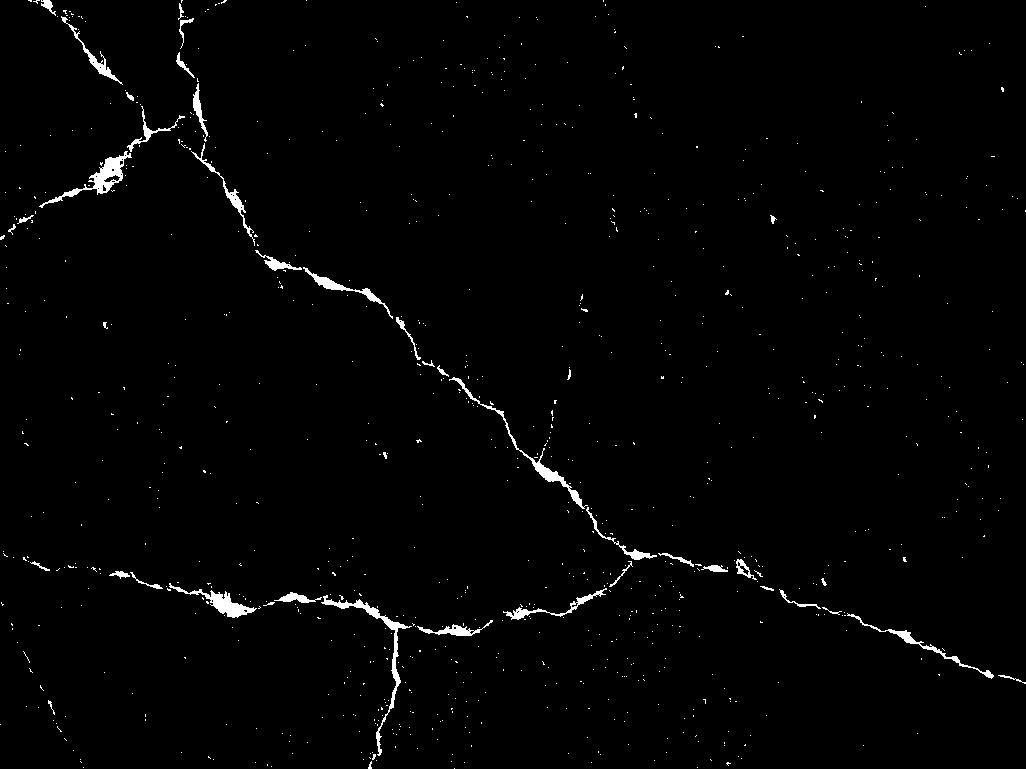}\label{fig:tabc}\end{subfigure}&
			\begin{subfigure}{0.15\textwidth}\centering\includegraphics[width=\linewidth]{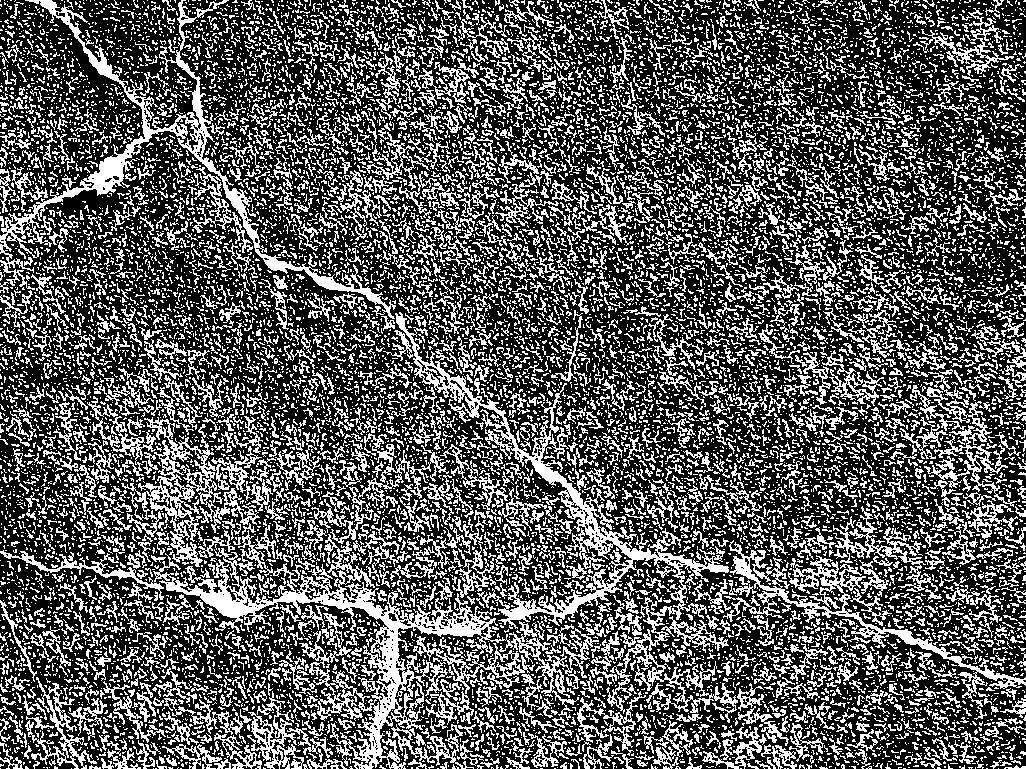}\label{fig:tabc}\end{subfigure}&
			\begin{subfigure}{0.15\textwidth}\centering\includegraphics[width=\linewidth]{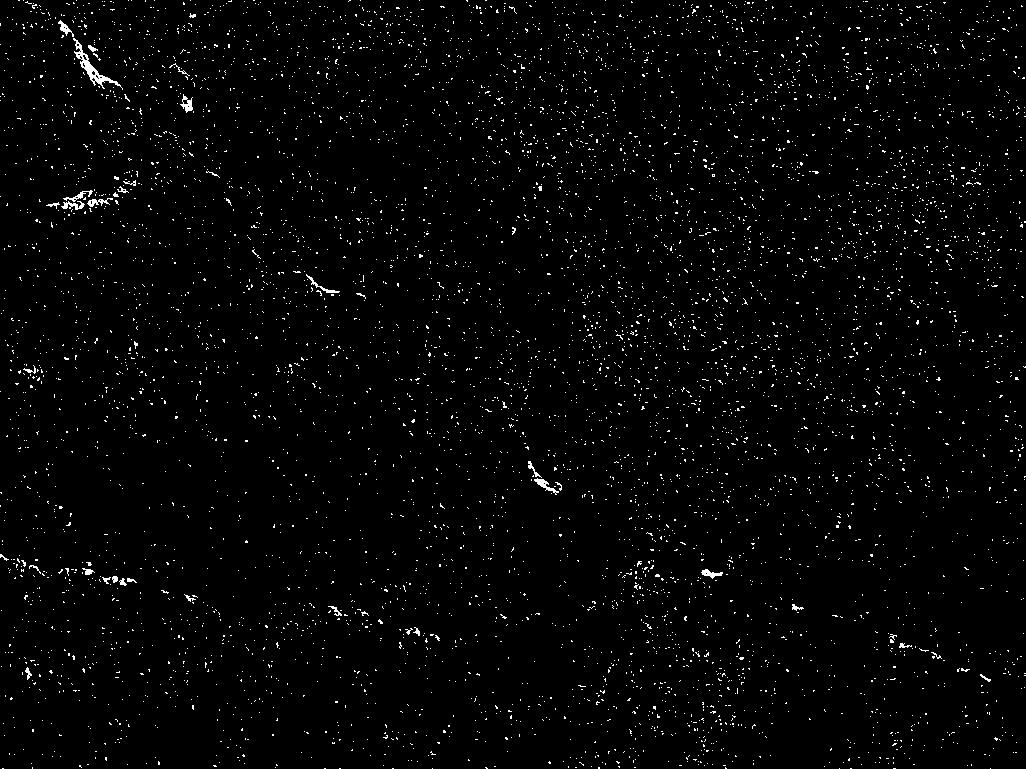}\label{fig:tabc}\end{subfigure}\\[8ex]
			Image 7&\begin{subfigure}{0.15\textwidth}\centering\includegraphics[width=\linewidth]{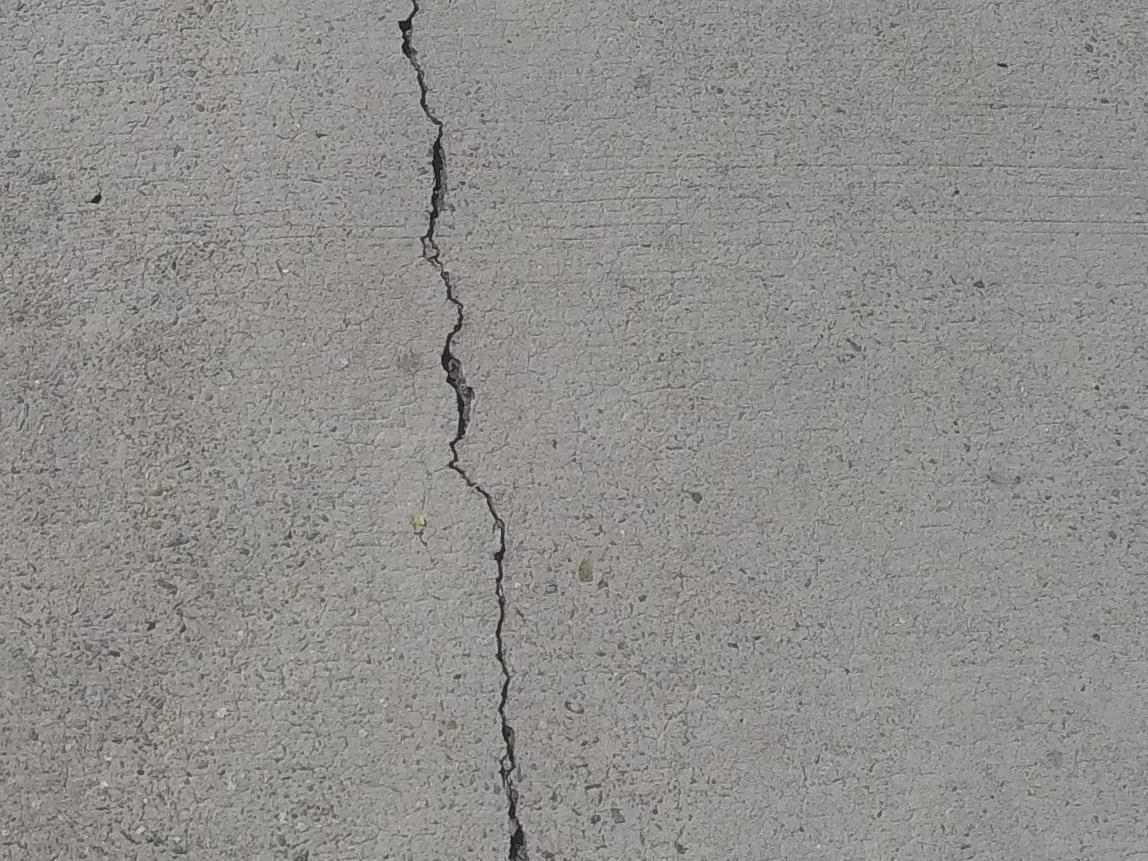}\label{fig:taba}\end{subfigure}&
			\begin{subfigure}{0.15\textwidth}\centering\includegraphics[width=\linewidth]{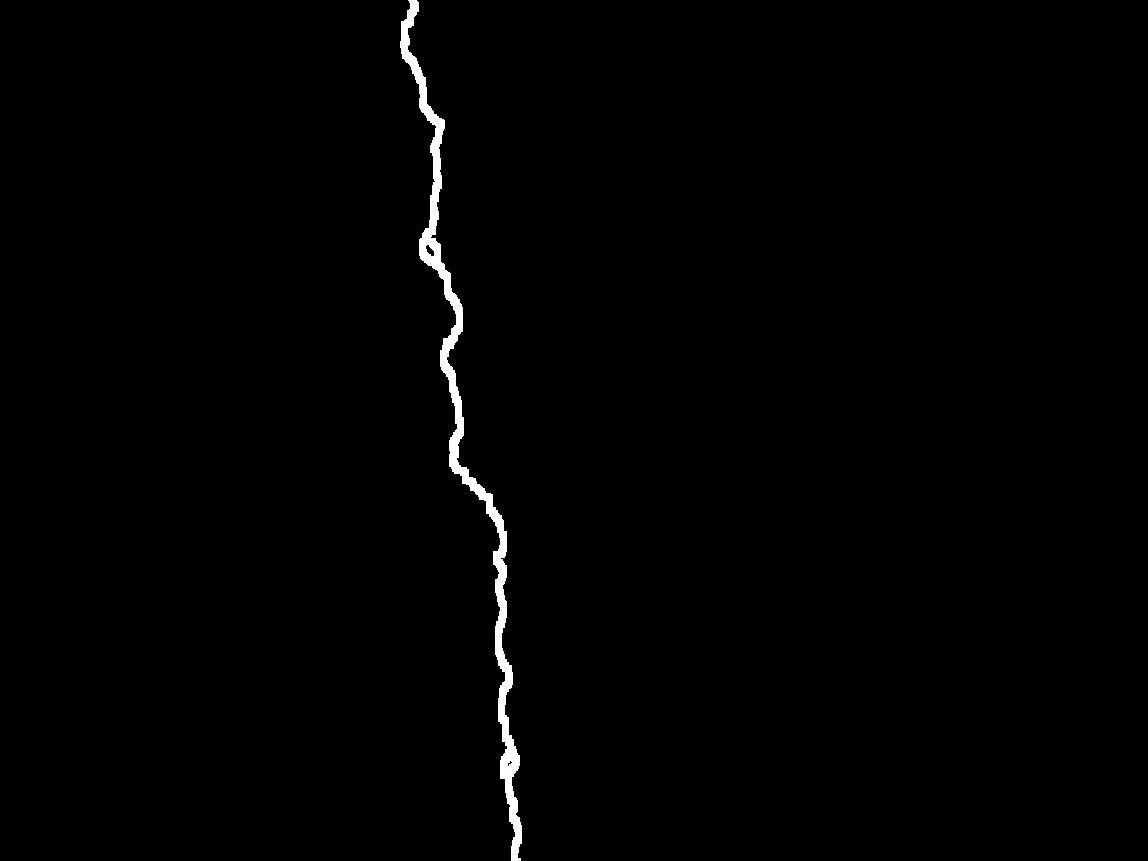}\label{fig:tabc}\end{subfigure}&
			\begin{subfigure}{0.15\textwidth}\centering\includegraphics[width=\linewidth]{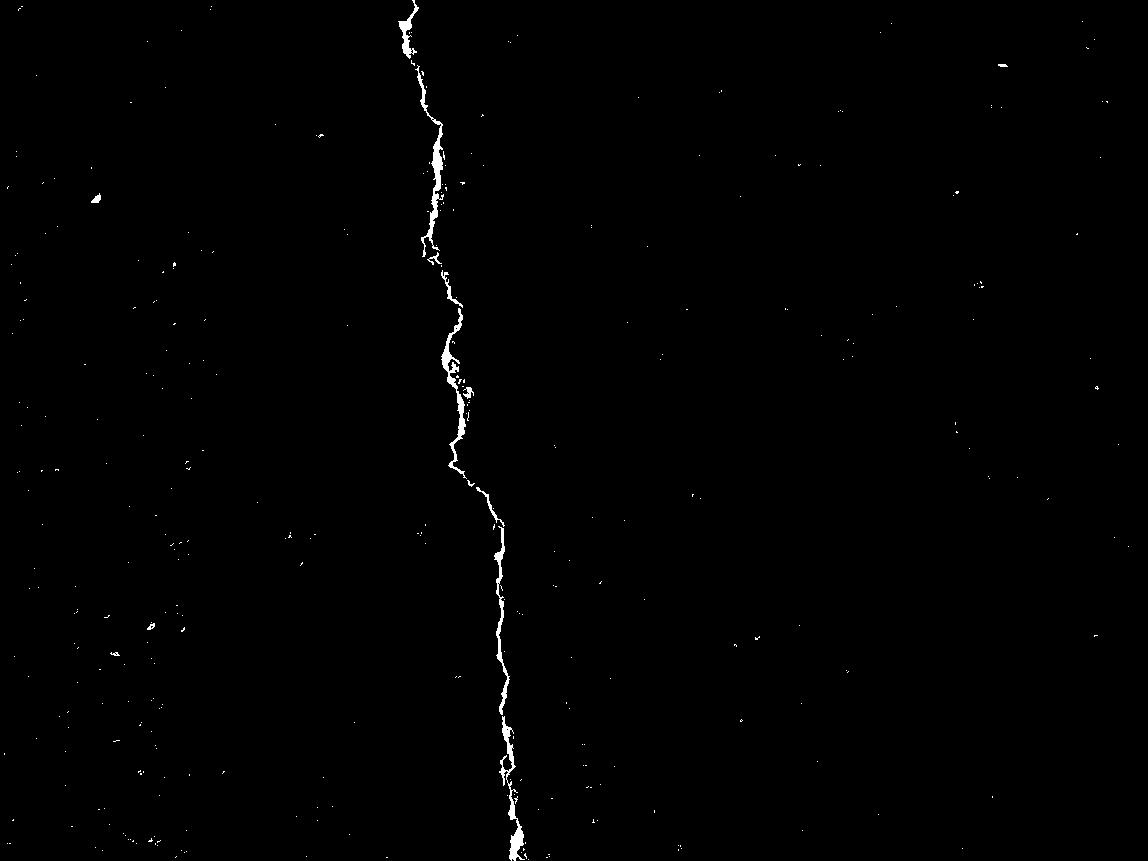}\label{fig:tabc}\end{subfigure}&
			\begin{subfigure}{0.15\textwidth}\centering\includegraphics[width=\linewidth]{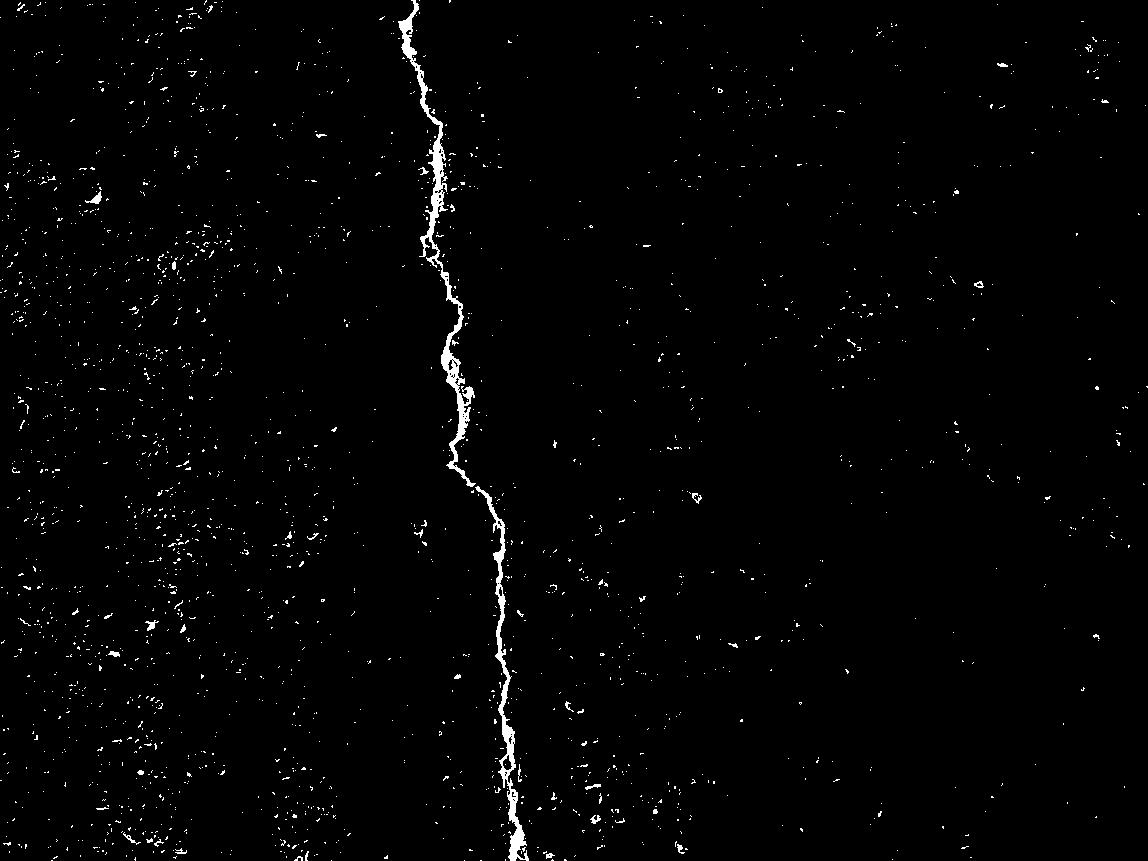}\label{fig:tabc}\end{subfigure}\\[7ex]
			&\begin{subfigure}{0.15\textwidth}\centering\includegraphics[width=\linewidth]{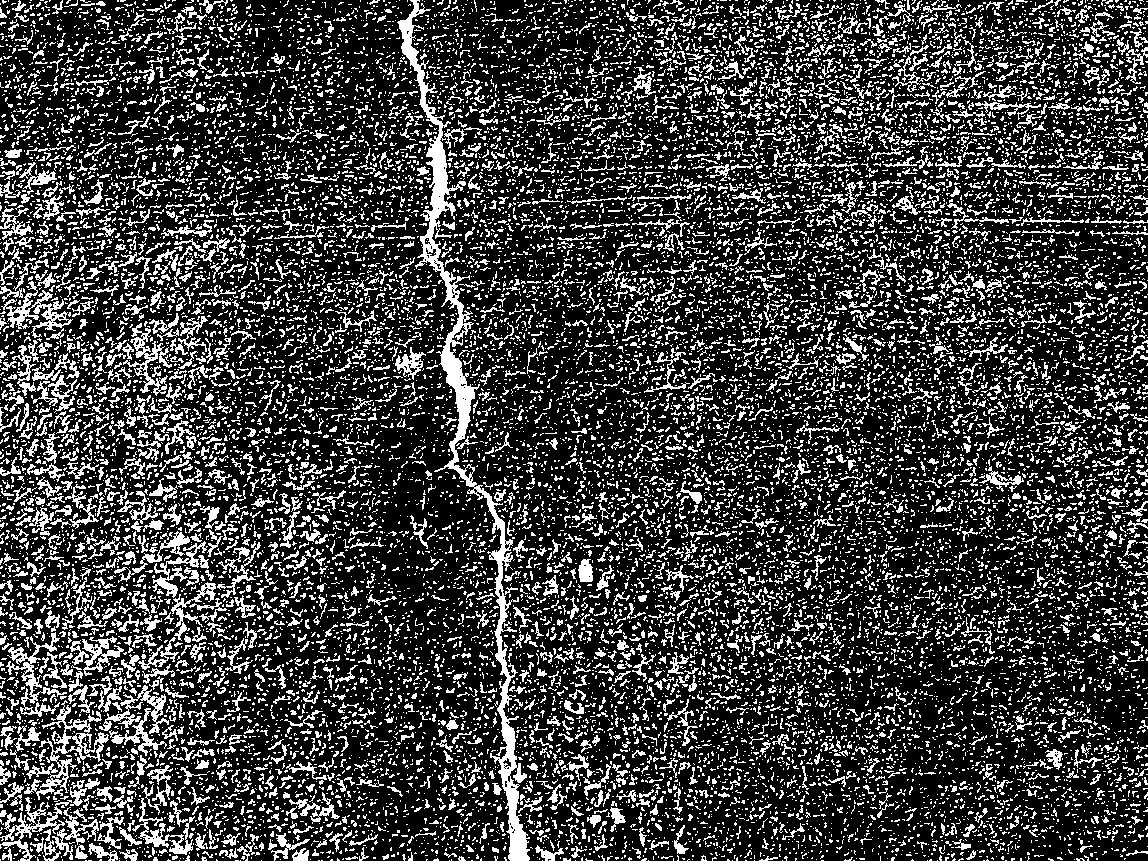}\label{fig:taba}\end{subfigure}&
			\begin{subfigure}{0.15\textwidth}\centering\includegraphics[width=\linewidth]{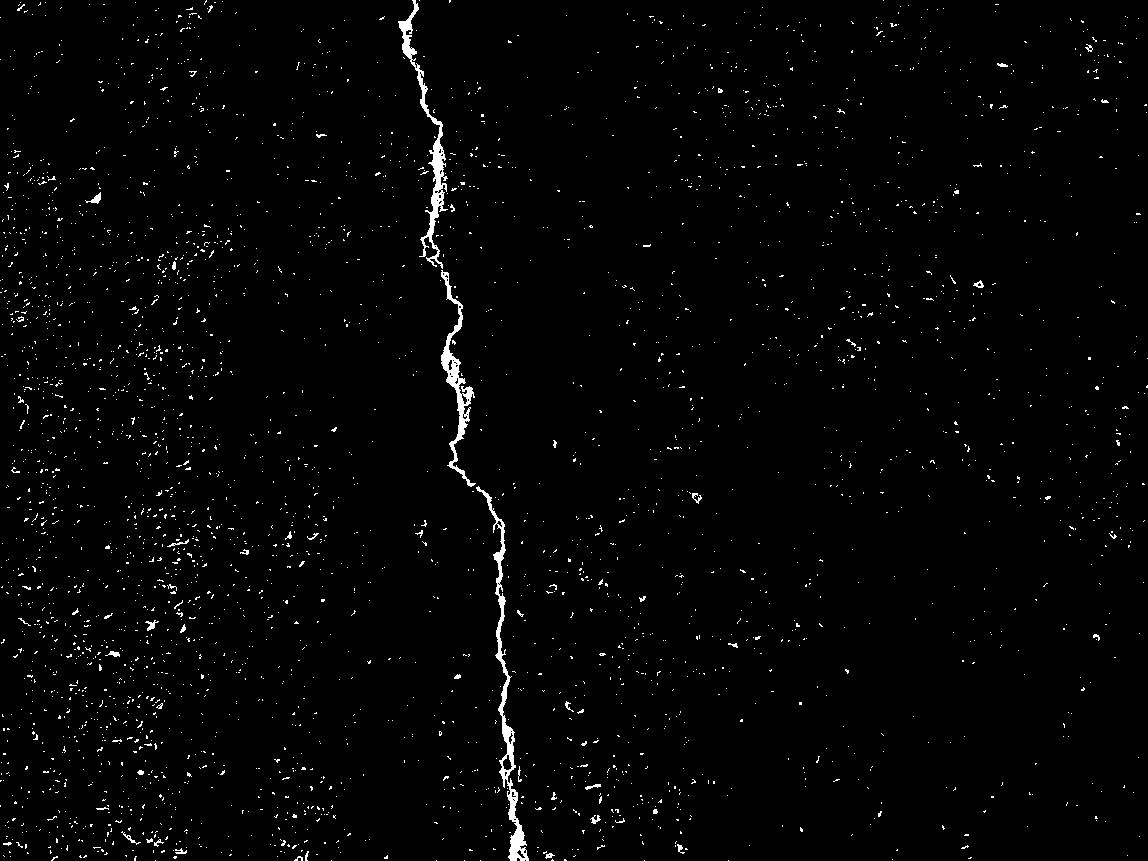}\label{fig:tabc}\end{subfigure}&
			\begin{subfigure}{0.15\textwidth}\centering\includegraphics[width=\linewidth]{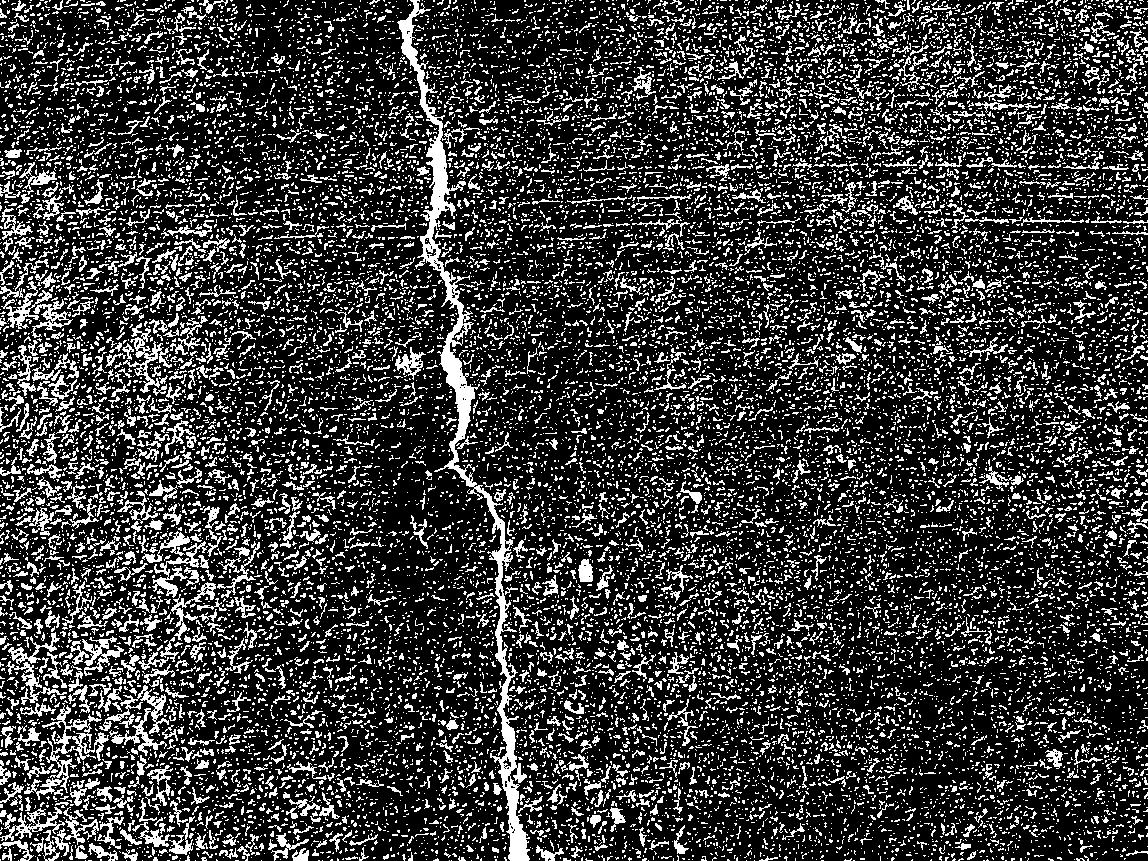}\label{fig:tabc}\end{subfigure}&
			\begin{subfigure}{0.15\textwidth}\centering\includegraphics[width=\linewidth]{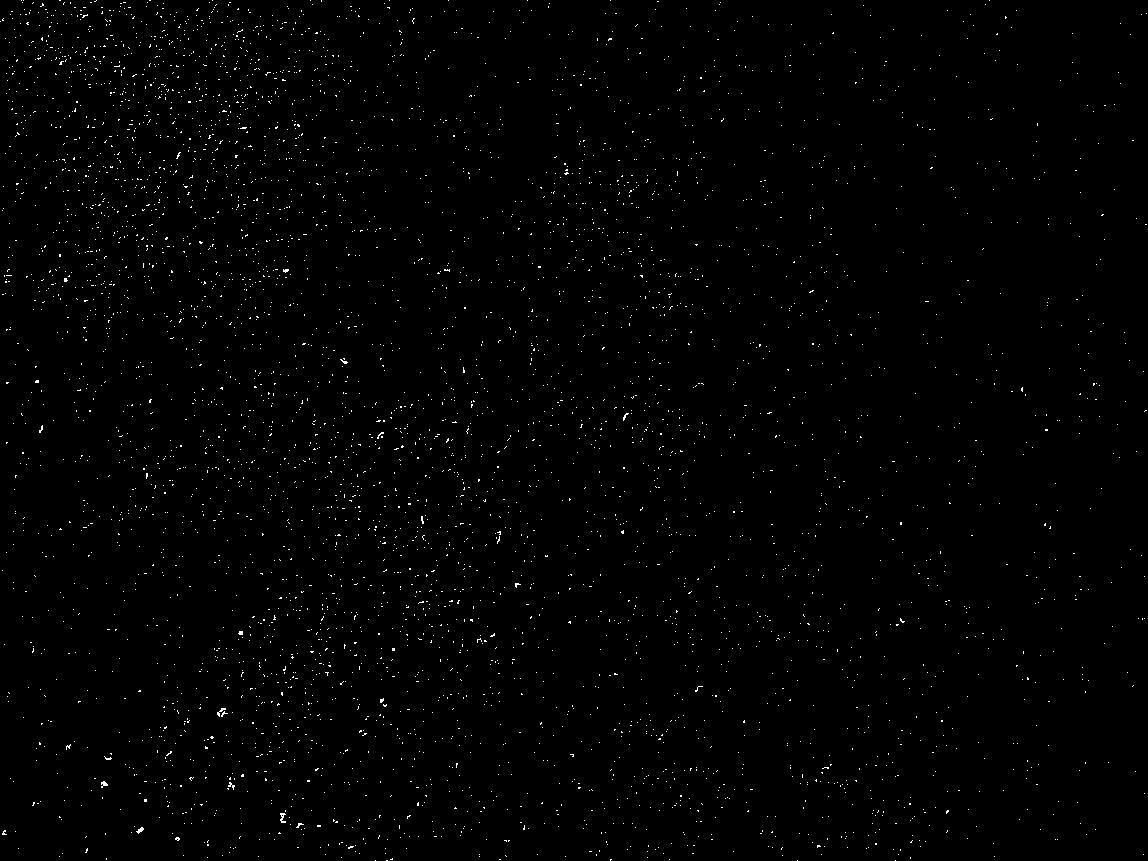}\label{fig:tabc}\end{subfigure}\\[8ex]
			Image 8&\begin{subfigure}{0.15\textwidth}\centering\includegraphics[width=\linewidth]{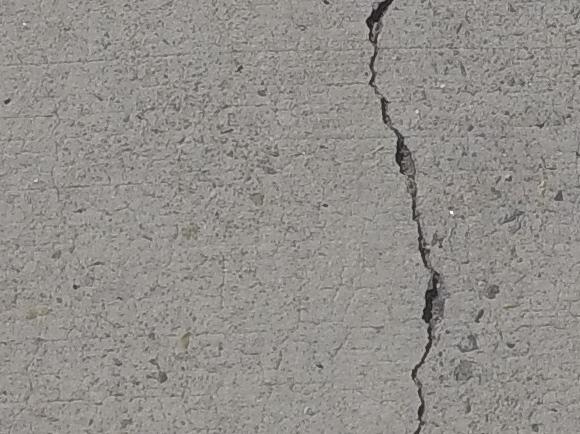}\label{fig:taba}\end{subfigure}&
			\begin{subfigure}{0.15\textwidth}\centering\includegraphics[width=\linewidth]{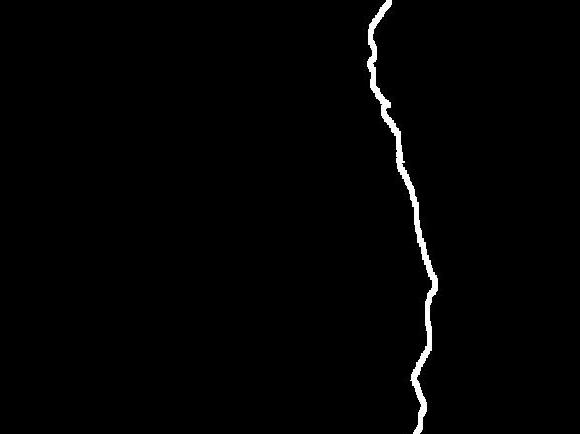}\label{fig:tabc}\end{subfigure}&
			\begin{subfigure}{0.15\textwidth}\centering\includegraphics[width=\linewidth]{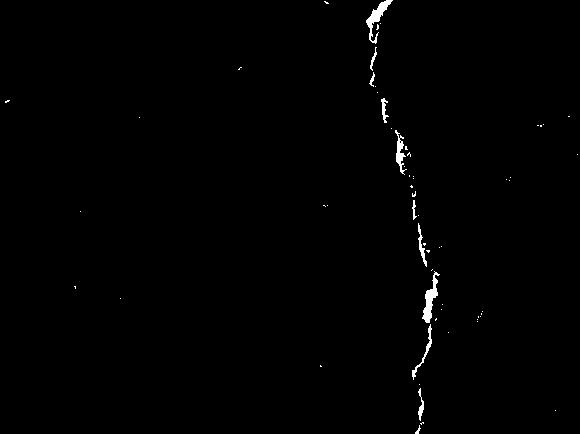}\label{fig:tabc}\end{subfigure}&
			\begin{subfigure}{0.15\textwidth}\centering\includegraphics[width=\linewidth]{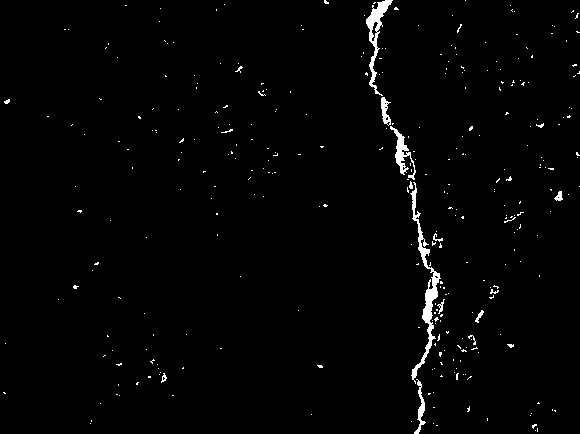}\label{fig:tabc}\end{subfigure}\\[7ex]
			&\begin{subfigure}{0.15\textwidth}\centering\includegraphics[width=\linewidth]{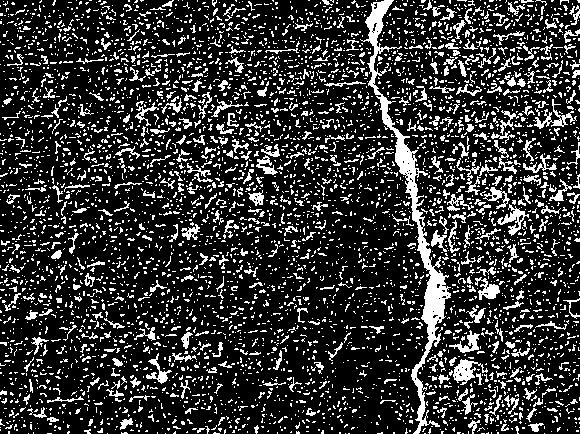}\label{fig:taba}\end{subfigure}&
			\begin{subfigure}{0.15\textwidth}\centering\includegraphics[width=\linewidth]{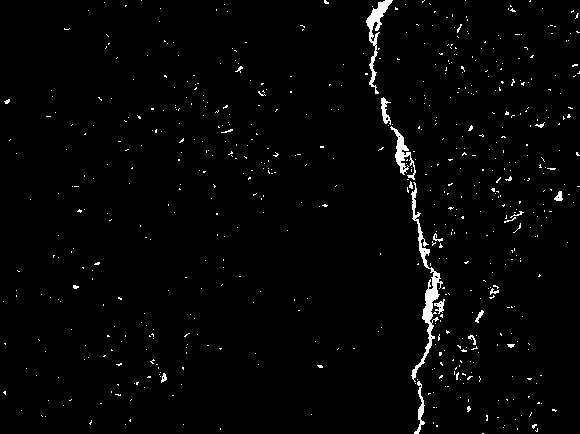}\label{fig:tabc}\end{subfigure}&
			\begin{subfigure}{0.15\textwidth}\centering\includegraphics[width=\linewidth]{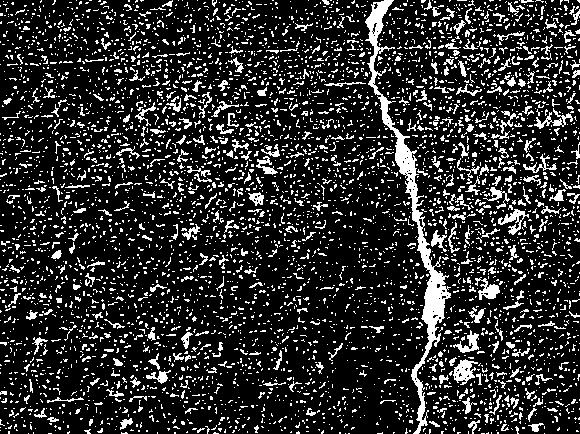}\label{fig:tabc}\end{subfigure}&
			\begin{subfigure}{0.15\textwidth}\centering\includegraphics[width=\linewidth]{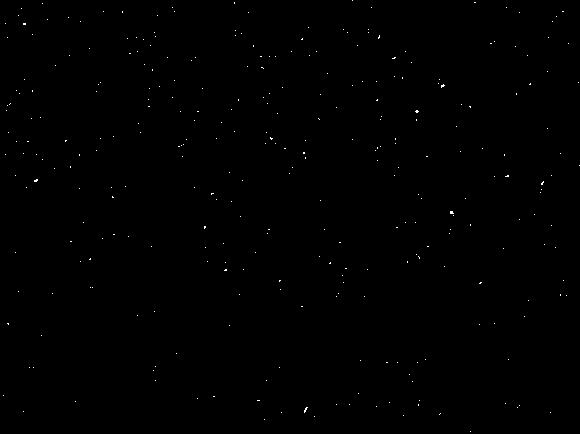}\label{fig:tabc}\end{subfigure}\\[8ex]
			Image 9&\begin{subfigure}{0.15\textwidth}\centering\includegraphics[width=\linewidth]{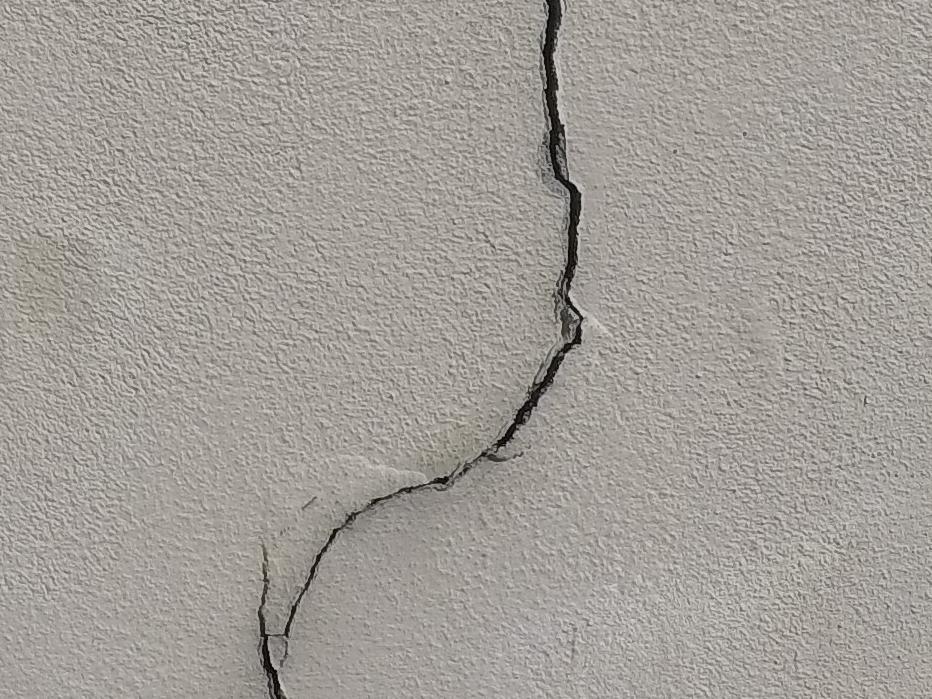}\label{fig:taba}\end{subfigure}&
			\begin{subfigure}{0.15\textwidth}\centering\includegraphics[width=\linewidth]{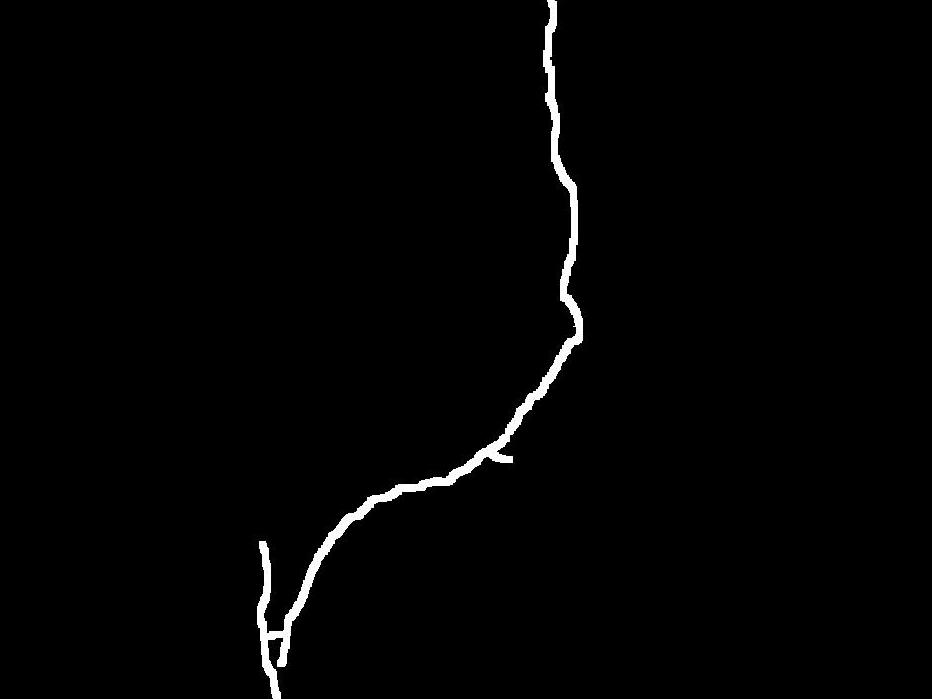}\label{fig:tabc}\end{subfigure}&
			\begin{subfigure}{0.15\textwidth}\centering\includegraphics[width=\linewidth]{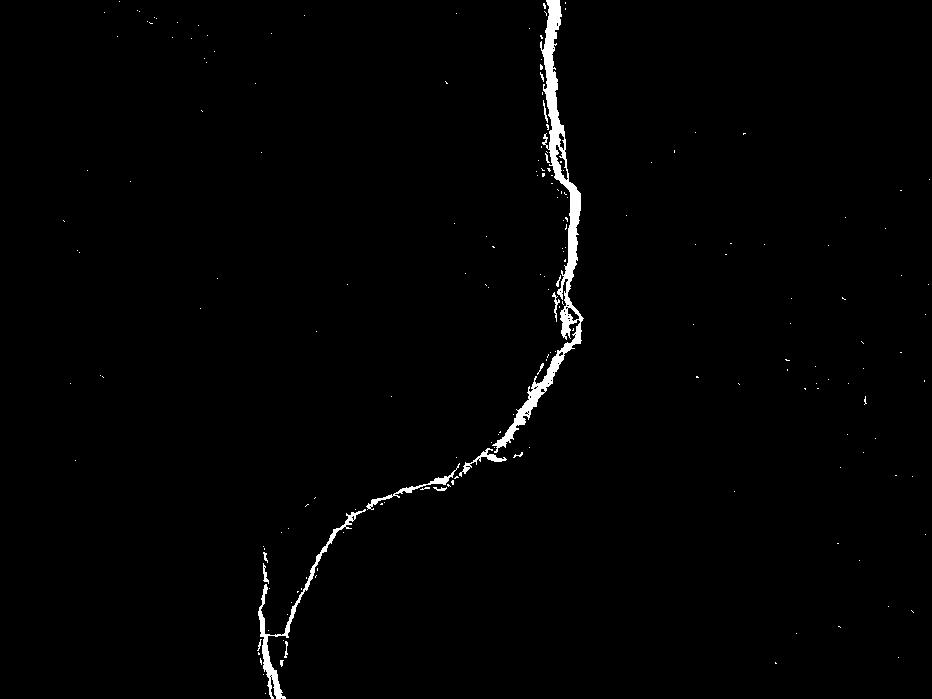}\label{fig:tabc}\end{subfigure}
			&\begin{subfigure}{0.15\textwidth}\centering\includegraphics[width=\linewidth]{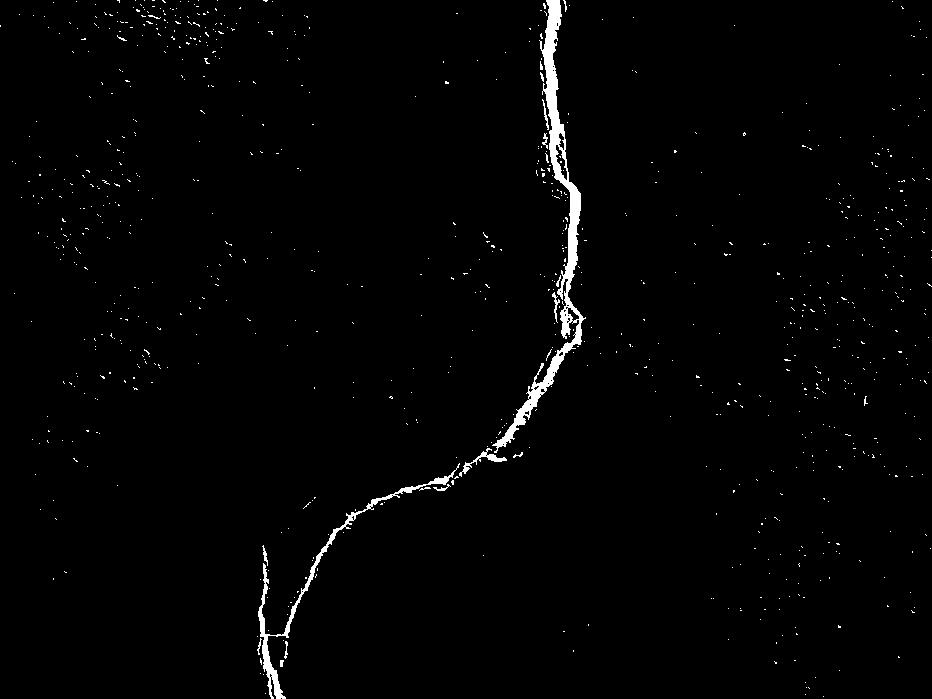}\label{fig:taba}\end{subfigure}\\[7ex]
			&\begin{subfigure}{0.15\textwidth}\centering\includegraphics[width=\linewidth]{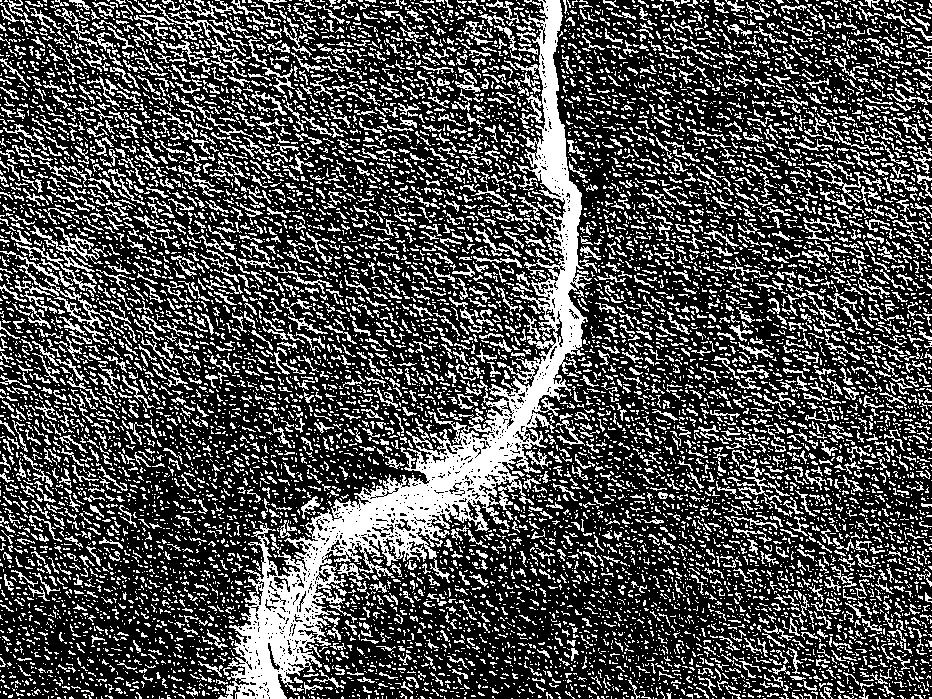}\label{fig:taba}\end{subfigure}&
			\begin{subfigure}{0.15\textwidth}\centering\includegraphics[width=\linewidth]{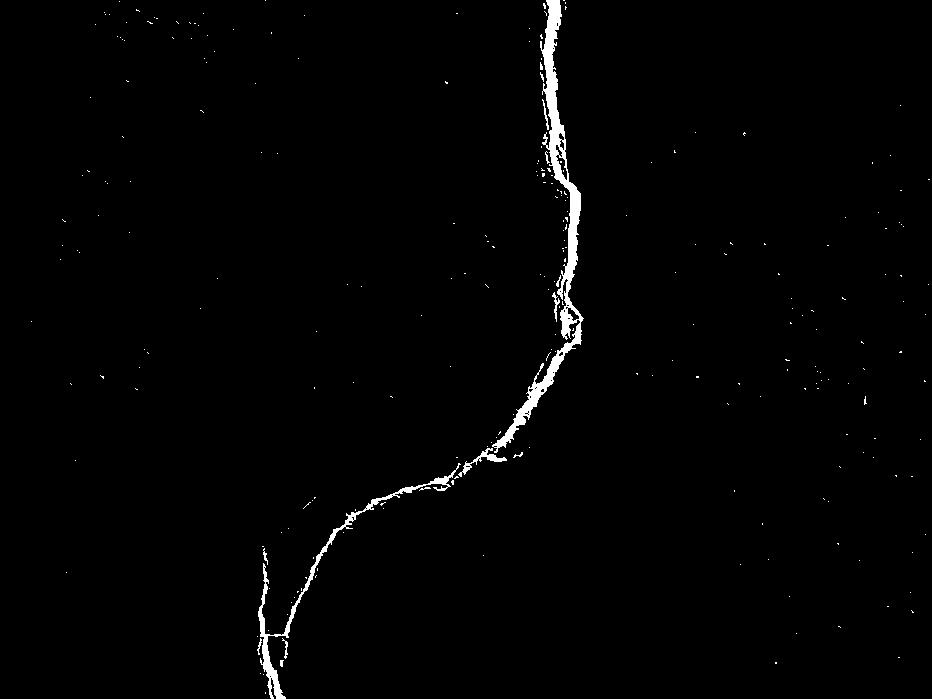}\label{fig:tabc}\end{subfigure}&
			\begin{subfigure}{0.15\textwidth}\centering\includegraphics[width=\linewidth]{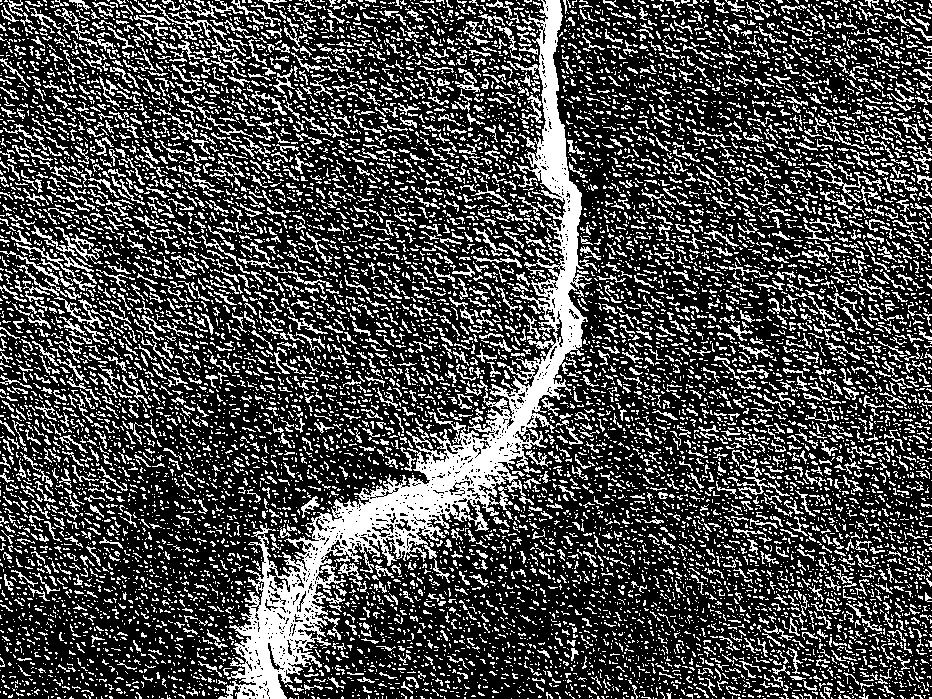}\label{fig:tabc}\end{subfigure}&
			\begin{subfigure}{0.15\textwidth}\centering\includegraphics[width=\linewidth]{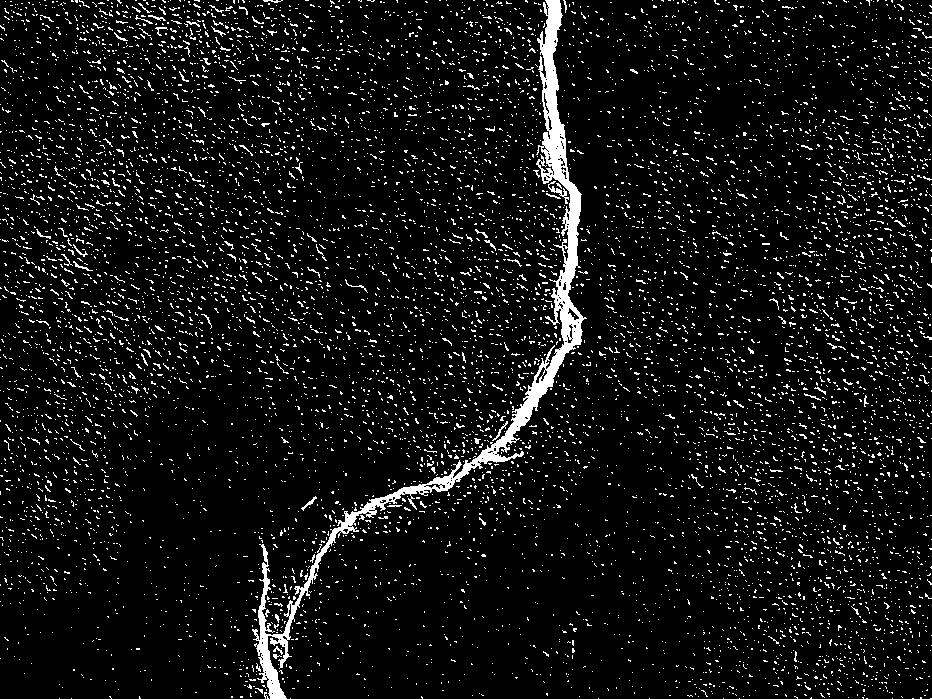}\label{fig:tabc}\end{subfigure}\\[8ex]
			

			
		\end{tabular}
	\end{center}
	\captionof{figure}{Defect detection results. First row: image name, original image, ground truth, our result, Sauvola; second row: detection respectively by Otsu, VE, ITTH and SDD.}
	\label{fig:ImCrack2}
\end{table*}

The comparison results can be quantitatively evaluated via the $F$-measure, a compromise between recall and precision \cite{PILLAI2017}. Let $t_p$ and $t_n$ be the correctly reported positive and negative results, whereas $f_p$ and $f_n$ be the falsely reported positive and negative results. The $F$-measure for binary classification is calculated as:
\begin{equation}
F=\dfrac{2pr}{p+r},
\end{equation}
where $p$ and $r$ denote the precision and recall measures defined respectively as: 
\begin{equation}
p=\dfrac{t_p}{t_p+f_p},
\end{equation}
\begin{equation}
r=\dfrac{t_p}{t_p+f_n}.
\end{equation}
Table \ref{tbl:ImCrack} reports the $F$ values obtained for each method which clearly shows the superiority of our proposed algorithm.

\begin{table*}[]
	\renewcommand{\arraystretch}{1.3}
	\renewcommand\thetable{II}
	\centering
	\caption{Comparison in defect detection between the proposed method and Sauvola, Otsu, VE, ITTH and SDD methods}
	\label{tbl:ImCrack}
	\begin{tabular}{lcccccc}
		\hline
		Image    & Our method      & Sauvola & Otsu   & VE              & ITTH   & SDD    \\
		\hline
		Image 1  & \textbf{0.9889} & 0.9163  & 0.8800 & 0.9206          & 0.9416 & 0.6881 \\
		Image 2  & \textbf{0.9774} & 0.9743  & 0.5101 & 0.9736          & 0.5112 & 0.7626 \\
		Image 3  & \textbf{0.9818} & 0.9547  & 0.5083 & \textbf{0.9818} & 0.5076 & 0.8534 \\
		Image 4  & \textbf{0.9542} & 0.6251  & 0.5141 & 0.5203          & 0.5147 & 0.6974 \\
		Image 5  & \textbf{0.9451} & 0.5239  & 0.5071 & 0.5119          & 0.5069 & 0.4965 \\
		Image 6  & \textbf{0.9140} & 0.8256  & 0.5183 & 0.8774          & 0.5208 & 0.5198 \\
		Image 7  & \textbf{0.8693} & 0.6776  & 0.5117 & 0.6371          & 0.5125 & 0.5017 \\
		Image 8  & \textbf{0.8685} & 0.7333  & 0.5214 & 0.6809          & 0.5244 & 0.5029 \\
		Image 9  & \textbf{0.8474} & 0.7876  & 0.5197 & 0.8367          & 0.5208 & 0.5730 \\
		Image 10 & \textbf{0.7537} & 0.6893  & 0.5179 & 0.7362          & 0.5222 & 0.5706 \\
		\bottomrule
	\end{tabular}
\end{table*}


We also evaluated the processing time executed by using MATLAB R2018b on an Intel(R) Core(TM) i5-5300U CPU @2.30 GHz with 64 bit Windows 7. The average processing time is 50 ms per image or 20 frames per second which is sufficient for real-time inspection. 

\section{Conclusion} \label{conclusion}
In this paper, we have presented a system architecture for surface inspection in real time using multiple UAVs. The system features a new communication platform based on the Internet of Things which can exploit processing capabilities of RCU and reduce communication distance burdens. Here, a multi-layer paradigm has been introduced to integrate various modules of a complicated system to fulfil a common inspection task with UAVs following a predefined geometrical shape. The coordination among UAVs is then carried out via a novel formation algorithm based on the angle-encoded PSO that can create optimal flying paths subject to constraints of the surface to be inspected and collision avoidance. Finally, a histogram-based segmentation algorithm has been developed for online detection of potential defects with high accuracy. A number of experiments have been conducted with real-world defects detected from the data acquired by a triangular formation of UAVs. Comparisons and discussions have been also presented to evaluate the performance of the proposed system.

\bibliography{IEEEabrv,bibi}

\balance

\end{document}